\documentclass[times,twocolumn,final]{elsarticle}

\usepackage{medima}
\usepackage{framed,multirow}

\usepackage{amssymb}
\usepackage{latexsym}

\usepackage{url}
\usepackage{xcolor}

\usepackage{hyperref}

\definecolor{newcolor}{rgb}{.8,.349,.1}

\journal{Medical Image Analysis}

\usepackage{mathtools}
\usepackage{nicefrac}
\usepackage{tikz}
\usepackage{bbm}
\usepackage{comment}
\usepackage{algorithm}
\usepackage{xspace}
\usepackage{tikz}

\newcommand{\sort}{\text{\tt sort}}

\newcommand{\plaus}[2]{\lambda^{#1}_{#2}}

\newcommand{\numcond}{K}

\newcommand{\block}[3]{b_{{#1}{#2}{#3}}}

\newcommand{\choice}[3]{{\sigma^{{#1}}_{{#2}{#3}}}\xspace}

\newcommand{\trace}{\text{Tr}}
\newcommand{\diag}{\text{diag}}

\newcommand{\ind}[1]{\delta\left[{#1}\right]}

\usepackage{xspace}
\makeatletter
\DeclareRobustCommand\onedot{\futurelet\@let@token\@onedot}
\def\@onedot{\ifx\@let@token.\else.\null\fi\xspace}

\makeatother

\newcommand{\E}{\mathbb{E}}
\newcommand{\PIRN}{PrIRN\xspace}

\newcommand{\red}[1]{#1}
\begin{document}

\verso{David Stutz \textit{et~al.}}

\begin{frontmatter}

\title{Evaluating medical AI systems in dermatology under uncertain ground truth}

\author[1]{David Stutz\fnref{fn1}}
\cortext[cor1]{Corresponding authors: dstutz@google.com, taylancemgil@google.com, agroy@google.com}
\author[1]{Ali Taylan Cemgil\fnref{fn1}}
\author[2]{Abhijit Guha Roy\fnref{fn1}}
\author[1]{Tatiana Matejovicova\fnref{fn1}}
\author[1,3]{Melih Barsbey\fnref{fn1}}
\fntext[fn1]{Equal first authors}
\author[2]{Patricia Strachan}
\author[2]{Mike Schaekermann}
\author[2]{Jan Freyberg}
\author[2]{Rajeev Rikhye}
\author[2]{Beverly Freeman}
\author[2]{Javier Perez Matos}
\author[2]{Umesh Telang}
\author[2]{Dale R. Webster}
\author[2]{Yuan Liu}
\author[2]{Greg S. Corrado}
\author[2]{Yossi Matias}
\author[1]{Pushmeet Kohli}
\author[2]{Yun Liu\fnref{fn2}}
\author[1]{Arnaud Doucet\fnref{fn2}}
\author[2]{Alan Karthikesalingam\fnref{fn2}}
\fntext[fn2]{Equal last authors}

\address[1]{Google DeepMind}
\address[2]{Google}
\address[3]{Bogazici University}

\received{XXX}
\finalform{XXX}
\accepted{XXX}
\availableonline{XX}
\communicated{XXX}

\begin{abstract}
For safety, medical AI systems undergo thorough evaluations before deployment, validating their predictions against a ground truth which is assumed to be fixed and certain. However, in medical applications, this ground truth is often curated in the form of differential diagnoses provided by multiple experts. While a single differential diagnosis reflects the uncertainty in one expert assessment, multiple experts introduce another layer of uncertainty through  potential disagreement. Both forms of uncertainty are ignored in standard evaluation which aggregates these differential diagnoses to a single label. In this paper, we show that ignoring uncertainty leads to overly optimistic estimates of model performance, therefore underestimating risk associated with particular diagnostic decisions. Moreover, point estimates largely ignore dramatic differences in uncertainty of individual cases.  To this end, we propose a \emph{statistical aggregation} approach, where we infer a distribution on probabilities of underlying medical condition candidates themselves, based on observed annotations. This formulation naturally accounts for the potential disagreements between different experts, as well as uncertainty stemming from individual differential diagnoses, capturing the entire \emph{ground truth uncertainty}. Practically, our approach boils down to generating multiple samples of medical condition probabilities, then evaluating and averaging performance metrics based on these sampled probabilities, instead of relying on a single point estimate. This allows us to provide uncertainty-adjusted estimates of common metrics of interest such as top-$k$ accuracy and average overlap. In the skin condition classification problem of \citep{LiuNATURE2020}, our methodology reveals significant ground truth uncertainty for most data points and demonstrates that standard evaluation techniques can overestimate performance by several percentage points. We conclude that, while assuming a crisp ground truth \textit{may} be acceptable for many AI applications, a more nuanced evaluation protocol acknowledging the inherent complexity and variability of differential diagnoses should be utilized in medical diagnosis.
\end{abstract}


\end{frontmatter}

\section{Introduction}
\label{sec:introduction}

Prior to deployment, predictive AI models are usually evaluated by comparing model predictions to a known ground truth on a held-out test set. In a supervised classification context, this typically assumes the availability of a unique and certain ground truth per example.
Commonly used supervised training and evaluation procedures are usually built on this assumption. However, in certain domains, predominantly in health, such ground truth is often not available.

In medical diagnosis, uncertainty is the norm as doctors have to constantly make complex decisions with limited information. The common strategy to deal systematically with uncertainty is to come up with a differential diagnosis. A differential diagnosis is a working list of medical conditions ordered according to the perceived subjective plausibility of each condition. In a standard diagnosis process, a doctor would usually proceed to ask the patient further questions or perform additional tests to converge on a final diagnosis. However, in machine learning tasks annotated by such experts, their feedback remain in the form of differential diagnoses, reflecting the existing uncertainty in their assessment limited by the information provided. Moreover, multiple physicians with different backgrounds and medical experiences, constrained by an imperfect labeling tool, will typically provide different differentials, especially on ambiguous cases \citep{FreemanHCOMP2021}.

In medical AI applications, ground truth labels are usually derived by simplistic aggregations of multiple differential diagnoses to fit existing standard machine learning evaluation metrics, such as top-$k$ accuracy.
\red{The choice of this aggregation mechanism is often not made explicit in evaluations.}
\red{More importantly,} this largely ignores the uncertainty and wealth of information in the potentially disagreeing expert annotations. 
In this paper, we show that ignoring this uncertainty in expert annotations leads to overly confident \textit{and} optimistic estimates of model performance, thereby underestimating risk associated with particular diagnostic decisions. Instead, we propose a principled evaluation procedure to obtain \red{uncertainty-adjusted estimates of the real model performance.}

\subsection{Ground truth uncertainty and annotator disagreement}

Throughout the paper, we consider two sources of ground truth uncertainty, \textit{annotation uncertainty} and \textit{inherent uncertainty}. Annotation uncertainty stems from labeling process complexities: even expert annotators can make mistakes, tasks might be subjective, annotators could be biased, or lack experience with the labeling tool or task.  Inherent uncertainty arises from limited observational information, such as diagnosing a medical condition using a single medical image in the skin classification application explored here \citep{LiuNATURE2020}. We ignore
other sources of uncertainty like task ambiguity \citep{UmaFAI2022} that arises for example when individual classes are not clearly agreed upon \citep{Phene2019,Medeiros2023}. 

Ground truth uncertainty is observed most prominently through \textit{annotator disagreement}. While inter-annotator disagreement can be measured through disagreement metrics, attributing it to annotation or inherent uncertainty remains challenging \citep{AbercrombieARXIV2023,RottgerNAACL2022}. Annotation uncertainty can potentially be reduced by expanding annotator groups, improving training, or refining labeling tools. In contrast, inherent uncertainty is generally irresolvable \citep{SchaekermannSIGCHIWORK2016}, even with costly adjudication protocols \citep{Schaekermann2019,Duggan2021}. Moreover, simple aggregations of opinions, such as majority voting, may eliminate minority views \citep{FieldACL2021}.

\begin{figure*}
    \centering
    \includegraphics[width=0.85\textwidth]{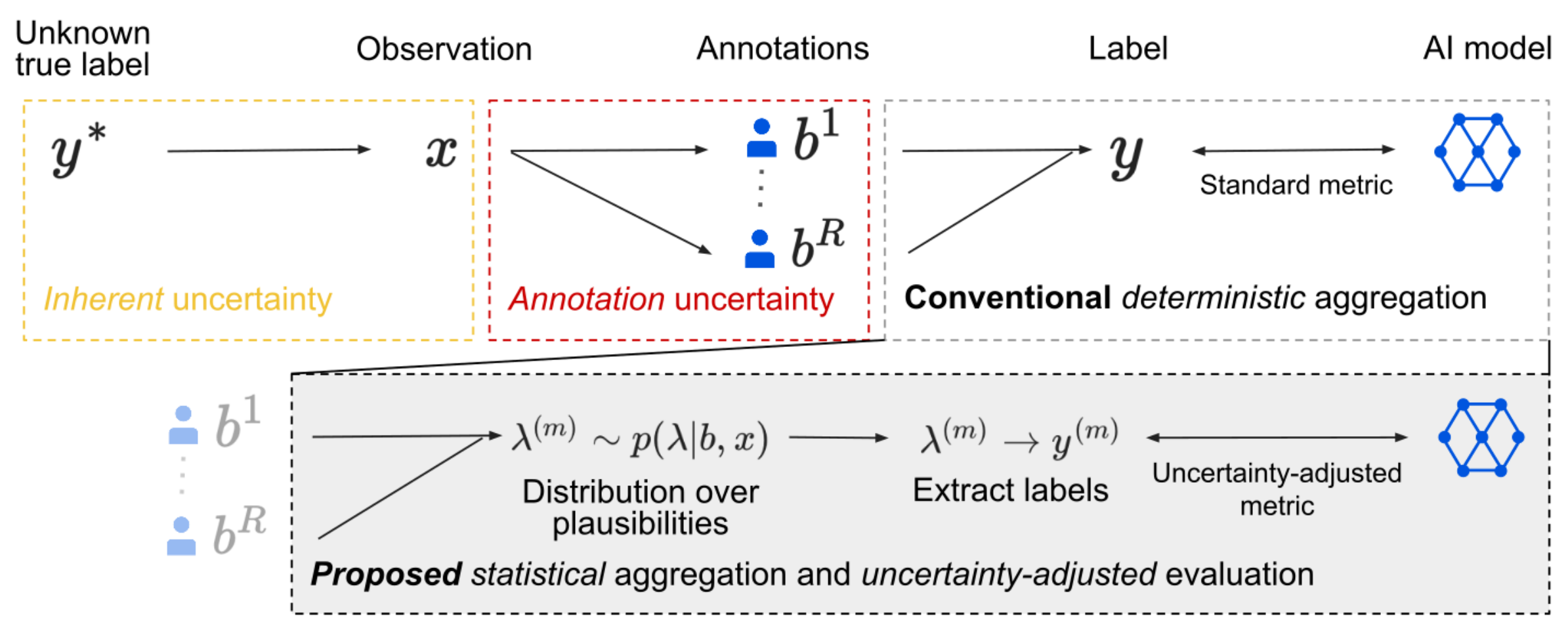}
    \vspace*{-8px}
    \caption{Ground truth uncertainty in supervised AI systems such as classification. The unobserved ground truth $y^*$ is assumed to yield an observation $x$. Here, $y^*$ could be the true underlying medical condition and the observation $x$ can be a medical image. Typically, $y^*$ is not uniquely identifiable from $x$ due to \textit{inherent uncertainty}. 
    To get an estimate of $y^*$, experts are asked to provide their annotations $b$ about a case. The experts 
    can have different opinions how to interpret a case; therefore, to deal with \textit{annotation uncertainty} instead of a single expert, multiple experts $R$ are consulted to give annotations $b^1,\dots,b^R$ based on $x$. These annotations are then typically (deterministically) aggregated and a single estimate $y$ of the true label $y^*$ is extracted and evaluation proceeds, ignoring annotation or inherent uncertainty. See Figure \ref{fig:results-motivation} for a concrete example in skin condition classification.
    In contrast, we use statistical aggregation to explicitly model distributions over probabilities of ground truth labels, that we call {\it plausibilities} $\lambda$. The distribution over $\lambda$ captures annotation uncertainty; and a single plausibility vector $\lambda$ captures the inherent uncertainty of a case. We discuss how to evaluate AI systems using this modeling approach.
    }
    \label{fig:introduction-disagreement}
\end{figure*}

While annotator disagreement can be observed in many standard datasets \citep{Krizhevsky2009,PetersonICCV2019}, it is particularly common in medicine. For example, in skin condition classification \citep{JainJAMA2021,EngBJD2019}, a significant portion of examples is typically subject to ground truth uncertainty  \citep{SchaekermannSIGCHIWORK2016}.
As a result, the problem of ground truth uncertainty has been recognized in several previous works; see e.g. \citep{SculleySDM2007,CabitzaAS2020,NorthcuttNIPSWORK2021,UmaJAIR2021,GordonCHI2021,DavaniTACL2022,PlankARXIV2022,LeonardelliARXIV2023}. Often, however, the focus is on mitigating the symptoms of ground truth uncertainty rather than tackling it directly. For example, there is a large body of work on dealing with label noise \citep{NorthcuttNIPSWORK2021}. In cases where ground truth uncertainty is modeled explicitly, related work focuses on training \citep{WelinderNIPS2010,RodriguesAAAI2018,GuanAAAI2018} but still assumes ground truth labels for evaluation. These shortcomings have recently been highlighted in \citep{MaierheinNATURE2018,GordonCHI2021}, but remain unaddressed.

\subsection{Evaluation under ground truth uncertainty}

To address evaluation under ground truth uncertainty, we \red{propose a framework that considers} a distribution over the ground truth rather than a single point estimate. This approach more faithfully captures the information communicated by annotators and is suitable for a broad spectrum of performance metrics beyond accuracy. Concretely, \red{as illustrated in Figure \ref{fig:introduction-disagreement}}, given multiple annotations $b^1,\ldots, b^R$ for example $x$, conventional evaluation obtains a single point estimate $y$ of the ground truth \red{through deterministic aggregation (top). Instead, we perform \textit{statistical aggregation} by Bayesian inference (bottom), using an algorithm that samples \textit{plausibilities} $\lambda$ - that is probability vectors over conditions - from a posterior distribution $p(\lambda|b,x)$.}
As plausibilities are probability vectors, even a single sample captures inherent uncertainty; for example, a posterior plausibility sample allocating high probability mass to several conditions indicates higher uncertainty about the underlying case. In contrast, annotation uncertainty is represented by variation in plausibility samples. When no variation exists in the plausibility samples, the only remaining source of uncertainty is inherent uncertainty. We control variance using a single annotator reliability parameter, which the user can choose. This choice could be informed by a domain expert who designed the rating task or tuned using rating data (similar to crowd sourcing work  \citep{YanML2014,ZhengPVLDB2017}). If selecting an appropriate rater reliability proves difficult, evaluating across a range of reliabilities is the best alternative."

\red{We propose measuring ground truth uncertainty and evaluating a trained model's performance using such plausibility samples from the above posterior in a Monte Carlo fashion. Specifically,} we define a case based measure, \textit{annotation certainty}, \red{as} the fraction of times that\red{, e.g., the top-1 label from plausibility samples agrees.}
This allows us to quantify annotation uncertainty on individual examples as well as on whole datasets. Moreover, we define \textit{uncertainty-adjusted} variants of common classification metrics such as top-$k$ accuracy or average overlap \cite{WuSAC2003,WebberTOIS2010} \red{in a similar fashion by averaging these metrics across plausibility samples. These metrics explicitly take ground truth uncertainty into account during evaluation. As a result, we avoid unfairly penalizing or rewarding the model's predictions if the ground truth is deemed uncertain.}

\subsection{Application to skin condition classification}

We apply our framework to skin condition classification from images in dermatology following the setting of \citep{LiuNATURE2020}. Annotations are expressed as differential diagnoses, which we formalize as \emph{partial} rankings that may exhibit ties and generally include only a small fraction of conditions. Due to the difficulty of diagnosis solely from images, we observe significant disparity among these differential diagnoses. In the literature, an established technique for aggregating these partial rankings is inverse rank normalization (IRN) \citep{LiuNATURE2020}. This 
technique provides a single point estimate of the probabilities of the underlying conditions.
Then, the condition attaining the highest IRN probability is chosen as the ground truth label. Evaluating e.g. the accuracy of an AI model in this way ignores any potential ground truth uncertainty. As a result, this evaluation is overconfident and ignores alternative scenarios where other conditions could have been more probable. This is particularly problematic in settings where AI models are meant to output multiple alternatives themselves (e.g., prediction sets or differential diagnosis).
Moreover, this approach is unable to capture the impact that individual annotations may have on metrics \red{or determine when improvements are statistically significant.}

As part of our framework, we propose two approaches for probabilistic aggregation of differential diagnoses expressed as partial rankings: First, we consider a re-interpretation of IRN as a point estimator of a categorical distribution. Based on this interpretation, we propose a Probabilistic IRN (PrIRN) where we model the posterior distribution over plausibilities as a Dirichlet distribution.
Second, we consider the Plackett--Luce (PL) model \citep{Plackett1975,Luce2012} which models probability distributions over full rankings through sampling without replacement. We adapt PL to the case of partial rankings, overcoming several technical challenges.
Based on these models, our analysis highlights that ignoring uncertainty leads to unrealistically confident and optimistic estimates of classifier performance. We observe that standard evaluation on curated labels that are derived from point estimates disregards large variations in performance, and hide the fact that results can be over-sensitive to the opinions of a single reader.
We discussed these results with dermatologists and highlight medical implications such as the inability to clearly categorize cases by risk. 

\subsection{Summary of contributions}

\red{
\begin{enumerate}
\item We address the problem of ground truth uncertainty in evaluating medical AI systems, comprising inherent uncertainty due to limited observational information and annotation uncertainty arising from finite and imperfect annotators.

\item We argue that conventional, deterministic approaches to aggregating annotations \emph{implicitly} ignore this uncertainty. We propose a Bayesian framework that \emph{explicitly} aggregates annotations into a posterior over \emph{plausibilities} (distributions over classes).

\item We introduce \emph{annotation certainty} to measure ground truth uncertainty on individual cases and entire datasets, and propose \emph{uncertainty-adjusted} variants of classification metrics to account for this uncertainty during evaluation.

\item We demonstrate our framework on a skin condition classification task where annotations are expressed as differential diagnoses and aggregated using either probabilistic IRN (PrIRN) or Plackett-Luce (PL), showing that conventional evaluation overestimates performance and disregards significant variation due to ground truth uncertainty.
\end{enumerate}
}
\section{Evaluation with uncertain ground truth from differential diagnoses}
\label{sec:methods-metrics}

This section introduces our framework for evaluation of AI systems with ground truth uncertainty. As this paper focuses on a case study in dermatology, we particularly concentrate on its application to the differential diagnosis annotations from prior work \citep{LiuNATURE2020}. We start by introducing notation and give an intuition of our approach on a toy example as well as a concrete example from our dermatology dataset. We then formalize this intuition and present two concrete statistical models we use for modeling and aggregating differential diagnosis annotations. Based on these models, we present measures for annotation uncertainty as well as uncertainty adjusted performance metrics for evaluating AI models.

\subsection{Notation and introductory examples}
\label{subsec:introduction-examples}

\begin{figure*}
    \centering
    \hspace*{-0.3cm}
    \includegraphics[height=2.75cm]{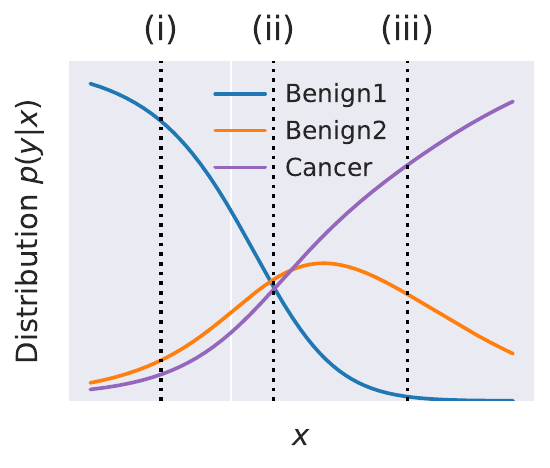}
    \includegraphics[height=2.75cm]{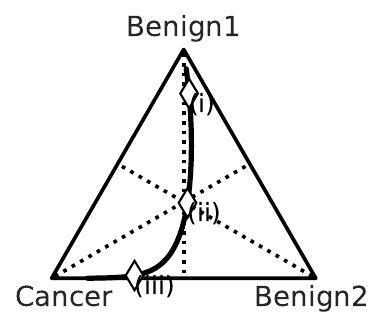}
    \includegraphics[height=2.75cm]{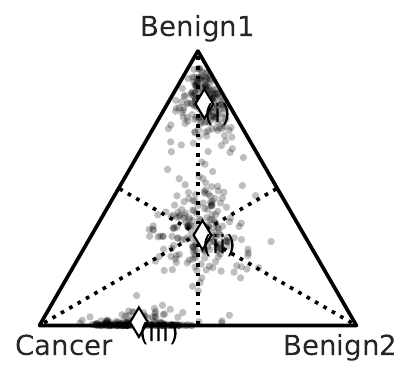}
    \includegraphics[height=2.75cm]{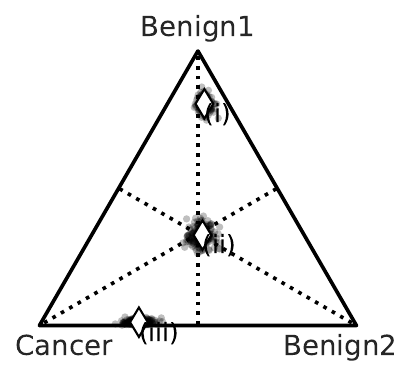}
    \includegraphics[height=2.75cm]{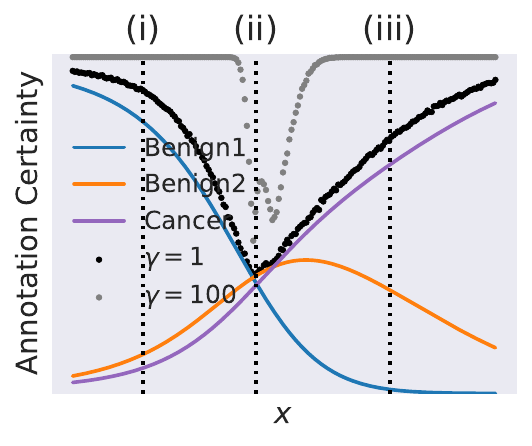}
    \vspace*{-6px}
    \caption{
    Illustration of annotation certainty on a toy dataset. Left: Synthetic dataset showing one-dimensional observations $x$ and corresponding true distribution $p(y|x)$ over labels $y$ for three imaginary labels Benign1, Benign2 and Cancer. Example (ii) is particularly ambiguous in the sense that $p(y|x)$ is not crisp. Middle left: For illustration, we plot the distributions $p(y|x)$ as ``diamonds'' on the 3-simplex. Middle and middle right: Modeling aggregation of annotator opinions $b$ statistically, see text, allows us to sample plausibilities $\lambda^{(m)}\sim p(\lambda|b, x)$. The spread of these plausibilities around the actual true distributions $p(y|x)$ captures the annotation uncertainty. This is influenced by the annotations as well as a prior \emph{reliability} parameter $\gamma$ \red{($\gamma = 1$ and $\gamma = 100$ in this case)}. This parameter reflects our prior trust in the annotators. Right: Measuring how often the top-1 label changes given plausibility samples $\lambda^{(m)}$, we can compute \emph{annotation certainty} for all examples. We plot annotation certainty across different reliabilities $\gamma$ (black to {\color{gray}gray}) indicating the effect of different prior trust levels.
    }
    \label{fig:introduction-ambiguity}
\end{figure*}

For illustration and introducing notation, we consider the synthetic toy dataset of Figure \ref{fig:introduction-ambiguity} (left). Here, observations $x$ are illustrated as one-dimensional on the x-axis and we plot the true distribution $p(y| x)$ for three different classes ({\color{blue}blue}, {\color{orange}orange} and {\color{violet}violet}) on the y-axis. \red{These could be different diagnoses, such as {\color{violet}cancer} and two benign conditions, {\color{blue}benign1}, {\color{orange}benign2}. For illustration, we pick} three examples, (i), (ii), and (iii). \red{These could correspond to patients to diagnose. Here,} the latter two examples, (ii) and (iii), are inherently ambiguous: the corresponding distribution over classes $p(y|x)$ is not crisp \red{as can be observed at the corresponding intersecting dotted line. For these cases, multiple classes have significant probability $p(y|x)$. For example, for example (ii), all three classes have roughly equal probability. We can visualize this more intuitively on a 3-simplex where the corners would correspond to crisp distributions (100\% probability for cancer, benign1 or benign3), see the diamonds in Figure \ref{fig:introduction-ambiguity} (middle left). Of course, in practice, we never observe the true distributions $p(y|x)$ for these examples (represented as diamonds on the graph).}

To address this, we assume access to a finite set of annotations $b$ \red{obtained from human expert raters}. For simplicity, in this example we assume these annotations to be single labels, \red{i.e., we assume that each rater samples a single label from the true $p(y|x)$ (or what they believe to be the true distribution)}. Then, we aggregate these opinions to obtain an approximation $\hat{\lambda}$ of $\lambda := p(y|x)$. \red{This can be achieved by counting frequencies. A ground truth label is then selected using}, e.g., majority voting. This represents how labels for many common benchmarks such as CIFAR and ImageNet \citep{RussakovskyIJCV2015}
have been obtained and this process is illustrated in Figure \ref{fig:introduction-disagreement} (top). Here, we refer to $\hat{\lambda}$ as \emph{plausibilities} as they construct a \red{categorical} distribution denoted $p(y|\hat{\lambda})$ over labels from which the $\arg\max$ corresponds to the majority voted label. \red{Unfortunately, this approach ignores any uncertainty present in the annotations since the rich information in these plausibilities is discarded in favor of a single, majority-voted label.}

\red{Instead, we intend to work directly with these plausibilities. To this end,} we assume a distribution $p(\lambda|b, x)$ over plausibilities. \red{For the toy example in Figure \ref{fig:introduction-ambiguity},} we use a Dirichlet distribution with concentration parameters reflecting the annotator opinions as well as a prior \emph{reliability} parameter ($\gamma$ in Figure \ref{fig:introduction-ambiguity}). This reliability parameter will quantify our \red{a priori} trust in the annotators. For example, we expect reliability to increase with the number of annotators or their expertise and training. For now, we consider reliability to be a variable parameter that affords us various perspectives on the data.
This is illustrated in Figure \ref{fig:introduction-ambiguity} (middle and middle right), showing plausibilities $\lambda^{(m)} \sim p(\lambda|b,x)$, $m \in [M]$, sampled from the statistical aggregation model on the 3-simplex.
The spread in these plausibilities represents annotation uncertainty and can be reduced using a higher reliability. The position on the simplex represents inherent uncertainty which is lower for plausibilities close to the corners such as (i) and higher for plausibilities close to the center such as (ii).
Then, we compute the top-1 label for every sampled $\lambda^{(m)}$, i.e., the label with the largest plausibility $\arg\max_{k \in [K]} \lambda^{(m)}_k$. This allows us to measure \emph{annotation certainty} (defined formally in Section \ref{subsec:methods-certainty}) -- if the top-1 label is always the same label, there is no annotation uncertainty. Specifically we can measure the fraction of plausibilities $\lambda^{(m)}$ among all $M$ samples where the top-1 label is $k$. Taking the maximum over all labels $k \in [K]$ defines \emph{annotation certainty}. In Figure \ref{fig:introduction-ambiguity} \red{(right)} we plot this annotation certainty for different reliabilities (black and {\color{gray}gray}) and we clearly see that annotation certainty is consistently low for example (ii), while annotation certainty increases with higher reliability for example (iii). This follows our approach as depicted in Figure \ref{fig:introduction-disagreement} (bottom) and allows us to easily summarize annotation uncertainty across the whole dataset.

\begin{figure*}
    \centering
    \begin{minipage}[t]{0.15\textwidth}
        \vspace*{0px}
        \centering
        \normalsize
        \textbf{Input $x$}
        \vskip 4px
        
        \includegraphics[width=1\textwidth]{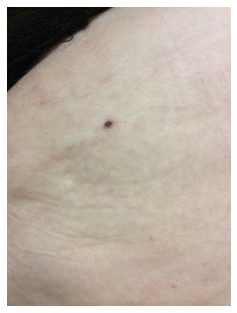}
    \end{minipage}
    \hspace*{0.5cm}
    \begin{minipage}[t]{0.65\textwidth}
        \vspace*{0px}
        \normalsize
        \textbf{Annotations $b$}
        \vskip 4px
        
        A0: \{Pyogenic granuloma (\textcolor{green!50!black}{Low})\} \{\underline{Hemangioma} (\textcolor{yellow!85!black}{Med})\} \{\textit{Melanoma} (\textcolor{red}{High})\}\\
        A1: \{Angiokeratoma of skin (\textcolor{green!50!black}{Low})\} \{Atypical Nevus (\textcolor{yellow!85!black}{Med})\}\\
        A2: \{\underline{Hemangioma} (\textcolor{yellow!85!black}{Med})\} \{Melanocytic Nevus (\textcolor{green!50!black}{Low}), \textit{Melanoma} (\textcolor{red}{High}),\\
        O/E - ecchymoses present (\textcolor{green!50!black}{Low})\}\\
        A3: \{\underline{Hemangioma} (\textcolor{yellow!85!black}{Med}), \textit{Melanoma} (\textcolor{red}{High}), Skin Tag (\textcolor{green!50!black}{Low}=\}\\
        A4: \{\textit{Melanoma} (\textcolor{red}{High})\}\\
        A5: \{\underline{Hemangioma} (\textcolor{yellow!85!black}{Med})\} \{\textit{Melanoma} (\textcolor{red}{High})\} \{Melanocytic Nevus (\textcolor{green!50!black}{Low})\}
    \end{minipage}
    \space*{-4px}
    \caption{An example input and corresponding expert dermatologist annotations from the dataset introduced in \citep{LiuNATURE2020}. We highlight the conditions' risk categorizations in parentheses \red{and indicate ranks using curly brackets where tied conditions are grouped together. We further} underline the majority voted label used for evaluation in \citep{LiuNATURE2020}, ``Hemangioma''. For contrast, we mark an alternative, high-risk condition, ``Melanoma'', in italics. There is clearly significant disagreement among experts, highlighting high ground truth uncertainty which could stem from either inherent or annotation uncertainty.}
    \label{fig:introduction-derm}
\end{figure*}

The main challenge in applying the aforementioned approach to our dermatology case study is that the expert annotations collected are not single labels per example. As illustrated in Figure \ref{fig:introduction-derm}, the annotations collected in \citep{LiuNATURE2020} come in the form of differential diagnoses. Mathematically, differential diagnoses can be formalized as \emph{partial} rankings, including a limited number of possible conditions that are ranked while allowing for ties (see Section \ref{subsec:method-diff}). This has to be accounted for in the aggregation model $p(\lambda|b,x)$, which will get more complex compared to the simple Dirichlet model above. The example in Figure \ref{fig:introduction-derm} highlights this difficulty on a concrete example with high level of disagreement among expert annotators. Here, the majority-voted label (used for evaluation in \citep{LiuNATURE2020}; see Section \ref{subsec:method-diff} for formal definition) is ``Hemangioma'', a medium-risk condition. However, this ignores a higher-risk, cancerous condition, ``Melanoma'' that is also present in many of the provided differential diagnoses. Clearly, limiting evaluation to ``Hemangioma'' is overly optimistic and disregards the rich information contained in the annotations. In our framework, we expect the top-1 label to switch frequently between ``Hemangioma'' and ``Melanoma'' for different samples $\lambda^{(m)} \sim p(\lambda|b,x)$, indicating high uncertainty. Adjusting $p(\lambda|b,x)$ to different settings makes this approach very flexible, see \ref{sec:app-introduction-examples} for further examples.

\subsection{Statistical model}
\label{subsec:methods-model}

The statistical model $p(\lambda|b,x)$ informally introduced above and summarized in
Figure \ref{fig:methods-model} is at the core of our framework. Essentially, we propose to replace deterministic aggregation of annotator opinions with a statistical model. To this end, we view aggregation as computing the posterior distribution $p(\plaus{}{}|b, x)$ over plausibilities, given annotations $b$ and observations $x$. Then, plausibilities represent distributions over labels:
\begin{align}
    p(y|b,x)=\int p(y|\lambda)p(\lambda|b,x) \mathrm{d}\lambda.
\end{align}
 In all the examples discussed within this paper, we make the simplifying assumption $p(\plaus{}{}|b, x) = p(\plaus{}{}|b)$ such that plausibilities depend only from the annotator opinions. Computing the posterior distribution $p(\plaus{}{}|b)$ is a central task for our purposes. We compute this posterior by specifying the annotation process, i.e., $p(b|\plaus{}{})$ and assuming a prior $p(\lambda)$ independent of the input $x$. The annotation process specifies how we expect experts to provide their annotations if we knew the underlying distribution $ \plaus{}{}$ over labels.
Additionally, we assume a \emph{reliability} parameter as part of our statistical model. In practice, this often corresponds to a temperature parameter in $p(b|\plaus{}{})$ (e.g., $\gamma$ in our toy example from Figure \ref{fig:introduction-ambiguity}). However, we interpret it as quantifying the prior trust we put in the annotators. As fixing this parameter based on domain expertise or data is challenging, we treat it as a free parameter that is to be explored during evaluation.

In the toy example of Section \ref{subsec:introduction-examples}, we assumed \red{that each annotator provides the top-1 label}. Thus, it was simple to derive the posterior in closed form for a Dirichlet prior distribution. Here, the plausibilities $\plaus{}{}$ explicitly correspond to the categorical distribution over classes $p(y|x)$.
In other cases, where annotations do not directly match the label space (see \ref{sec:app-introduction-examples} for examples), this statistical model might be more complex.
For our case study, a skin-condition classification problem detailed in Section \ref{sec:derm}, the expert annotations are partial rankings instead of top-1 labels. In this case, the plausibilities $\lambda$ still approximate the categorical distribution $p(y|x)$ but we need to rely on more complex models for $p(b|\plaus{}{})$. 
This might entail that that the posterior  $p(\plaus{}{}|b)$ is not available in closed-form, requiring specialized techniques for sampling.

\begin{figure}
    \centering
    \begin{tikzpicture}
        \node[circle,draw=black] (lambda) at (0,0){$\plaus{}{}$};
        \node[circle,draw=black] (y) at (-0.5,-1.5){${y}$};
        \node[circle,draw=black, double] (h) at (0.5,-1.5){$b^1$};
        \node[circle,draw=black, double] (h2) at (2.0,-1.5){$b^r$};
        \node[] (h3) at (1.25,-1.5){$\dots$};
        \node[circle,draw=black, double] (x) at (0,1.5){$x$};
        \draw[->] (lambda) -- (y);
        \draw[->] (lambda) -- (h);
        \draw[->] (lambda) -- (h2);
        \draw[->] (x) -- (lambda);
    \end{tikzpicture}
    \caption{Graphical model corresponding to the statistical aggregation model at the core of our framework: This describes the joint distribution of the observation $x$, plausibilities $\lambda$, annotations $b=(b^1,...,b^R)$ and label $y$. We observe $x$ and $b$ from which one can infer plausibilities $p(\lambda|b, x)$ and label $p(y|\lambda)$, see text. As discussed in the text, in order to simplify inference of $\lambda$, we assume a simplified model with $p(\lambda|b,x) = p(\lambda|b)$.}
     \label{fig:methods-model}
\end{figure}

Beyond the specific choice of $p(b|\plaus{}{})$, we assume the annotators to be conditionally independent given the plausibilities. This is a simplification and there is significant work in crowd sourcing and truth discovery considering alternative models \citep{YanML2014}. Moreover, in our experiments, we consider a simplified version of the model in Figure \ref{fig:methods-model} that assumes conditional independence $p(\plaus{}{}|b,x)=p(\plaus{}{}|b)$. This makes inferring the posterior over $\plaus{}{}$ easier but clearly reduces our ability to model input dependent uncertainty and thereby disentangle the different sources of uncertainty.
For example, it is difficult to distinguish between an inherently ambiguous example where we observed low disagreement by chance and an actually unambiguous example with low disagreement.
However, this distinction would be extremely valuable, e.g., to inform relabeling.

One might argue that we introduce a model assumption into evaluation and there is no guarantee that $p(y|b,x)$ converges to the true $p(y|x)$ as the number of annotators goes to infinity. However, it is important to realize that almost all existing benchmarks are based on an assumed annotation model. This is because these benchmarks obtain ground truth labels through expert annotations. That the assumed annotation model is implicit and often unacknowledged in the corresponding publications and leaderboards does not change the fact that it has tangible effects on evaluation. We make this model assumption \red{explicit} and actually use it for our advantage: quantifying the underlying ground truth uncertainty.

\subsection{Model for differential diagnosis annotations}
\label{subsec:method-diff}

To apply this framework to dermatology, we model differential diagnoses as partial rankings and discuss meaningful aggregation models $p(\lambda|b,x)$ that facilitate sampling plausibilities. Specifically, we follow \citep{LiuNATURE2020}, and use inverse rank normalization (IRN) as a deterministic aggregation baseline. We then propose a probabilistic variant of IRN as \red{a} first, simple statistical aggregation model before considering a more sophisticated Plackett--Luce model \citep{Plackett1975,Luce2012}.

For each case, each annotator selects a subset of conditions they think might be the true diagnosis. In addition, they assign a confidence score to each condition, indicating how likely they think that condition is the true diagnosis. The subsets of conditions selected by different annotators may differ in size. Moreover, the confidence scores are usually not comparable across annotators which is why they are instead used to produce a ranking of the conditions for each annotator \citep{LiuNATURE2020}. This results in \textit{partial} rankings because there may be ties among conditions (i.e., conditions with the same confidence score) and the large majority of possible conditions remains unranked. Formally, we write
\begin{align}
    \block{}{}{} = (\block{}{}{1} \succ \block{}{}{2} \ldots \succ \block{}{}{L}).\label{eq:methods-blocks}
\end{align}
We call groups of tied conditions ``blocks'', and each block contains one or more labels, $\block{}{}{i} = \{\choice{}{i}{1},\ldots,\choice{}{i}{|\block{}{}{i}|}\}$ with $\choice{}{i}{j} \in [K]$ and $K$ being the number of classes. Here, $[P]:=\{1,...,P\}$ for integer $P$ and $\block{}{}{i}\succ\block{}{}{j}$ indicates that classes in the $i$-th block are ranked higher than those in the $j$-th block (according to the confidence scores). We assume the blocks are non-empty, mutually exclusive, i.e., $\emptyset \neq \block{}{}{i} \subseteq [K]$  and  $\block{}{}{i} \cap \block{}{}{j} = \emptyset$, and that $\bigcup_{i=1}^{L} b_{i} = [K]$. Note that there might exist multiple equivalent permutations within the individual blocks $b_i$ such that we use $\mathcal{S}(\block{}{}{})$ to denote the set of all permutations compatible with Equation \eqref{eq:methods-blocks}. In the literature discussing ranking metrics and aggregation \citep{Sakai2013,WebberTOIS2010,SculleySDM2007,WuSAC2003}, partial rankings are also commonly referred to as top-$k$ rankings with ties. While some metrics such as Kendall's tau and Spearman's footrule have been adapted to ties \citep{FaginPODS2004,VignaWWW2015}, modeling partial rankings is generally non-trivial. This also holds for any aggregation model $p(\lambda|b,x)$ as we will see in Section \ref{subsec:methods-pl}.

As an example, we recall annotator 2 from Figure \ref{fig:introduction-derm} where the blocks are defined as
\begin{align*}
    b = (&b_1 = \{\textit{Hemangioma}\}, \\ &b_2 = \{\textit{Melanocytic Nevus}, \textit{Melanoma}, \textit{O/E}\}).
\end{align*}
We interpret the above example as follows: the annotator declares that \textit{Hemangioma} is their first choice, followed by either \textit{Melanocytic Nevus}, \textit{Melanoma}, or \textit{O/E}. However, for the latter three conditions, the annotator is unable to say which one is more likely than the others. That is, these three conditions are tied for the second rank. Implicitly, this means that all $3! = 6$ possible permutations of the second rank would be equivalent. Moreover, annotators usually ignore a large set of conditions. Strictly speaking, the ignored conditions can be thought of as another block
\begin{align*}
    b_3 = \{\text{any other condition}\}
\end{align*}
with the last block $b_L = b_3$ capturing all unranked classes.

\subsubsection{Inverse rank normalized (IRN) aggregation}
\label{subsec:methods-irn}

A heuristic method for aggregating differential diagnoses to arrive at a plausibility point estimate is \textbf{Inverse Rank Normalization (IRN)} \citep{LiuNATURE2020}. Given a partial ranking $\block{}{}{}$ from a single annotator with $L$ blocks as in Equation \eqref{eq:methods-blocks}, we define the unnormalized IRN score of a condition $y$ as 
\begin{align}
    \overline{\text{IRN}}(y; \block{}{}{}) 
    =  \sum_{i = 1}^{L-1} \frac{\nicefrac{1}{i}}{|\block{}{i}{}|} \delta[y \in \block{}{i}{}].\label{eq:methods-irn-singleton}
\end{align}
\red{where $\delta$ is an indicator function where $\delta[S]$ is one when the statement $S$ is true and zero otherwise.}
Unnormalized IRN assigns a weight $\nicefrac{1}{i}$ to each block $\block{}{i}{}$ for $i \in [L-1]$ and a weight of $0$ for all the classes in the last block $\block{}{}{L}$ of unranked classes. The weight is distributed equally across all classes in a block, i.e., class $\choice{}{i}{j}$ obtains weight $\nicefrac{{1/i}}{|\block{}{i}{}|}$. 
Given $R$ partial rankings $\block{}{}{}^1, \ldots, \block{}{}{}^R$ from $R$ annotators, we define the unnormalized IRN score of a condition $y$ as 
\begin{align}
    \overline{\text{IRN}}(y; \block{}{}{}^1,\ldots,\block{}{}{}^R) 
    = \sum_{r = 1}^R \overline{\text{IRN}}(y; \block{}{}{}^r) \label{eq:methods-irn}
\end{align}
with $L_r$ being the number of blocks for annotator $r$ and $b_j^r$ indexing these blocks.
Normalized IRN is 
\begin{align}
    \text{IRN}(y; \block{}{}{}^1,\ldots,\block{}{}{}^R) =\frac{  \overline{\text{IRN}}(y; \block{}{}{}^1,\ldots,\block{}{}{}^R)} { \sum_{k = 1}^K \overline{\text{IRN}}(k; \block{}{}{}^1,\ldots,\block{}{}{}^R)}.
\end{align}
Note that normalization comes after aggregation rather than normalizing per annotator $r$ first and then averaging.

Overall, IRN can be viewed as an adhoc point estimate of the plausibilities $\lambda$ without an explicit generative model for $p(\lambda|b,x)$. This also implies that we cannot sample plausibilities from $p(\lambda|b,x)$.
\red{Moreover, because it is deterministic, it also has to resolve ties deterministically. However, distributing a rank's weight equally across tied conditions can lead to unintuitive plausibilities (see Figure \ref{fig:results-motivation} for an example and discussion).}
In contrast, the next section describes a probabilistic version of IRN that allows to sample plausibilities.

\subsubsection{Probabilistic IRN (\PIRN) aggregation}
\label{subsec:method-birn}

To derive a \textbf{probabilistic variant of IRN (\PIRN)}, we interpret the calculation of the unnormalized IRN as a counting method for accumulating the statistics of a histogram on discrete data. Then, the normalized $\text{IRN}(y; \block{}{}{}^1,\ldots,\block{}{}{}^R)$ corresponds to a maximum likelihood estimate of the parameters of a multinomial distribution (the sampling distribution of a histogram) given the accumulated statistics. In this spirit, we define the sampling distribution $p(\lambda| \block{}{}{}^1,\ldots,\block{}{}{}^R)$ directly as a Dirichlet distribution $\mathcal{D}(\gamma \hat{\lambda})$ where $\hat{\lambda}_y = \text{IRN}(y; \block{}{}{}^1,\ldots,\block{}{}{}^R)$ and $\gamma>0$ (the concentration parameter) represents our reliability parameter. Here, the $\mathcal{D}(\gamma \hat{\lambda})$ distribution is the true posterior distribution of the parameters of a categorical distribution if labels were drawn i.i.d. and the observed counts would be $\gamma \hat{\lambda}$. 
Then, the reliability parameter $\gamma$ can be interpreted as the number of annotators: with increasing number of annotators we have less annotation uncertainty and for $\gamma \rightarrow \infty$ the Dirichlet distribution degenerates to a point mass on $\hat{\lambda}$ -- the IRN estimate.

\subsubsection{Plackett--Luce (PL) aggregation}
\label{subsec:methods-pl}

One conceptual problem of \PIRN is that its weighting and aggregation choices appear to be fairly arbitrary and the corresponding generative model for the annotation process based on a categorical distribution is not realistic for differential diagnoses. An alternative is defining a generative model, that takes the ordering into account: 
when raters provide a partially ordered list of conditions, they exclude already mentioned conditions when choosing the next option. 

The \textbf{Plackett--Luce (PL)} \citep{Plackett1975,Luce2012} model is a natural choice for this setting. The PL distribution models how annotators, given the true plausibilities $\lambda$, draw \emph{full} rankings by sampling without replacement from all conditions. This model is very appropriate for our problem. Unfortunately, in a Bayesian framework, the resulting posterior distribution of the plausibilities cannot be derived analytically. However, efficient Gibbs sampling algorithms \citep{Caron2010,CaronARXIV2012} and optimization procedures \citep{HunterAS2003} do exist. Here, we extend the Gibbs sampler of \citep{Caron2010} to \emph{partial} rankings. 

In the following, we assume a single case for brevity, with the annotator being indexed by $r \in [R]$. The inferential procedure we describe can be repeated for any other case independently. We also assume that
\begin{align}
\label{eq:pl_likelihood_rankings}
p(\choice{}{}{}|\lambda) = \prod_{r = 1}^R p(\choice{r}{}{}|\lambda),
\end{align}
meaning that the annotators' rankings are independent, conditioned on plausibilities.
Then, moving to the description of $p(\choice{r}{}{}|\lambda)$, the classical PL model makes the assumption that each annotator $r$ reports a \emph{full} ranking of classes by sampling without replacement, proportional to the plausibilities $\plaus{}{}$. So the first class $\choice{r}{}{1}$ is drawn with probability
\begin{align*}
p(\choice{r}{}{1} = k) = \frac{\plaus{}{k}}{Z}
\end{align*}
where the normalization constant is
$Z = \sum_{k'} \plaus{}{k'}$. Then the annotator draws the next choice 
$\choice{r}{}{2}$ (by excluding the previous choice $\choice{r}{}{1}$) from $\{1\dots K\} \setminus \{\choice{r}{}{1}\}$ with probability
\begin{align*}
p(\choice{r}{}{2}  = k | \choice{r}{}{1}) = \frac{ \plaus{}{k} }{Z - \plaus{}{\choice{r}{}{1}}}
\end{align*}
and so on until no choices are left. Then, the log-likelihood of the plausibility vector $\plaus{}{}$ given the choices $\choice{r}{}{}$ is given as
\begin{align*}
\mathcal{L}_r(\plaus{}{}) = \log \plaus{}{\choice{r}{}{1}} - \log Z + &\log \plaus{}{\choice{r}{}{2}} - \log(Z - \plaus{}{\choice{r}{}{1}}) + \dots \\ &+ \log \plaus{}{\choice{r}{}{\numcond}}
- \log\left(Z - \sum_{k=1}^{\numcond-1} \plaus{}{\choice{r}{}{k}}\right).
\end{align*}
Given Equation \eqref{eq:pl_likelihood_rankings}, the log-likelihood for the entire parameter vector $\plaus{}{}$ is simply the equal contributions of all the annotators that have rated the case $\mathcal{L}(\plaus{}{}) = \sum_r \mathcal{L}_r(\plaus{}{})$. 

The inferential goal of interest is estimating the latent plausibilities $\plaus{}{}$ based on the annotations $b$ in the form of full rankings $\sigma$. This can be achieved by maximizing the likelihood $\mathcal{L}(\plaus{}{})$, or in a Bayesian way by assigning a prior distribution $p(\plaus{}{})$ and sampling from the posterior distribution $p(\plaus{}{}| \choice{}{}{} ) \propto p(\plaus{}{}) \prod_r p(\choice{r}{}{}| \plaus{}{})$. As the maximum likelihood solution would provide a point estimate of the plausibilities, similar to IRN, here we focus on the sampling approach.

We introduce the equivalent of a reliability parameter $\gamma$ for the PL distribution by defining a new tempered PL distribution with parameters $\bar{\lambda} = \lambda^{\gamma}$
and normalization $\bar{Z} = \sum_{k} \bar{\lambda}_{k} = \sum_{k} \lambda^{\gamma}_{k}$. Unfortunately, the resulting tempered model no longer enjoys the conjugacy properties needed for deriving a Gibbs sampling algorithm. While it is possible to use more generic sampling methods such as Hamiltonian Monte Carlo (HMC) for this tempered PL model, qualitatively we obtain the same concentration effect by simply repeating the partial rankings for each annotator such that the likelihood contribution of each annotator to the posterior is $\gamma \mathcal{L}_r(\plaus{}{})$ where $\gamma$ is a small integer. This also matches the intuition of the reliability parameter $\gamma$ in \PIRN~from the previous section.

The PL model as introduced above defines a distribution over full rankings. While it is straightforward to compute the probability of a given full ranking under the PL distribution, partial rankings -- especially with larger block sizes -- provide a computational challenge as we need to consider all the permutations compatible with the partial rankings. This would amount to a computational complexity that scales factorially in the block size. In contrast, in \ref{app:pl_exact}, we describe a procedure for computing the exact likelihood for partial rankings that scales only as $O(2^{|B|})$ with the block sizes $|B|$ making it practical for typical differential diagnoses.

\subsection{Annotation certainty}
\label{subsec:methods-certainty}

Having formalized two aggregation models $p(\lambda|b)$ to use for differential diagnosis annotations, we now formalize our measure of annotation uncertainty.
We first consider a label $y \in [K]$, and define the certainty for a specific label $y$ as the probability that $y$ corresponds to the top-1 label of the plausibilities $\lambda$, given the input $x$ and annotations $b$ under the chosen statistical model for aggregation. More formally, we can write this as an expectation under $p(\lambda|b,x)$:
\begin{align}
    \text{Certainty}(y; b, x) &= p(y = \arg \max_{j} \plaus{}{j}) \label{eq:definition-certainty-singleton} \\ &=  \E_{p(\plaus{}{}| b, x)} \Big\{\ind{y = \arg \max_{j} \plaus{}{j}}\Big\}.\nonumber 
\end{align}
In practice, we compute this expectation using a Monte Carlo average
\begin{align}
    \text{Certainty}(y; b, x) \approx \frac{1}{M} \sum_{m = 1}^M &\ind{y = \arg \max_{j} \lambda_j^{(m)}},  \label{eq:definition-certainty}\\ &\text{where~}\plaus{(m)}{} \overset{\textup{i.i.d.}}{\sim} p(\lambda|b, x). \nonumber
\end{align}
This estimates certainty for a specified label $y$ for a specific example. To summarize the overall annotation certainty for the example, we compute the maximum certainty across all possible labels $y$:
\begin{align}
    \text{AnnotationCertainty}_1(b, x) = \max_{y\in [K]} \text{Certainty}(y; b, x). \label{eq:definition-annotation-certainty-singleton}
\end{align}
Here, the subscript $1$ indicates that annotation certainty refers to the top-1 label from plausibilities. 

An important feature of the above definition is that we can generalize it to sets of labels $C$, e.g., top-$j$ sets with $|C| = j$:
\begin{align}
    \text{AnnotationCertainty}_j(b, x) = \max_{C:\text{ top-}j\text{ labels}} \text{Certainty}(C; b, x)  \label{eq:definition-annotation-certainty-topset}\\
    \text{with}\quad\text{Certainty}(C; b, x) = \E_{p(\plaus{}{}| b, x)} \Big\{\ind{C = Y_{\text{top}-|C|}(\lambda)}\Big\},
\end{align}
where $Y_{\text{top}-|C|}(\lambda)$ refers to the set that includes $C$ most plausible labels based on $\lambda$. This certainty measure can also be estimated using $M$ Monte Carlo samples. This prevents us from having to enumerate all $K!/(j!(K-j)!)$ subsets of size $j$ as seemingly implied by (\ref{eq:definition-annotation-certainty-topset}), which would be prohibitive. Instead, we only need to consider top-$j$ sets corresponding to the $M$ samples. However, larger $j$ might require a higher $M$ for reliable estimates of annotation certainty. To measure annotation certainty of a dataset as a whole, we can average \text{AnnotationCertainty}($b, x$) across examples.

Note that this measure depends on the aggregation procedure $p(\lambda|b,x)$.
For example, Equation \eqref{eq:definition-annotation-certainty-topset} is always $1$ for any input where the model $p(\lambda|b,x)$ is a point mass; i.e., it gives a single deterministic point estimate $\hat{\lambda}$ (subject to there being ties, which we ignore for simplicity). This is, by construction the case in any deterministic aggregation procedure for human annotations.

To understand intuitively what the measures are calculating, considering the simplex in Figure \ref{fig:introduction-ambiguity}. Each point on this simplex corresponds to an individual plausibility vector. For a given case, if we use a deterministic aggregation methods, the plausibility vector is a single point estimate and the top-$1$ label is associated with one of the three sectors shown on the simplex. \red{The annotation certainty measures the impact of the spread in plausibilities on the top-$j$ labels}. Thereby, annotation certainty implicitly also captures inherent uncertainty: even a small spread in plausibilities changes the top-$j$ labels more often for inherently uncertain examples, i.e., plausibilities close to the center of the simplex.
More generally, this means that annotation certainty will be high for easy, unambiguous examples that all annotators agree on.
The reverse, however, is not necessarily true: low annotation certainty \emph{can} indicate high inherent uncertainty but it does not necessarily have to.
For example, many experienced annotators consistently disagreeing might indicate high inherent uncertainty; however, annotators might also disagree for other reasons such as unclear annotation instructions, even if the example is generally easy to classify.
This also explains our naming: annotation certainty explicitly measures annotation uncertainty and only implicitly accounts for inherent uncertainty.
Partly, this also stems from our assumption that $p(\lambda|b,x) = p(\lambda|b)$ in Section \ref{subsec:methods-model}, making it more difficult to disentangle annotation and inherent uncertainty.

\subsection{Uncertainty-adjusted accuracy}
\label{subsec:methods-measuring}

Given a measure of annotation certainty, we intend to take it into account when evaluating performance of AI models. Specifically, we assume the label set $C$ used in Equation (\ref{eq:definition-annotation-certainty-topset}) to be a prediction set from a classifier -- for example, corresponding to the top-$k$ logits of a deep neural networks (denoted $C_{\text{top}-k}$. Then, we wish to measure the quality of this prediction set.
If we knew the true plausibilities $\lambda^* = p(y|x)$, our ground truth target $Y_{\text{top}-j}(\lambda^*)$ would be chosen as the top-$j$ elements of $\lambda^*$.
The quality of the prediction can then be computed by evaluating the indicator of the event that the target set is contained within the prediction set, $\ind{Y_{\text{top}-j}(\lambda^*) \subseteq C_{\text{top}-k}(x)} \in \{0, 1\}$, assuming $j \leq k$ for simplicity). For example, the standard top-$k$ accuracy is an estimate of the probability where we specifically take $j=1$ and define
\begin{align}
\text{Accuracy}_{\text{top}-k} &= p(Y_{\text{top}-1}(\lambda^*) \subseteq C_{\text{top}-k}(x)) \\ &= \E_{p(\lambda^*, x)} \{\ind{Y_{\text{top}-1}(\lambda^*)\subseteq C_{\text{top}-k}(x)}\}.
\end{align}
However, we acknowledge that $\lambda^*$ is unknown and that there is uncertainty on $\lambda$. Given $p(\lambda|b,x)$, we can quantify this uncertainty using $\text{Certainty}$ from Equation \eqref{eq:definition-certainty}. Integrating this into the above definition of accuracy yields the proposed \emph{uncertainty-adjusted} version:
\begin{align}
    \text{UA-Accuracy}_{\text{top}-k} = p(Y_{\text{top-}1}(\lambda) \subseteq C_{\text{top-}k}(x))\\
    = \E_{p(b, x)} \E_{p(\lambda| b, x)}\{ \delta[Y_{\text{top-}1}(\lambda) \subseteq C_{\text{top-}k}(x)]\} \label{eq:methods-evaluation-ua-accuracy}.
\end{align}
Note that this metric now implicitly depends on the annotations $b$ through our statistical model $p(\lambda|b,x)$.
As with annotation certainty, this metric mainly accounts for annotation uncertainty and is only indirectly aware of inherent uncertainty (see Section \ref{subsec:methods-certainty}). In order to explicitly account for inherent uncertainty, i.e., the plausibilities $\lambda$ not having a clear top-$1$ label, we can compare the top-$j$ target set $Y_{\text{top-}j}(\lambda)$ with the top-$k$ prediction set $C_{\text{top-}k}(x)$ set. However, to avoid extensive notation, we consider only target sets having the same cardinality as the prediction set. We call this metric \emph{uncertainty-adjusted top-$k$ set accuracy}: 
\begin{align}
    \text{UA-SetAccuracy} = p(Y_{\text{top}-|C|}(\lambda)  = C(x))\\
    = \E_{p(b, x)} \E_{p(\plaus{}{}| b, x)} \{\ind{Y_{\text{top}-|C|}(\lambda)  = C(x)}\}.
    \label{eq:methods-evaluation-ua-set-accuracy}
\end{align}
This deviates slightly from the notion of standard top-$k$ (vs. top-$1$) accuracy, but can be more appropriate when evaluating under uncertain ground truth where a large part of the uncertainty stems from inherent uncertainty. In such cases, the top-$1$ label will often be a poor approximation of the ground truth. In practice, we approximate these uncertainty-adjusted metrics using a Monte Carlo estimate.

Uncertainty-adjusted accuracy reduces to standard accuracy as soon as there is no annotation uncertainty or it is ignored through deterministic aggregation (i.e., when a point estimate $\hat{\lambda}$ is used). As with annotation certainty, uncertainty-adjusted metrics primarily capture annotation uncertainty and do not necessarily capture inherent uncertainty if it is not reflected in the annotations. This is particularly important to highlight for uncertainty-adjusted top-$k$ accuracy, which only considers the top-1 label from sampled plausibilities $\lambda^{(m)}$. For very ambiguous examples, this can ignore inherent uncertainty if the annotation uncertainty is low such that most plausibilities $\lambda^{(m)}$ agree on the top-1 label. To some extent this is mitigated by using our uncertainty-adjusted set accuracy. However, as our approach of constructing uncertainty-adjusted metrics is very general, this can also be addressed by considering different ``base'' metrics.

\begin{figure*}[t]
    \centering
    \begin{minipage}{0.05\textwidth}
        \rotatebox{90}{Plausibilities}
    \end{minipage}
    \begin{minipage}{0.93\textwidth}
        \small\hspace{4cm}PL plausibilities\hspace{6cm} PRIRN plausibilities
        
        \includegraphics[trim={0 0 0 0.65cm},clip,width=0.49\textwidth]{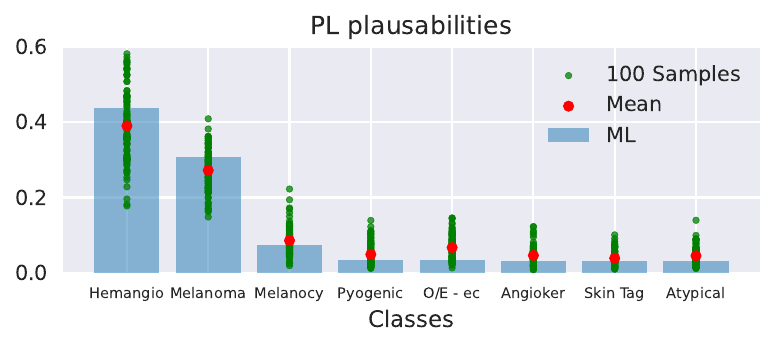}
        \includegraphics[trim={0 0 0 0.65cm},clip,width=0.49\textwidth]{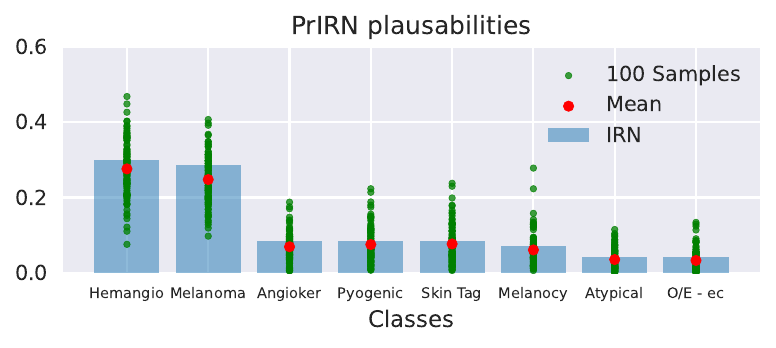}
    \end{minipage}
    
    \vskip 3px
    {\color{black!50!white}\rule{\textwidth}{1pt}}
    \vskip 3px
    
    \begin{minipage}{0.05\textwidth}
        \rotatebox{90}{Pred.}
    \end{minipage}
    \begin{minipage}{0.93\textwidth}
        \begin{minipage}{0.49\textwidth}
            \small
            Model A: $C_{\text{top-}3} =$\\ \{Atyp. Nevus, Hemangioma, Melanocytic Nevus\}
        \end{minipage}
        \hfill
        \begin{minipage}{0.49\textwidth}
            \small
            Model B: $C_{\text{top-}3} =$\\ \{Hemangioma, Melanocytic Nevus, Melanoma\}
        \end{minipage}
    \end{minipage}
    
    \vskip 3px
    {\color{black!50!white}\rule{\textwidth}{1pt}}
    \vskip 3px
    
    \begin{minipage}{0.05\textwidth}
        \rotatebox{90}{Eval.}
    \end{minipage}
    \begin{minipage}{0.93\textwidth}
        \begin{minipage}{0.49\textwidth}
            \small
            \centering
            \begin{tabular}{|r | c | c|}
                Model A & PL & \PIRN\\\hline
                $\E_{p(\lambda| b, x)}\{ \delta[Y_{\text{top-}1}(\lambda) \subseteq C_{\text{top-}3}(x)]\}$ & $0.7$ & $0.52$
            \end{tabular}
        \end{minipage}
        \hfill
        \begin{minipage}{0.49\textwidth}
            \small
            \centering
            \begin{tabular}{|r | c | c|}
                Model B & PL & \PIRN\\\hline
                $\E_{p(\lambda| b, x)}\{ \delta[Y_{\text{top-}1}(\lambda) \subseteq C_{\text{top-}3}(x)]\}$ & $1$ & $0.99$
            \end{tabular}
        \end{minipage}
    \end{minipage}
    \caption{Plausibilities and evaluation results for the example and annotations from Figure \ref{fig:introduction-derm}. Our statistical aggregation models, PL and \PIRN, produce distributions over plausibilities (samples indicated in {\color{green!50!black}green}, \red{medium reliability}), contrasted with deterministic aggregation such as IRN or ML ({\color{blue}blue}). These plausibilities represent categorical distributions over all conditions (first row). We then evaluate uncertainty-adjusted metrics based on these plausibilities: given prediction sets for models A and B (second row), we evaluate how often, in expectation, the top-1 label from the plausibilities is included (third row).}
    \label{fig:results-motivation}
\end{figure*}

For example, we illustrate the applicability of our framework on a class of ranking metrics:
Specifically, for large prediction set sizes $k = |C|$, the annotation certainty can be very low as achieving exact match of large prediction and target sets can be a rare event. Thus, it seems natural to consider less stringent metrics based on the intersection of prediction set $C$ and ground truth set $Y$ instead of requiring equality:
\begin{align}
    \text{Overlap}(C, Y) = \frac{|C \cap Y|}{|C|}.\label{eq:methods-overlap}
\end{align}
As we do not know the target ground truth set, we use instead the expectation under the aggregation procedure $p(\lambda| b, x)$ where we define an uncertainty adjusted overlap as 
\begin{align}
    \text{UA-Overlap}(C, x) = \E_{p(\plaus{}{}| b, x)} \Big\{\text{Overlap}(C, Y_{\text{top}-|C|}(\lambda)) \Big\}.
\end{align}
Following \citep{WuSAC2003,WebberTOIS2010}, we also consider overlaps 
of increasingly larger prediction and ground truth sets and define uncertainty adjusted average overlap as
\begin{align}
\begin{split}
    \text{UA-AverageOverlap@L}\hspace*{3cm}\\
    = \E_{p(b, x)} \Big\{\tfrac{1}{L} \sum_{k=1}^L \text{UA-Overlap}(C_{\text{top-}k}(x), x)\Big\}.\label{eq:methods-ua-average-overlap}
\end{split}
\end{align}
Note that using this formulation also avoids difficulties computing average overlap directly against partial rankings, see \ref{appendix-average-overlap}.
Similar uncertainty-adjusted variants can be defined for other ranking-based metrics \citep{Sakai2013}, e.g., Kendall's tau or Spearman's footrule \citep{Shieh1998,FaginSIDMA2003,FaginPODS2004,KumarWWW2010,VignaWWW2015}. In general, we expect that uncertainty-adjusted variants can easily be defined for almost all relevant metrics in machine learning.
\section{Experiments}
\label{sec:derm}
\label{sec:results}

We present comprehensive experiments on the skin condition classification data of \citep{LiuNATURE2020}, where we expect high annotation uncertainty across a significant proportion of the dataset. As outlined in Section \ref{subsec:method-diff}, we initially use inverse rank normalization (IRN) for deterministic aggregation.
We experiment with our two statistical aggregation models, probabilistic IRN (\PIRN) and Plackett--Luce \citep{Plackett1975,Luce2012} (PL).
Both models include a reliability parameter, described in Section \ref{sec:methods-metrics}, that reflects our trust in the annotators.
\red{While domain expertise might help select an appropriate rater reliability, we consider the case where it may be unclear} what the ``right'' reliability for our statistical aggregation models should be. \red{Thus,} we conduct experiments across a range of reliabilities, exploring different levels of annotator trust.

\subsection{Dataset, \red{Model and Example}}

The dataset of \citep{LiuNATURE2020} includes $K=419$ different conditions which are to be predicted from three $448\times448$ pixel images taken with consumer-grade cameras. Each annotation includes a variable number of conditions combined with a confidence value. As these confidence values are not comparable across dermatologists, previous work \citep{EngBJD2019,RoyMIA2022,AziziARXIV2022} uses them to obtain differential diagnoses, formalized as partial rankings of conditions. Details and statistics on the annotations are provided in \ref{sec:app-data}. We train several classifiers using cross-entropy against the IRN plausibilities by \red{randomly varying architecture and hyper-parameters and picking four models for our experiments. We follow the exact same set up as described in detail in \citep{RoyMIA2022} and \ref{app:training}}. We randomly selected four of these classifiers for evaluation and comparison.

\begin{figure*}[t]
    \centering
    \hspace*{-0.35cm}
    \begin{tabular}{@{}c@{}c@{}c@{}c@{}c@{}}
        \includegraphics[height=4cm]{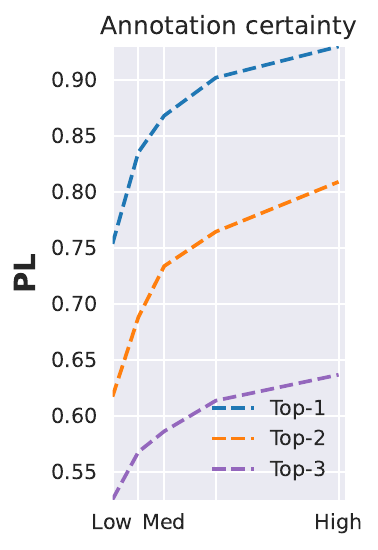}
        & \includegraphics[height=4cm]{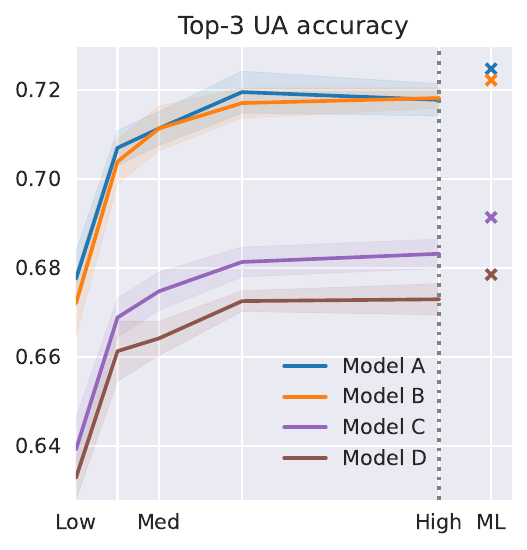}
        & \includegraphics[height=4cm]{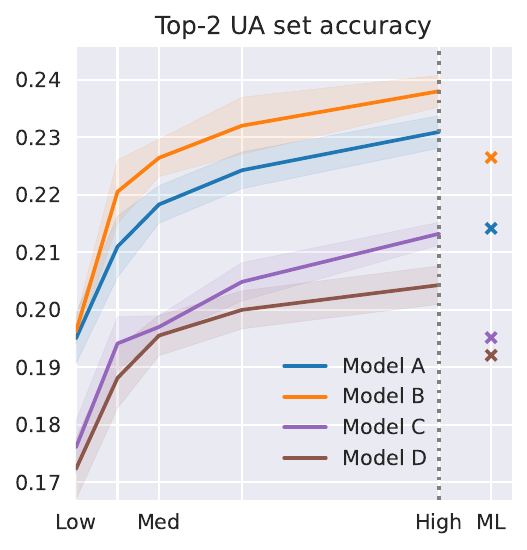}
        & \includegraphics[height=4cm]{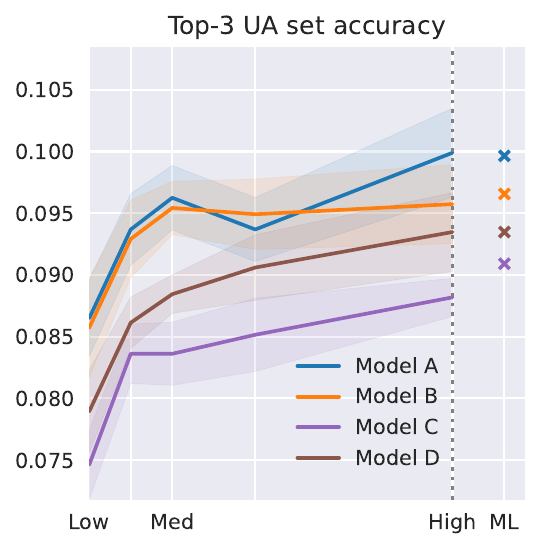}
        & \includegraphics[height=4cm]{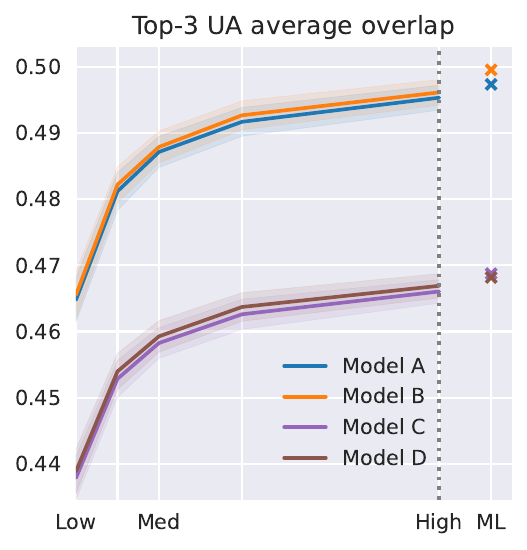}
        \\
        \includegraphics[height=4cm]{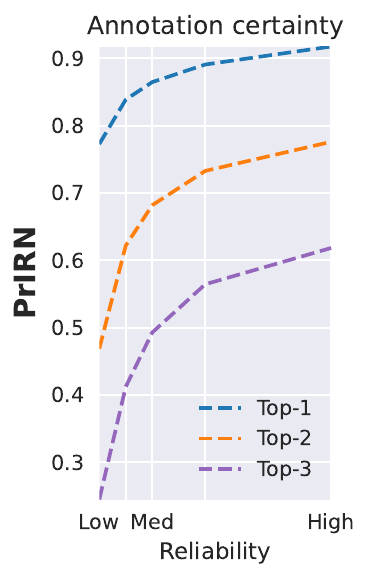}
        & \includegraphics[height=4cm]{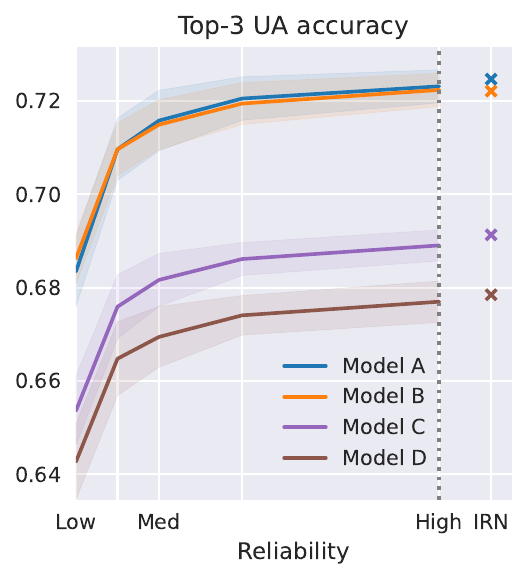}
        & \includegraphics[height=4cm]{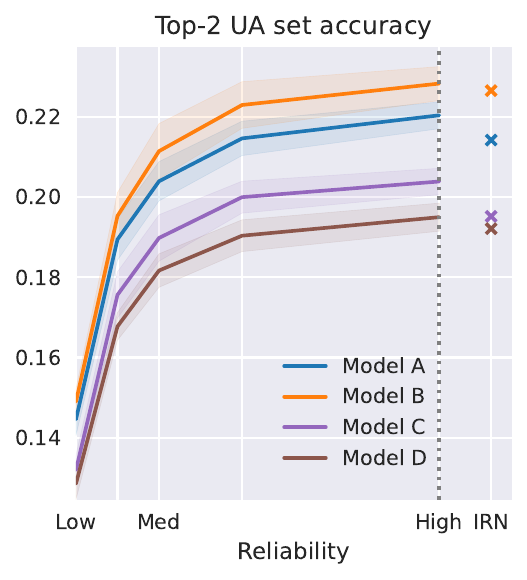}
        & \includegraphics[height=4cm]{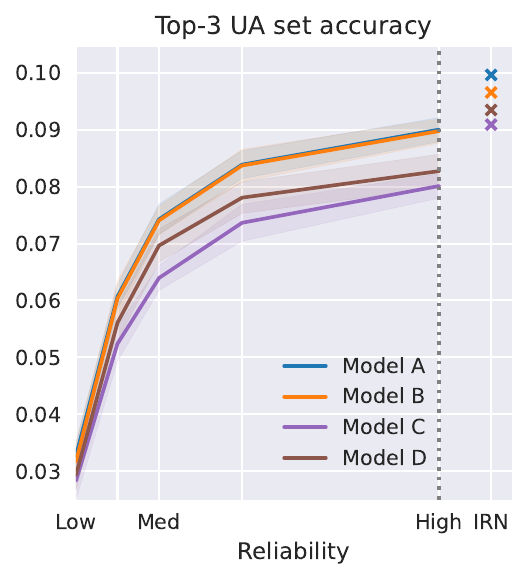}
        & \includegraphics[height=4cm]{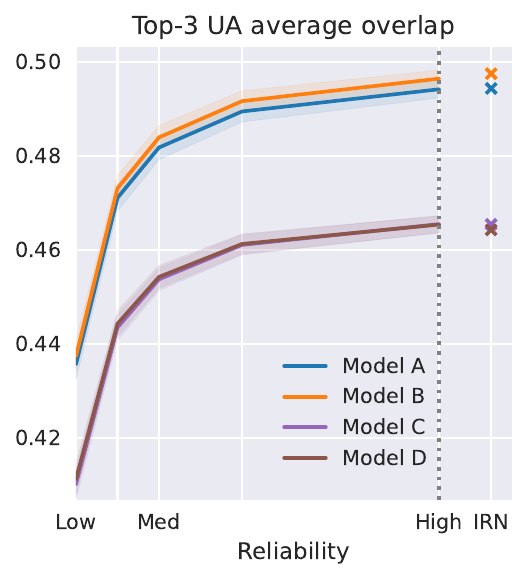}
    \end{tabular}
    \vspace*{-8px}
    \caption{For PL (top) and \PIRN~(bottom), we plot annotation certainty as well as uncertainty-adjusted top-3 accuracy, top-2 and top-3 \emph{set} accuracy and top-3 average overlap (y-axis). We consider four classifiers, A to D, across various reliabilities (x-axis). We omit exact reliability values as they are inherently not comparable between PL and \PIRN. Annotation certainty and evaluation metrics are averages across all examples and $M = 1000$ plausibility samples. The shaded region additionally reports the standard deviation of metrics after averaging across plausibility samples; colored $\times$ mark performance against point estimates (ML for PL, IRN for \PIRN). We find that reliability severely impacts evaluated performance and that there is a significant variation induced by annotation uncertainty. Classifiers tend to perform significantly worse when evaluated against more than the top-1 labels using set accuracy.}
    \label{fig:results-comparison}
\end{figure*}

Plausibilities from \PIRN and PL for the case of Figure \ref{fig:introduction-derm} are shown in Figure \ref{fig:results-motivation} (top row). Compared to ML and IRN ({\color{blue}blue bars}), there is clearly significant variation in the sampled plausibilities ({\color{green!50!black}green dots}) to the extent that the two most likely conditions (Hemangioma and Melanoma) may swap their positions. \red{Note that some plausibility samples can change the order of conditions radically from their deterministic, i.e., infinite reliability, counterparts (ML for PL and IRN for \PIRN). The example also highlights differences between PL and \PIRN. As IRN resolves ties deterministically by equally distributing weight across tied conditions, the rank of conditions such as ``Melanocytic Nevus'' that are involved in ties can be unintuitive after aggregation.} This also impacts evaluation: considering the two prediction sets from Figure \ref{fig:results-motivation} (second row), the first model (A) does \emph{not} include both conditions, while the second model (B) does. Concretely, model A includes the top-1 label in only 70\% of the sampled plausibilities, while model B always includes the top-1 label (100\%), cf. Figure \ref{fig:results-motivation} (third row). In a nutshell, this summarizes our evaluation methodology on one specific case with fixed reliability, which we extend to the whole test set and across multiple reliabilities next.

\subsection{Main results}

\begin{figure*}[t]
    \centering
    \includegraphics[height=3cm]{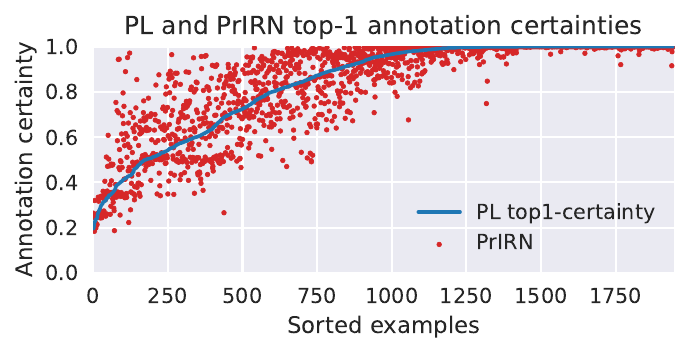}
    \includegraphics[height=3cm]{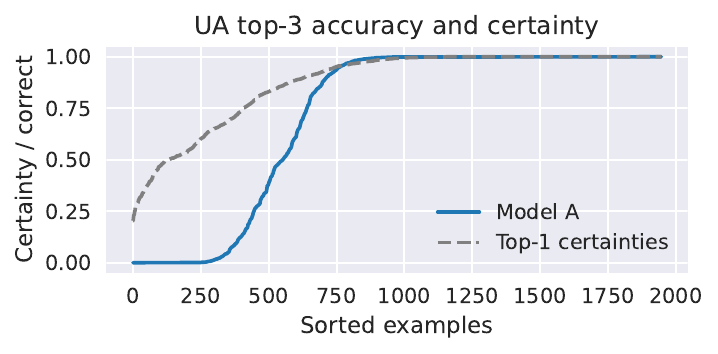}
    \includegraphics[height=3cm]{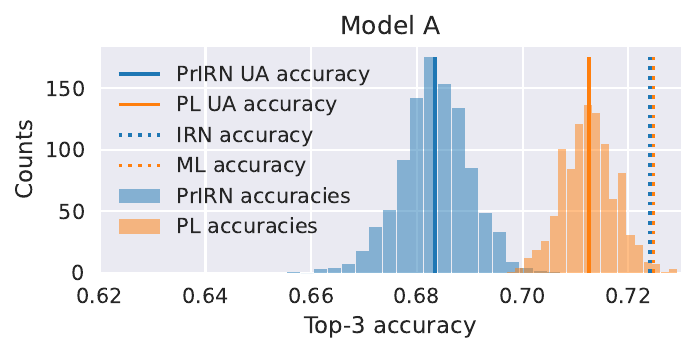}
    \vspace*{-8px}
    \caption{For a fixed, medium reliability, we present annotation certainty and uncertainty-adjusted top-3 accuracy across examples and plausibility samples. Left: PL top-1 annotation certainty plotted against sorted examples ({\color{blue}blue}) in comparison to \PIRN ({\color{red}red}). While there is high correlation between PL and \PIRN, there can be significant difference for individual examples.
    Middle: Uncertainty-adjusted top-3 accuracy for model A against sorted examples ({\color{blue}blue}). For many examples, the classifier does not consistently include all possible top-1 ground truth label in its predictions, resulting in values between $0$ and $1$. \red{Note that top-1 certainties are sorted separately.}
    Right: Histogram plot of uncertainty-adjusted top-3 accuracy (averaged over examples) across the $M = 1000$ plausibility samples. Between worst- and best-case plausibilities, there can be up to 4\% difference in accuracy.}
    \label{fig:results-certainty}
\end{figure*}

Our main results focus on evaluating (top-k) annotation certainty alongside uncertainty-adjusted accuracy across classifiers and reliabilities.
As reliabilities are inherently not comparable across PL and \PIRN, we omit specific values.
Then, Figure \ref{fig:results-comparison} (first column) highlights that average annotation certainty clearly increases with higher reliability; eventually being $1$ at infinite reliability. Also, top-$2$ and top-$3$ annotation certainty is significantly lower than top-$1$ annotation certainty, meaning that there is significant uncertainty not only in the top-$1$ condition but also lower ranked ones. This annotation uncertainty also has a clear impact on accuracy: Uncertainty-adjusted top-$3$ accuracy (second column) reduces significantly for lower reliability. As indicated by the shaded region, there is also high variation in accuracy across plausibility samples. In contrast, the ML and IRN plausibility point estimates, as used in previous work \citep{LiuNATURE2020}, typically overestimate performance and cannot provide an estimate of the expected variation in performance.

In order to account for inherent uncertainty in the evaluation, we additionally consider \emph{set} accuracy in Figure \ref{fig:results-comparison} (third and fourth columns).
Strikingly, set accuracy is dramatically lower than standard accuracy, highlighting that the trained classifiers perform poorly on conditions likely ranked second or third. This is further emphasized by the fact that the reduction in accuracy is significantly larger than the corresponding drop in annotation certainty. Moreover, shifting focus to set accuracy also impacts the ranking of classifiers. This can have severe implications for hyper-parameter optimization and model selection which is typically based purely on standard accuracy at infinite reliability. We also observe more significant differences between using \PIRN and PL, e.g., in terms of absolute accuracy numbers, their variation, or differences across classifiers. This highlights the impact that the statistical aggregation model can have on evaluation.
Lastly, Figure \ref{fig:results-comparison} (last column) also shows top-3 uncertainty-adjusted average overlap, where second and third conditions are weighed by $\nicefrac{1}{2}$ and $\nicefrac{1}{3}$, respectively. Besides resulting in generally higher numbers, this also reduces variation significantly.

\begin{figure*}[t]
    \centering
    {\small
    \hspace{0cm}PL risk and condition annotation certainty\hspace{2.5cm}Expected risk from PL plausibilities
    
    \includegraphics[trim={0 0 0 0.65cm}, clip,height=3.25cm]{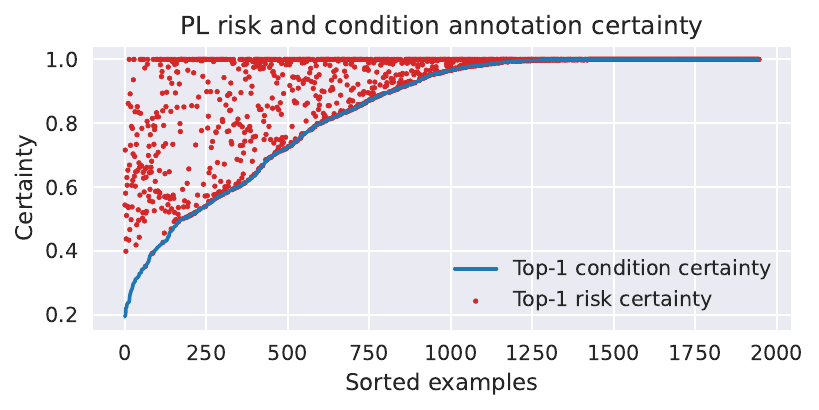}
    \includegraphics[trim={0 0 0 0.65cm}, clip,height=3.25cm]{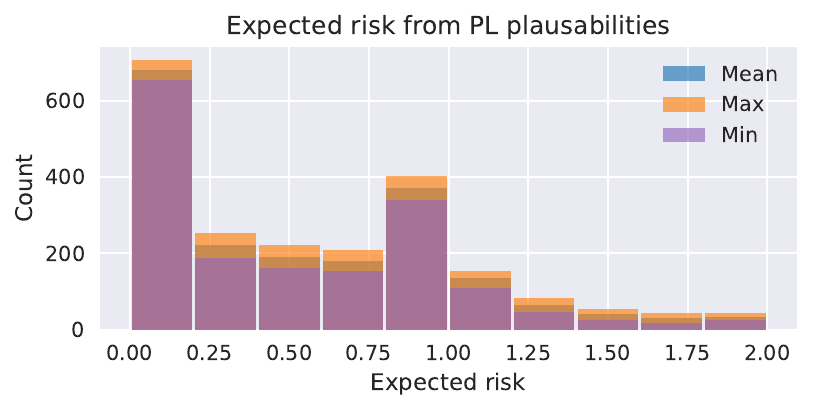}}
    \vspace*{-8px}
    \caption{Implications of ground truth uncertainty on risk categories.
    Right: Annotation certainty computed for risks in {\color{red}red} in comparison to conditions {\color{blue}blue}. Going from 419 conditions to 3 risk levels clearly increases annotation certainty on average. But for many examples, risk categories remain very uncertain.
    Right: For many examples, there is evidence for multiple risk categories within the annotations. We plot a histogram of \emph{expected risk}, the \red{mean/max/min} expected risk category given the plausibilities, i.e., distributions over conditions, \red{after mapping risk levels to low = 0, medium = 1 and high = 2 (see text)}.
    }
    \label{fig:results-risk}
\end{figure*}

\subsection{Analysis}

In the following, we focus on a fixed, medium reliability (as annotated in Figure \ref{fig:results-comparison}) and consider annotation certainty and uncertainty-adjusted accuracy across examples.
Specifically, Figure \ref{fig:results-certainty} (left) plots annotation certainty from PL ({\color{blue}blue}) and \PIRN ({\color{red}red}) over examples (sorted for PL):
For at least a quarter of the examples there is significant annotation uncertainty, i.e., top-$1$ annotation certainty is well below $1$.
We also observe that annotation certainty is strongly correlated between PL and \PIRN (correlation coefficient 0.9). This indicates that similar examples are identified as having high annotation uncertainty. However, on individual examples, there can still be a significant difference. Similarly, we found that annotation certainty correlates well with annotator disagreement (see \ref{sec:app-results}). Figure \ref{fig:results-certainty} (middle) also shows uncertainty-adjusted top-3 accuracy against (sorted) examples. Again, for at least a quarter of examples, uncertainty-adjusted accuracy is \red{well} below $1$, i.e., the top-$3$ prediction sets do not always include all possible top-1 ground truth labels. In Figure \ref{fig:results-certainty} (right), we also show results across plausibility samples. While Figure \ref{fig:results-comparison} depicted the variability based on plausibility samples only through standard deviation error bars, these histograms clearly show that accuracy can easily vary by up to 4\% between best and worst case.

Besides performance evaluation across all $419$ distinct conditions, previous work \citep{RoyMIA2022} also put significant focus on classifying risk categories, considering low, medium or high risk conditions. These categories are assigned to each condition independent of the actual case (e.g., Melanoma is a high-risk condition). As recommendations to users (e.g., whether the user should see a specialist) are similar for conditions in the same risk categories, it is often more important to correctly classify risk categories compared to individual conditions. Figure \ref{fig:results-risk} (left), however, shows that these risk categories are also subject to significant uncertainty. This is made explicit by computing top-$1$ annotation certainty for \red{risk categories} ({\color{red}red}) derived from the plausibilities over conditions. While this is generally higher than annotation certainty for conditions ({\color{blue}blue}) due to the smaller label space (3 risks vs. 419 conditions), annotation certainty remains low for many cases. 
This also has far-reaching consequences for evaluation. For example, evaluation metrics such as accuracy are often conditioned on high-risk cases. That is, for evaluation, we are interested in a classifier's accuracy only considering high-risk cases. This conditioning, however, is not well defined in light of this uncertainty. This is made explicit in Figure \ref{fig:results-risk} (right) which plots \emph{expected risk}: the expected risk assignment for cases, based on plausibility samples, after mapping risk levels to an ordinal scale; low = 0, medium = 1, and high = 2. \red{That is, taking the expectation of the risk level (0, 1, or 2) over a plausibility sample. The figure then shows mean/max/min across our plausibility samples.} Most cases do not yield crisp risk assignments as there is typically evidence for multiple risk categories present in the annotations.

\subsection{Discussion}
\label{subsec:results-annotator}

The results in this paper indicate, using our annotation certainty measure, that a large portion of the dataset exhibits high ground truth uncertainty. 
The current approach \citep{LiuNATURE2020} of deterministically aggregating annotations using IRN and then evaluating against the corresponding top-1 labels largely ignores this uncertainty. In our framework, using the \PIRN model, this implicitly corresponds to evaluation at \emph{infinite} reliability, i.e., full trust in all annotators. Instead, our approach to evaluation paints a more complete picture by computing \emph{uncertainty-adjusted} (top-k) accuracy across a range of reliabilities, corresponding to different trust ``scenarios''. In practice, this not only allows to compare models across these different scenarios but also highlights the expected variation in performance. Moreover, we show that performance is always relative to the chosen aggregation model, as highlighted using our alternative PL-based models. \red{Differences between results obtained using \PIRN and PL stem from different assumptions or biases. For example, \PIRN's treatment of ties appears unintuitive, suggesting PL is more reliable for our application. This also leads to generally lower performance when evaluating against \PIRN compared to PL as plausibility mass is more equally distributed for many ambiguous examples due to these ties.}

\red{Another important aspect of our uncertainty-adjusted metrics is that no model may be able to achieve, e.g., 100\% \emph{top-1} uncertainty-adjusted accuracy if the ground truth is deemed sufficiently uncertain. As our uncertainty-adjusted metrics are comparable to their regular counterparts, this indicates that the top-1 prediction is not sufficient.}
\red{Analogously, we also show that considering} more than the top-1 condition for evaluation can be important. Our results show performance drops rather drastically, indicating potential negative consequences for patients when \red{critical lower-ranked conditions are not considered.}
For example, seemingly random conditions on the 2nd or 3rd place of the prediction set can lead to confusion or anxiety.

We also qualitatively evaluated our framework in an informal study with two US board-certified dermatologists familiar with the labeling tool \citep{LiuNATURE2020}. Specifically, we discussed individual cases with particularly low annotation certainty by showing input images alongside meta information (sex, age, etc.) and the corresponding annotations (cf. Figure \ref{fig:results-motivation}). Discussing these cases takes considerable time as the dermatologists try to understand how the annotators came to their respective conclusions. In most cases, the disagreement was attributed to inherent uncertainty, i.e., missing information, inconclusive images, etc. In only \red{a} few cases, the disagreement was attributed to annotator mistakes or annotation quality in general -- e.g., inexperienced annotators, annotators ignoring meta information etc.
Once again, this highlights the difficulty of disentangling annotation and inherent uncertainty in cases with high disagreement (as discussed in Sections \ref{subsec:methods-model} and \ref{subsec:methods-certainty}), aligning  with related work on ``meta-annotation'' for understanding sources of disagreement \citep{SandriEACL2023,BhattacharyaICCV2019}. However, it also emphasizes that our uncertainty-adjusted metrics effectively consider both sources of uncertainty.

\red{Overall, we believe that our framework will help model development and improve model selection robustness, potentially benefiting patient outcomes. We also emphasize that annotation certainty and uncertainty-adjusted metrics are not dependent on the exact task and aggregation model used. Thus, these can readily be applied to other tasks with different annotation formats. Additionally, although PL was motivated here by our dermatology application, it is applicable across a wide range of diagnosis tasks where annotations often come in the form of partial rankings (i.e., differential diagnoses).}
\section{Related work}\label{sec:relatedwork}

Annotator disagreement has been discussed extensively and early on in medicine \citep{Feinstein1990,McHughBM2012,RaghuICML2019,Schaekermann2020} as well as machine learning \citep{DawidJRSS1979,SmythNIPS1994}. Natural language processing, for example, has particularly strong work on dealing with disagreement \citep{ReidsmaACL2008,AroyoHC2014,AroyoAI2015,SchaekermannSIGCHIWORK2016,DumitracheNAACL2019,RottgerNAACL2022,AbercrombieARXIV2023}, see \citep{PavlickTACL2019} for an overview. As crowdsourcing human annotations has become a standard tool in creating benchmarks across the field \citep{KovashkaFTCGV2016,SorokinCVPRWORK2008,SnowEMNLP2008} -- though not without criticism \citep{RottgerACL2021} -- most work focuses on resolving or measuring disagreement and aggregating annotations. Methods for measuring disagreement \citep{Feinstein1990,PowersEACL2012,UmaJAIR2021} are often similar across domains. However, measures such as Fleiss'/Cohen's kappa \citep{Cohen1960,Fleiss2003}, percent agreement \citep{McHughBM2012}, or intra-class correlation coefficient \citep{Landis1997} are only applicable to annotations with single class responses such that generalized approaches \citep{BraylanWWW2022} or custom measures are used for more structured annotations \citep{PavlickTACL2019}. Resolving disagreement is typically done \textit{computationally} (e.g., through majority vote). However, recent work has explored domain-specific \textit{interactive} approaches for resolving disagreement, involving discussions or deliberation \citep{SchaekermannCHI2020,SchaekermannCHI2020b,SchaekermannACMHCI2019,Schaekermann2020,PradhanFAI2022,DrapeauHCOMP2016,ChenCHI2019,Silver2021,BakkerNIPS2022} or relabeling \citep{ShengKDD2008}, and reducing disagreement by co-designing labeling with experts \citep{FreemanHCOMP2021}. Recent work also considers properly modeling disagreement \citep{VitsakisARXIV2023} or performing meta analysis \citep{SandriEACL2023,BhattacharyaICCV2019}, trying to understand sources of disagreement.

For benchmarks, disagreement is generally addressed by aggregating labels from multiple annotators to arrive at what is assumed to be the single correct label. This can involve basic majority voting or more advanced methods \citep{DawidJRSS1979,DemarneffeCL2012,PhamTPAMI2017,Warby2014,CarvalhoIJCAI2013,GauntUAI2016,TianTPAMI2019}, including inverse rank normalization (IRN) as discussed in this paper. Often, aggregation is also performed using probabilistic models from the crowdsourcing and truth-discovery literature \citep{YinTKDE2008,WelinderCVPR2010,LiPVLDB2012,DongPVLDB2009,ZhaoPVLDB2012,WangIPSN2012,BachrachICML2012,RodriguesAAAI2018,GuanAAAI2018,ChuAAAI2021,GordonCHI2022}, see \citep{YanML2014,ZhengPVLDB2017} for surveys. However, evaluation is often based on point estimates and the impact of annotator disagreement on evaluation is generally poorly understood \citep{GordonCHI2021}. While, e.g., \citep{ReidsmaACL2008,CollinsHCOMP2022} train models on individual annotators, \citep{DemarneffeCL2012,NieEMNLP2020} perform evaluation on aggregated probabilities instead of top-1 labels, and \citep{GaoTIP2016} trains on label distributions rather than discrete aggregated labels, there is no common understanding for dealing with annotation uncertainty for evaluation. Instead, following \citep{PlankARXIV2022}, disagreement is often treated as label noise.
Here, early work \citep{AngluinML1987,Kearns1998,Kearns1993,LawrenceICML2001} assumes uniform or class-conditional label noise, while more recent work \citep{BeigmanACL2009,OyenARXIV2022} also considers feature-dependent or annotator-dependent noise. Popular methods try to estimate the label noise distributions \citep{NorthcuttARXIV2019,HendrycksNIPS2018} in order to prune or re-weight examples.
Such approaches have also been utilized to infer annotator confusion matrices \citep{ZhangARXIV2020e,TannoCVPR2019}, similar to annotator quality in crowdsourcing. We refer to \citep{ChenICML2019,ZhangARXIV2022b} for good overviews and note that there is also some similarity to partial label learning \citep{HullermeierIDA2005,NguyenKDD2008,CourJMLR2011,WangICLR2022b}. 

Overall, work on handling annotation disagreement is very fragmented \citep{UmaJAIR2021} and focused on treating symptoms such as label noise rather than the underlying uncertainty in the ground truth. This is emphasized in recent position papers \citep{BaanARXIV2022,PlankARXIV2022,Valerio2021} that argue for common frameworks to deal with this challenge.
Indeed, recent work \citep{SculleySDM2007,MaierheinNATURE2018,CabitzaAS2020,BelzARXIV2023} demonstrates that many results in machine learning are not reproducible, in part due to annotation uncertainty. This has also been the basis for several workshops and challenges \citep{LeonardelliARXIV2023} and the main motivation for this paper. Closest to our work, \citep{GordonCHI2021,LovchinskyICLR2020} propose methods to incorporate label disagreement into evaluation metrics. However, their work is limited to binary classification tasks. Moreover, the considered annotations are unstructured, i.e., annotators merely provide single labels. 
In contrast, our framework for evaluation with annotation uncertainty if independent of task, domain or annotation format.
Finally, \citep{CollinsHCOMP2022} evaluates against labels aggregated from random subsets of annotators, which can be seen as a bootstrapping approach to statistical aggregation in our framework.
\section{Conclusion}
\label{sec:conclusion}

\red{
In almost all supervised learning tasks, ground truth labels are implicitly or explicitly obtained by aggregating annotations, e.g., majority voting. These deterministic aggregation methods ignore the underlying uncertainty and tend to over-estimate performance of models. This can be  particularly critical when assessing the quality of health models.

In this paper, we propose a statistical model for aggregating annotations that explicitly accounts for ground truth uncertainty. Based on this model, we provide a novel framework for understanding and quantifying annotation ambiguity. Specifically, we introduce a measure of annotation certainty to identify ambiguous data points and develop uncertainty-adjusted performance metrics for more reliable AI system evaluation.

We have demonstrated our methodology on a challenging skin condition classification problem addressed in  \citep{LiuNATURE2020}. In this context, our annotation uncertainty measure has revealed that a large proportion of cases in the training and test sets exhibit high ground truth uncertainty which has been ignored previously. Moreover, we observe that classifier performance, as estimated using our uncertainty-adjusted metrics, can exhibit significant variations, hinting at the possibility that true performance may be much lower in a real deployment scenario. The variations in model performance estimates become more pronounced when evaluating not only against possible top-1 ground truth labels but prediction sets derived from differential diagnoses \red{(as highlighted using \emph{set} accuracy)}.

While we have focused on a dermatology application, the framework's versatility allows adaptation to other research domains, with direct relevance to diagnosis tasks involving differential diagnoses or partial rankings. This approach is especially valuable in medical and healthcare contexts where nuanced interpretation of results is crucial.
}

\section*{Acknowledgements}

We would like to thank Annisah Um'rani and Peggy Bui for their support of this project as well as Naama Hammel, Boris Babenko, Katherine Heller, Verena Rieser, and Dilan Gorur for their feedback on the manuscript.

\section*{Data availability}

The de-identified dermatology data used in this paper is not publicly available due to restrictions in the data-sharing agreements.

\bibliographystyle{abbrvnat}
\bibliography{generated_bibliography}

\newpage
\begin{appendix}
\onecolumn
\section{Further introductory examples and details}
\label{sec:app-introduction-examples}

\begin{figure*}
    \centering
    \begin{minipage}[t]{0.23\textwidth}
        \vspace*{0px}
        \includegraphics[height=3cm]{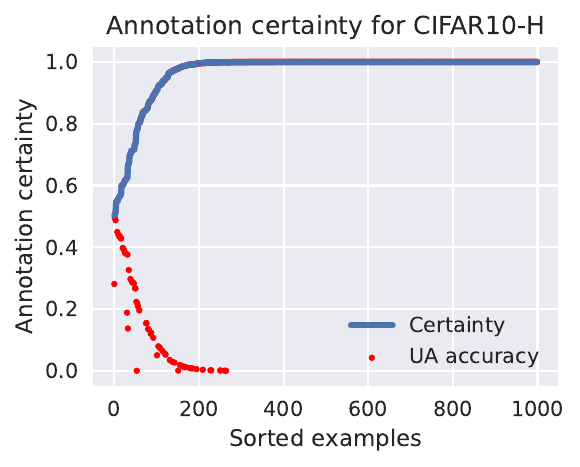}
    \end{minipage}
    \begin{minipage}[t]{0.17\textwidth}
        \vspace*{2px}
        \includegraphics[height=2.75cm]{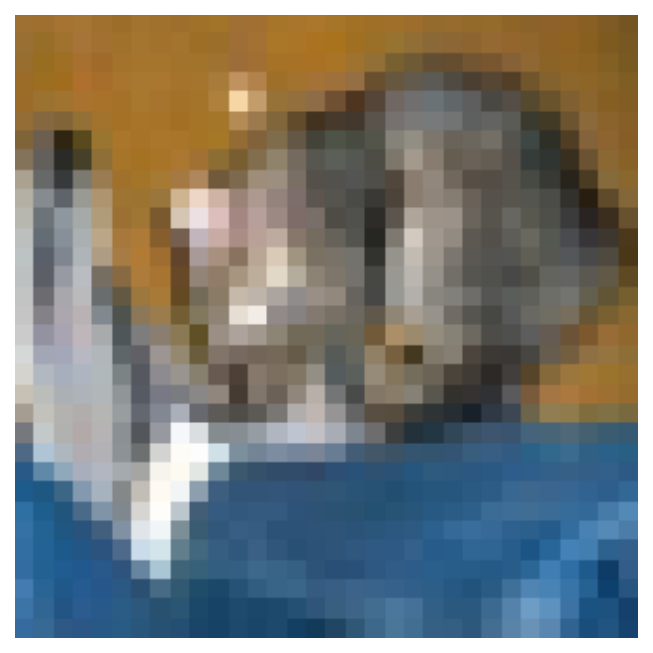}
    \end{minipage}
    \begin{minipage}[t]{0.2\textwidth}
        \vspace*{0px}
        \includegraphics[height=3cm]{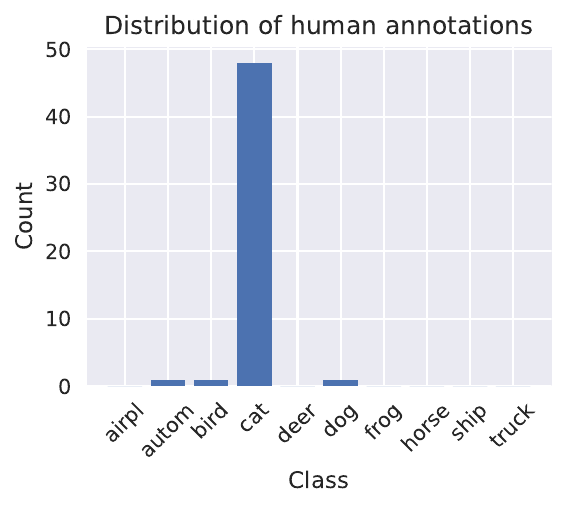}
    \end{minipage}
    \begin{minipage}[t]{0.17\textwidth}
        \vspace*{2px}
        \includegraphics[height=2.75cm]{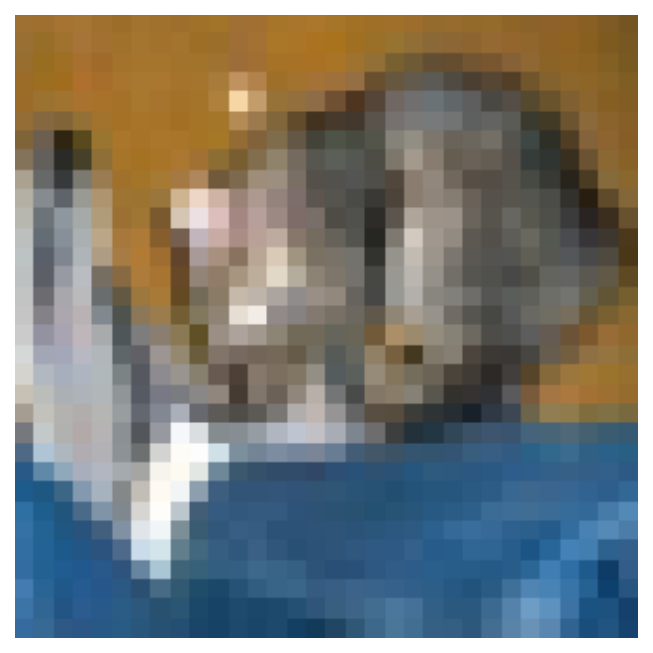}
    \end{minipage}
    \begin{minipage}[t]{0.2\textwidth}
        \vspace*{0px}
        \includegraphics[height=3cm]{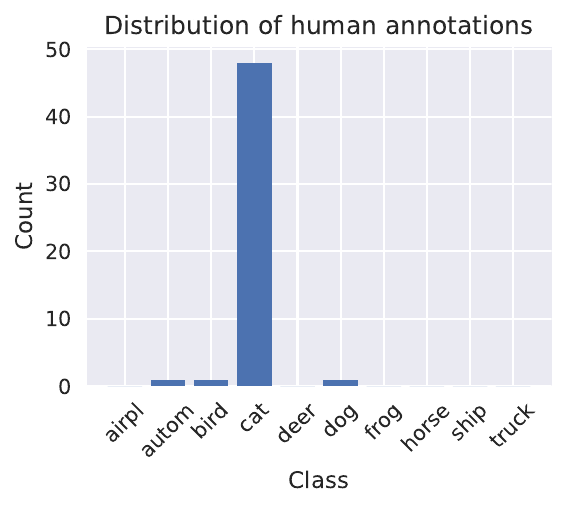}
    \end{minipage}
    \vspace*{-4px}
    \caption{We plot \emph{annotation certainty} and \emph{uncertainty-adjusted} (UA) accuracy for CIFAR10-H \citep{Krizhevsky2009,PetersonICCV2019}, highlighting the 1000 examples with lowest annotation certainty. There is a small set of $178$ images with certainty below $99\%$ for which annotators tend to disagree. Given the large number of annotators ($50$), this likely indicates examples with inherent uncertainty, as shown for two examples on the right (images and histogram of annotations).}
    \label{fig:problem-cifar10}
\end{figure*}

\textbf{CIFAR10:} 
To move from the toy example in Section \ref{subsec:introduction-examples} to a real benchmark, we consider CIFAR10 \citep{Krizhevsky2009} with the human annotations from \citep{PetersonICCV2019}. Again, each annotator provides a single label such that we can apply the same methodology. Specifically, we model $p(\lambda|b, x) = \mathcal{D}(\lambda;\mu)$ as Dirichlet distribution with concentration parameter vector $\mu$. We define $\mu = \gamma s + \alpha \mathbf{1}$ where $s$ is a vector of sufficient statistics with the $c$'th element $s_c$ denoting the number of annotators that have chosen $c$ as (top-1) label, $\gamma > 0$ is the reliability parameter, $\alpha>0$ is a small prior parameter that we assign to each label $c$, and $\mathbf{1}$ a vector of ones of the same shape as $s$. A higher reliability parameter $\gamma$ makes the plausibilities $\lambda \sim p(\lambda|b)$ more concentrated, resulting in higher annotation certainty. Here, the effect of $\alpha$ vanishes with higher reliability or more annotators; essentially, $\alpha$ represents a minimum number of annotators.

Figure \ref{fig:problem-cifar10} (left) reports the corresponding annotation certainty alongside an uncertainty-adjusted version of accuracy.
As can be seen, annotators agree for the majority of examples, resulting in certainties close to $1$. Nevertheless, on roughly $178$ test examples, certainty is below $0.99$. As there are 50 annotators per example with only 10 classes, it is likely that these are actually difficult cases, i.e. cases with inherent uncertainty, as confirmed in Figure \ref{fig:problem-cifar10}.
This corresponds to roughly $0.2\%$ of the examples. Interestingly, recent improvements in accuracy on CIFAR10 are often smaller than $0.2\%$\footnote{According to Papers with Code, \url{https://paperswithcode.com/sota/image-classification-on-cifar-10}.}. To measure accuracy while taking annotation uncertainty into account, we measure how often the original CIFAR10 ground truth labels coincide with the top-1 labels obtained from sampled plausibilities. We call this \emph{uncertainty-adjusted accuracy} (formally defined in Section \ref{subsec:methods-measuring}) and, unsurprisingly, observe that the original labels of \citep{Krizhevsky2009} (when interpreted as predictions) often perform poorly for examples with high annotation uncertainty ({\color{red}red} in Figure \ref{fig:problem-cifar10}).

\begin{figure*}[t]
    \centering
    \includegraphics[height=3.15cm]{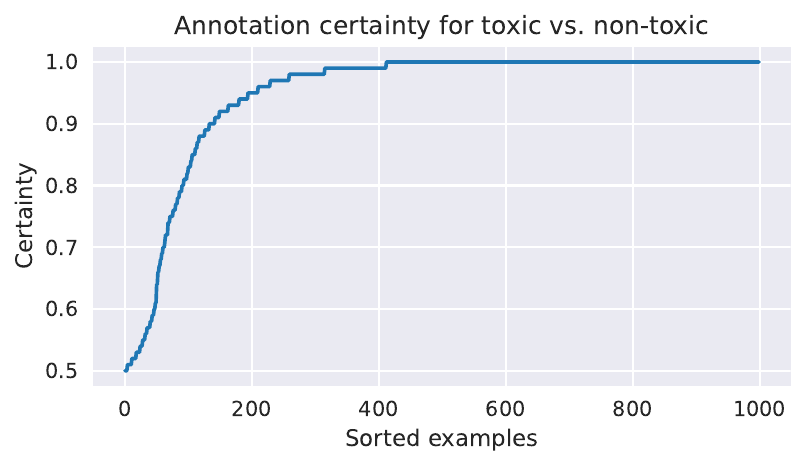}
    \includegraphics[height=3.15cm]{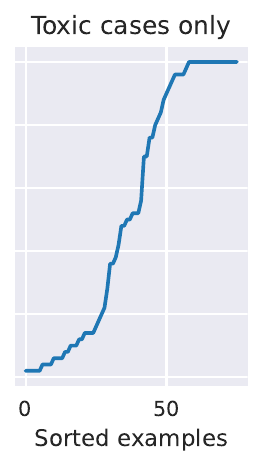}
    \includegraphics[height=3.15cm]{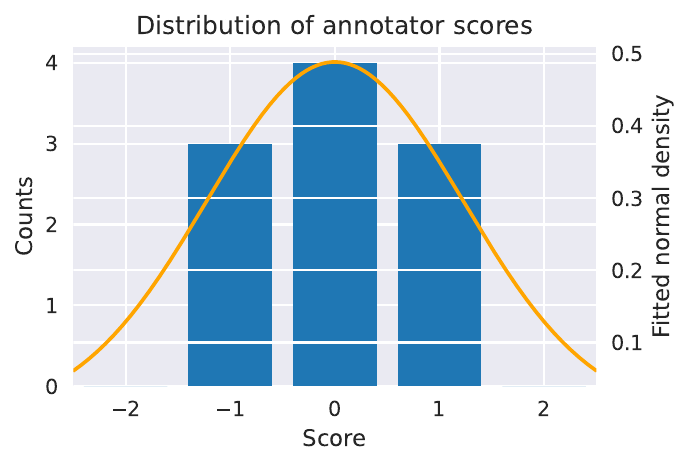}
    \includegraphics[height=3.15cm]{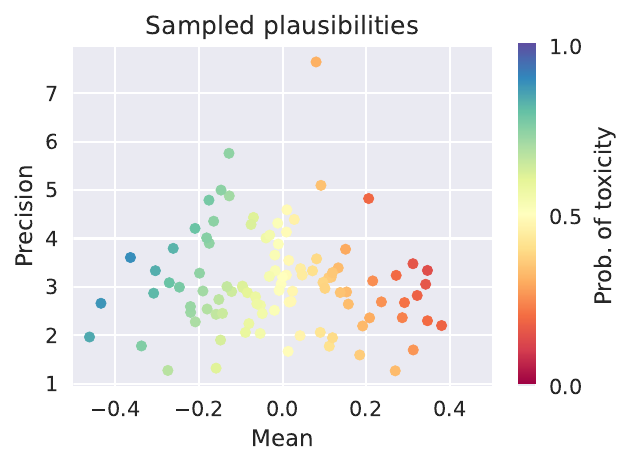}
    \vspace*{-12px}
    \caption{Annotation certainty for 1k examples from the Wikipedia toxicity dataset \citep{wikipediadetox} as well as those examples identified by the majority of annotators as toxic or very toxic. On the right, we show an example score distribution and samples from the corresponding posterior over plausibilities. Certainty is low for approximately $50\%$ of the toxic examples.}
    \label{fig:problem-wikipedia}
\end{figure*}

\textbf{Wikipedia Toxicity} \citep{wikipediadetox} is a natural language dataset for detecting toxicity in comments posted in Wikipedia page discussions. Around 10 annotators rate each comment between -2 (very toxic) and 2 (very healthy). Then, a comment can be treated as toxic if the mean rating is below 0. This suggests that the scores are assumed to be normally distributed. In this case, we define the plausibilities as the mean and standard deviation: $\plaus{}{} = (\mu, \sigma^2)$. As before, plausibilities parameterize the distribution over targets. Choosing an appropriate prior\footnote{We can put a normal-inverse-gamma prior on $\mu$ and $\sigma
^2$, i.e., $\mu \sim \mathcal{N}(\mu_0, \nu)$ and $\sigma^2 \sim \Gamma(\alpha, \beta)$ and estimate the parameters $\mu_0, \nu, \alpha$ and $\beta$ using the scores.} for $\mu$ and $\sigma$ allows to calculate certainty as the maximum of the two possible outcomes: $\mathcal{N}(y \leq 0; \mu, \sigma^2)$ and $\mathcal{N}(y > 0; \mu, \sigma^2)$. Figure \ref{fig:problem-wikipedia} shows the computed certainties for the first 1k examples of the dataset. A significant part of the examples is highly uncertain, especially among the toxic cases (majority of annotators responding -2 or -1, roughly 1\% of all cases). We also show one qualitative example with the corresponding annotations and plausibility samples. Note that, in this case, plausibilities correspond to one-dimensional Gaussian distributions which are plotted using mean and precision.

\begin{figure*}[t]
    \centering
    \includegraphics[height=3.15cm]{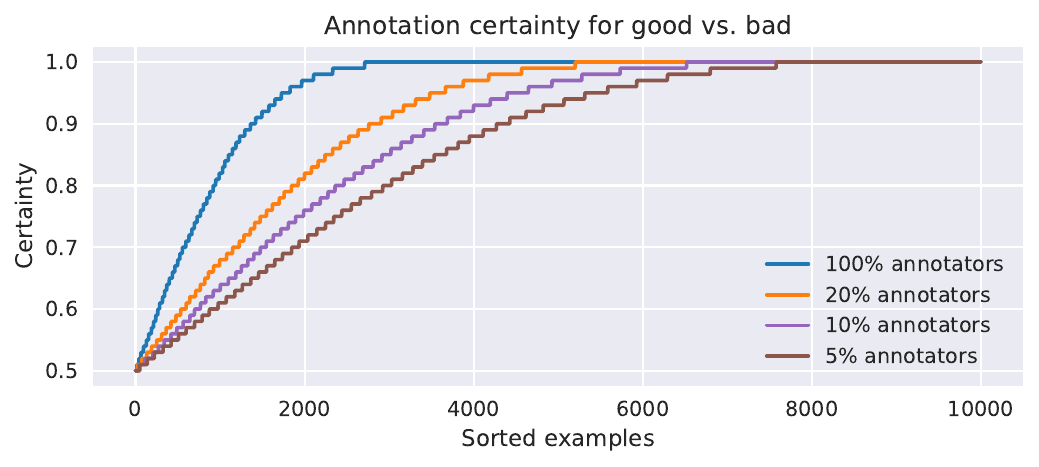}
    \includegraphics[height=3.15cm]{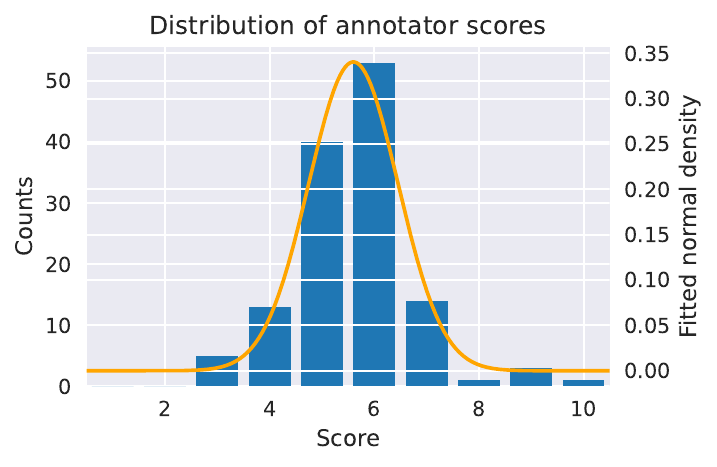}
    \includegraphics[height=3.15cm]{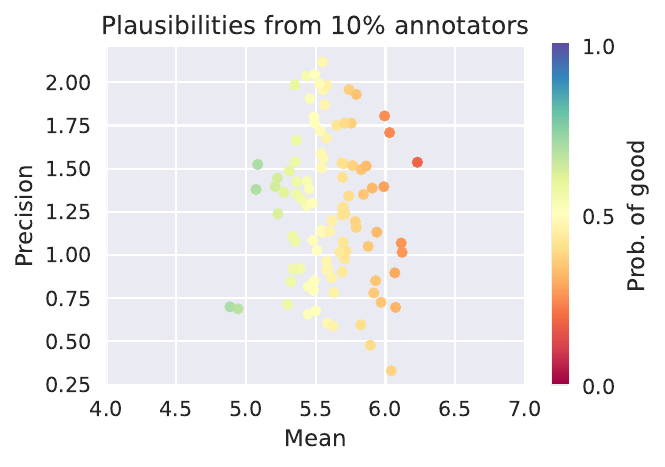}
    \caption{Left: Annotation certainty for the first 10k AVA \citep{MurrayCVPR2012} images considering a binary classification setting of poor images (score $<5$) against good images (score $\geq 5$). Certainty might be high primarily due to the large number of annotators. Right: Score distribution for one example together with samples from the corresponding posterior $p(\plaus{}{} = (\mu, \sigma^2)|b)$.}
    \label{fig:problem-ava}
\end{figure*}

\textbf{AVA} \citep{MurrayCVPR2012} is a dataset of visual aesthetics scores. Photography enthusiasts and experts rated more than 200k images on a scale between 1 and 10. Compared to CIFAR10, this is a more subjective and thus ambiguous labeling task. Nevertheless, the dataset is commonly used for predicting visual aesthetics, e.g., by predicting poor or good images \citep{LuMM2014,EsfandaraniTIP2018}. For simplicity, this can be done by taking the mean of the ratings and classifying images with mean score lower than $5$ as poor and those with mean score higher or equal to $5$ as good. This suggests that the scores are assumed to be normally distributed. Specifically, $\plaus{}{} = (\mu, \sigma^2)$, modeling mean and standard deviation. Using the same normal-inverse-gamma prior as above, we can compute certainty by estimating $\mathcal{N}(y \leq 5; \mu, \sigma^2)$ and $\mathcal{N}(y > 5; \mu, \sigma^2)$. Figure \ref{fig:problem-ava} shows the computed certainties as well as an example of a typical score distribution and plausibility samples. Surprisingly, for the first 10k images of AVA, most examples receive high certainty. However, this does \emph{not} imply that there is no inherent ambiguity, as highlighted in Figure \ref{fig:problem-ava}. Instead, it might just show that there is lower annotation uncertainty due to the high number of, on average, $222$ annotations.

\section{Plackett--Luce likelihood and Gibbs sampling}

Recall the definition of differential diagnoses as partial rankings in Section \ref{subsec:method-diff}. Furthermore, we say that annotators provide a \emph{full} ranking if all blocks except $b_L$ have size one, i.e., $|b_i| = 1$, $i < L$. We say the annotators provide a \textit{complete} ranking if all blocks are of size one, i.e., $|b_i| = 1$, $i \leq L$. Note that we can have multiple annotators providing partial rankings, in which case we let $\block{}{}{}^r$, $r \in [R]$, denote the annotation by the annotator $r$ out of $R$ annotators.

\subsection{Exact PL likelihood computation for partial rankings}\label{app:pl_exact}

\begin{figure}
    \centering
    \includegraphics[width=0.55\textwidth]{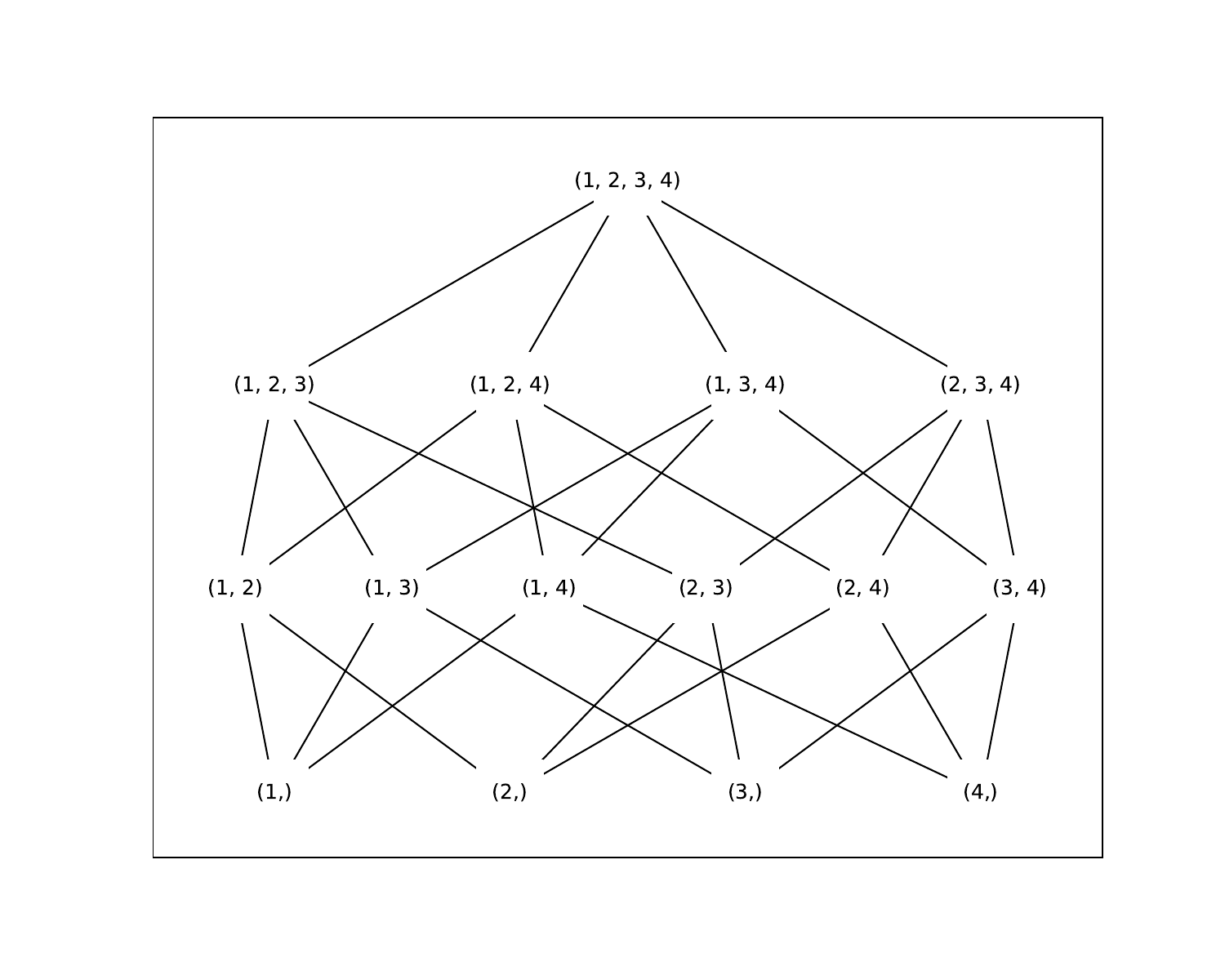}
    \vspace*{-12px}
    \caption{Hasse diagram of subsets of $\{1,2,3,4\}$. To compute the probability 
    of a partial order of the form $\{1,2,3,4\} \succ \bar{b}$ 
    exactly under a PL distribution with parameters $\lambda$,  we associate the value 
    $\mathcal{R}(A)$ with each node $A \subset \{1,2,3,4\}, A \neq \emptyset$. 
    After initialization of the bottom nodes, that correspond to subsets with only a single element, 
    the algorithm proceeds by computing values of $\mathcal{R}(A)$ layer by layer using 
    the update $\mathcal{R}_A \leftarrow {\sum_{a \in A} \mathcal{R}({A \setminus \{a\}}})/{(\bar{Z} + \sum_{a \in A} \lambda_a)} $
    for each node. Here, the total plausibility of the classes in $\bar{b}$ is given as $\bar{Z} = \sum_{k \in \bar{b}} \lambda_k$. The desired probability is $p(\{1,2,3,4\} \succ \bar{b}) = \Big( \prod_{k \in \{1,2,3,4\}} \lambda_k \Big) \mathcal{R}({\{1,2,3,4\}})$. 
    } 
    \label{fig:methods-recursion}
\end{figure}

For a more concise exposition, we will temporarily assume that we have a single annotator providing a partial ranking $b$, and will introduce the multiple annotator formulation at the end of Section~\ref{subsec:methods-gibbs}. We see that for the partitioning $b = (\block{}{}{1}, \block{}{}{2}, \dots, \block{}{}{L})$
the probability $p(\block{}{}{}|\plaus{}{}) = p(\block{}{}{1} \succ \block{}{}{2} \succ \dots \succ \block{}{}{L})$ admits the factorization
\begin{align*}
    p(\block{}{}{}| \lambda) & = p(\block{}{}{1}| \lambda) p(\block{}{}{2}| \lambda, \block{}{}{1} ) \dots p(\block{}{}{L-1}| \lambda, \block{}{}{1:L-2}).
\end{align*}
Note that above we have redefined $b$ to include all unranked classes, if any, in $b_L$, such that $b_1 \cup \dots \cup b_L = [K]$ with $[K]$ being the set of all $K$] classes. Hence we omit the term $p(b_L|\plaus{}{}, b_{1:L-1}) = 1$.
Here, the first term is the probability of drawing a permutation from the PL distribution that starts with a sequence of classes that are all in $\block{}{}{1}$:
\begin{align*}
    p(\block{}{}{1}| \lambda) = p(\block{}{}{1} \succ (\block{}{}{2} \cup \block{}{}{3} \cup \dots \cup \block{}{}{L}))
    = \sum_{\sigma} \ind{\sigma_{1:|\block{}{}{1}|} \in \mathcal{S}(\block{}{}{1})} p(\sigma | \lambda).
\end{align*}
where $\mathcal{S}(\block{}{}{})$ refers to all permutations compatible with $b = (\block{}{}{1}, \block{}{}{2}, \dots, \block{}{}{L})$. With a slight abuse of notation, we let $\mathcal{S}(\block{}{}{1})$ be the set containing all possible permutations of the first block.
Concatenating all classes such that $\sigma_{1:|\block{}{}{i}|}$ corresponds to the first block, this is essentially the probability that the prefix $\sigma_{1:|\block{}{}{1}|}$ is in the first block $\block{}{}{1}$. Now, we define the cumulative sum of block sizes as $c_l = |b_1| + |b_2| + \dots + |b_l|$ with $c_0 = 0$ and $c_L = K$ and get
\begin{align*}
\begin{split}
    p(\block{}{}{2}| \lambda, \block{}{}{1}) & = p(\block{}{}{2} \succ (\block{}{}{3} \cup \dots \cup \block{}{}{L}) | \block{}{}{1} \succ (\block{}{}{2} \cup \block{}{}{3} \cup \dots \cup \block{}{}{L}))\\
    & = \frac{p(\block{}{}{1}, \block{}{}{2}| \lambda)}{p(\block{}{}{1}| \lambda)} = \frac{\sum_{\sigma} \ind{\sigma_{c_0+1:c_1} \in \mathcal{S}(\block{}{}{1})} \ind{\sigma_{c_1+1:c_2} \in 
    \mathcal{S}(\block{}{}{2})} p(\sigma | \lambda)}{\sum_{\sigma} \ind{\sigma_{c_0+1:c_1} \in \mathcal{S}(\block{}{}{1})} p(\sigma | \lambda)}
\end{split}
\end{align*}
for the second block.
This suggests that we can generalize this for any arbitrary block $l \in [L]$ as follows:
\begin{align*}
p(\block{}{}{l}| \lambda, \block{}{}{1:l-1} )
= \frac{\sum_{\sigma} \Big(\prod_{\kappa=1}^l\ind{\sigma_{c_{\kappa-1}+1:c_\kappa} \in \mathcal{S}(\block{}{}{\kappa})} \Big) p(\sigma | \lambda)}{\sum_{\sigma} \Big(\prod_{\kappa=1}^{l-1}\ind{\sigma_{c_{\kappa-1}+1:c_\kappa} \in \mathcal{S}(\block{}{}{\kappa})} \Big) p(\sigma | \lambda)},
\end{align*}
with the convention that $\block{}{}{1:0} = \emptyset$. This still requires to enumerate all the permutations in $\mathcal{S}(\block{}{}{})$ which is factorial in the length of the blocks. However, the computation can be reduced through 
an algorithm reminiscent of dynamic programming as one can show that
\begin{align*}
    p(\block{}{}{l} | \lambda, \block{}{}{1:l-1}) & = \Big(\prod_{k\in\block{}{}{l} } \lambda_{k} \Big) \cdot \mathcal{R}_l({\block{}{}{l}}).
\end{align*}
Here, $\mathcal{R}_l(A)$ is a function that is defined recursively for any subset $A \subset \block{}{}{l}$ as
\begin{align*}
 \mathcal{R}_{l}(A) = \left\{ \begin{array}{cc} 
 1 & A = \emptyset \\
 \Big(\sum_{a \in A} \mathcal{R}_{l}(A \setminus \{a\})\Big)/{\Big(\bar{Z}_{l} + \sum_{a \in A} \lambda_a\Big)}  & A \neq \emptyset  
 \end{array} \right.  
\end{align*}
where $\bar{Z}_{l} = \sum_{k \in \bar{b}_l} \lambda_k$ denotes the total plausibility of the remaining elements $\bar{b}_l = \block{}{}{l+1} \cup \dots \cup \block{}{}{L}$. The probability of a partition is 
\begin{align*}
    p(\block{}{}{}|\plaus{}{}) & = \Pr\{\block{}{}{1} \succ \block{}{}{2} \succ \dots \succ \block{}{}{L}\} 
 = \Big( \prod_{k\notin b_L} \lambda_k \Big) \prod_{l=1}^{L-1} \mathcal{R}_l(b_l).
\end{align*}
We refer to \ref{app:proof} for a proof and an example, but illustrate the recursion in Figure \ref{fig:methods-recursion}.

\subsection{PL Gibbs sampling with partial rankings}
\label{subsec:methods-gibbs}
Following \citep{Caron2010}, we consider the following hierarchical Bayesian model for PL with partial rankings by introducing latent arrival times $\tau$:
\begin{equation}
\begin{aligned}
\label{eq:pl_generative_model}
\lambda_k & \sim \mathcal{G}(\alpha,\beta),\quad k = 1 \dots K\\
\tau_k & \sim \mathcal{E}(\lambda_k),\quad k = 1 \dots K \\
\sigma & = \arg\sort (\tau)  \\
B & = Q P_\sigma 
\end{aligned}
\end{equation}
where $\tau = \tau_{1:K}$, $\mathcal{G}(\alpha,\beta)$ is a Gamma distribution with shape $\alpha$ and rate $\beta$ and $\mathcal{E}(\rho)$ is an exponential distribution with rate $\rho$. Here, $P_\sigma$ denotes the $K \times K$ permutation matrix associated with $\sigma$ that would bring $\tau$ into an ascending sorted order\footnote{For exposition related to Equation \eqref{eq:pl_generative_model}, we assume the latent random variable $\choice{}{}{}$ to be a \textit{complete ranking}, rather than a \textit{full ranking}. As defined in Section~\ref{subsec:method-diff}, complete ranking implies that $\choice{}{}{}$ includes no ties or unranked classes.}.
$Q$ denotes a $L \times K$ matrix that represents the partition structure of the partial ranking and transforms a given permutation matrix $P_\choice{}{}{}$ to a matrix where each row represents the sets associated with each block. Finally, $B$ denotes a $L \times K$ matrix representation of the partial ranking $b$. For example, the partial ranking $\block{}{}{} = \{1, 5\} \succ  \{2, 3\} \succ \{4\}$ corresponds to
\begin{align*}
B & = 
\begin{pmatrix} 
1 & 0 & 0 & 0 & 1 \\
0 & 1 & 1 & 0 & 0 \\
0 & 0 & 0 & 1 & 0
\end{pmatrix}
&  Q & = 
\begin{pmatrix} 
1 & 1 & 0 & 0 & 0 \\
0 & 0 & 1 & 1 & 0 \\
0 & 0 & 0 & 0 & 1
\end{pmatrix}
\end{align*}
Hence for any permutation $\choice{}{}{} \in \mathcal{S}(\block{}{}{})$ that is consistent with $\block{}{}{}$, the associated permutation matrix $P_{\choice{}{}{}}$ satisfies the matrix equality
$B = Q P_{\choice{}{}{}}$. In other words, $P_{\choice{}{}{}}$ and $B$ are matrix representations for the random variables $\choice{}{}{}$ and $b$, respectively.

In this model, the goal is to construct a Markov chain with the target distribution $p(\lambda|B)$.
However, directly sampling form the desired posterior is difficult, thus we sample from an augmented target distribution $p(\lambda, \tau, P_\sigma| B)$ instead, extending \citep{Caron2010} to the case of partial rankings. Specifically, Gibbs sampling constructs such a Markov chain by iteratively sampling from full conditionals. In our model, the conditionals are
\begin{align*}
\lambda \sim  p(\lambda| \tau) & & \tau & \sim p(\tau| 
\lambda, P_\sigma) && P_\sigma \sim p(P_\sigma| \lambda, B)
\end{align*}
where $\lambda = \lambda_{1:K}$. In the sequel, we will derive each full conditional separately.

\textbf{Sampling from $p(\lambda| \tau)$:}
Each coordinate of $\lambda$ becomes conditionally independent from the rest, given the corresponding coordinate of $\tau$. The full conditional is obtained using the conjugacy of a gamma prior $p(\lambda_k)$ and exponential observation model $p(\tau_k| \lambda_k)$ as
\begin{align*}
    p(\lambda_k| \tau_k) & = \mathcal{G}(\lambda_k; \alpha+1,\beta+\tau_k).
\end{align*}

\textbf{Sampling from $p(\tau| \lambda, P_\sigma)$:}
Given $\lambda$ and $P_\sigma$, it is easier to generate $\tau_{\text{sorted}}$ rather than $\tau$ directly. The difficulty is that the arrival times must be drawn proportionally to $p(\tau | \lambda)$ while still satisfying the ranking prescribed by $P_\sigma$.
Fortunately, this is simplified by exploiting that the exponential distribution is memoryless. Specifically, define $\tilde{\lambda} = P_\sigma \lambda$ (note that $\tilde{\lambda}$ itself may not be sorted, it is just rearranged according to the permutation matrix $P_\sigma$). By the order statistics of the exponential distribution, the arrival time
of the winner is distributed by
\begin{align*}
p(\tau_{\text{sorted}, 1} |P_\sigma, \lambda) = \mathcal{E}(\sum_{k=1}^K \tilde{\lambda}_{ k}).
\end{align*}
Now, for the runner-up, we know that $\tau_{\text{sorted}, 2} > \tau_{\text{sorted}, 1}$ with probability one. Moreover, thanks to the memoryless property of the exponential distribution,  at time $\tau_{\text{sorted}, 1}$, when the winner has finished, the remaining race looks like a race between the remaining contenders (i.e., classes). 
Thus, the interarrival time for this second race is
$\Delta_2 = \tau_{\text{sorted}, 2}  - \tau_{\text{sorted}, 1} $ has the distribution
\begin{align*}
p(\Delta_2|
\tau_{\text{sorted}, 1} , P_\sigma, \lambda) 
= \mathcal{E}(\sum_{k=2}^K \tilde{\lambda}_{k}).
\end{align*}
For the $k$th position in the race, we see that the interarrival time 
$\Delta_k = \tau_{\text{sorted}, k}  - \tau_{\text{sorted}, k-1} $ has the distribution
\begin{align*}
p(\Delta_k| \tau_{\text{sorted}, 1:k-1} , P, \lambda) 
= \mathcal{E}(\sum_{j=k}^K \tilde{\lambda}_{j}).
\end{align*}
This observation provides the following procedure to draw from $p(\tau | P_\sigma, \lambda)$:
First, we express the cumulative sum operation using a $K \times K$ upper triangular matrix $U$ where the diagonal and upper diagonal entries are one and remaining elements are zero. Second, the exponential distribution with vector argument
$ x \sim  \mathcal{E}( v ) $ denotes $x_k \sim \mathcal{E}(v_k)$ for $k=1 \dots K$.
Overall, sampling boils down to:
\begin{enumerate}
\item Arranging the elements of $\lambda$ according to $P_\sigma$: $\tilde{\lambda} = P_\sigma \lambda$.
\item Sampling the interarrival times $\Delta \sim \mathcal{E}( U \tilde{\lambda})$.
\item Computing the sorted arrival times from the interarrival times $\tau_{\text{sorted}} = U^\top \Delta$.
\item Recovering the arrival times as $\tau = P_\sigma^\top \tau_{\text{sorted}}$.
\end{enumerate}
Compactly, this can be implemented with a one-liner: 
\begin{align}
\tau = P_\sigma^\top U^\top (-\log(\epsilon) / (U P_\sigma \lambda))
\label{eq:tau_sampler}
\end{align} 
where $\epsilon$ is a vector of length $K$ with uniform distributed entries on $[0, 1]$ and $/$ is element-wise division.

\textbf{Sampling from $p(P_\sigma| \lambda, B)$:}
The identity
\begin{align*}
p(P_\sigma| \lambda, B) = \frac{p(P_\sigma, B| \lambda)}{p(B| \lambda)} = \frac{p(B| P_\sigma) p( P_\sigma | \lambda)}{p(B| \lambda)}
\end{align*}
suggests that we need to sample from a permutation $\sigma$ from a Plackett-Luce distribution
with parameters $\lambda$ (second term of the nominator), but in such a way that $B = Q P_\sigma$ (first term of the nominator).
This can be achieved by sequentially sampling for each block a permutation using the following full conditionals:
\begin{align*}
    \sigma_1 & \sim p(\sigma_1 | \lambda, \block{}{}{1}) \\
    \sigma_2 & \sim p(\sigma_2| \lambda, \block{}{}{1},\sigma_1 ) \\
    \dots \\
    \sigma_k & \sim p(\sigma_k| \lambda, \block{}{}{1},\sigma_{1:k-1} )
\end{align*}
for $k = 1 \dots c_1$. The first full conditional is
\begin{align*}
p(\sigma_1 = s_1 | \lambda, \block{}{}{1}) & = \frac{p(\sigma_1 = s_1, \block{}{}{1} | \lambda)}{p(\block{}{}{1} | \lambda)} = \frac{\sum_{\sigma} \ind{\sigma_{1} = s_1} \ind{\sigma_{1:c_1} \in \mathcal{S}(\block{}{}{1})} p(\sigma | \lambda)}{\sum_{\sigma} \ind{\sigma_{1:c_1} \in \mathcal{S}(\block{}{}{1})} p(\sigma | \lambda)} \nonumber\\
& = \frac{\sum_{\sigma} \ind{\sigma_{1} = s_1} \ind{\sigma_{2:c_1} \in \mathcal{S}(\block{}{}{1} \setminus \{s_1\})} p(\sigma | \lambda)}{\sum_{s \in b_1}\sum_{\sigma} \ind{\sigma_{1} = s} \ind{\sigma_{2:c_1} \in \mathcal{S}(\block{}{}{1} \setminus \{s\})} p(\sigma | \lambda)}\nonumber \\
& = \frac{\mathcal{R}_1(\block{}{}{1} \setminus \{s_1\})}{\sum_{s \in b_1} \mathcal{R}_1(\block{}{}{1} \setminus \{s\})}
\end{align*}
It can be shown that, the $k$'th class can be sampled with probabilities
\begin{align*}
p(\sigma_k = s_k | \lambda, \block{}{}{1}, \sigma_{1:k-1} = s_{1:k-1}) = 
\frac{\mathcal{R}_1(\block{}{}{1} \setminus \{s_{1:k-1}, s_k\})}{\sum_{s \in b_1 \setminus \{s_{1:k-1}\}} \mathcal{R}_1(\block{}{}{1} \setminus \{s_{1:k-1}, s\})}.
\end{align*}
Again, this algorithm can be implemented using the diagram in Figure \ref{fig:methods-recursion}. On the first pass, we calculate the function $\mathcal{R}$ for each subset of the elements in the block. Then, starting from the top node, we sample first $\sigma_1$ according to $p(\sigma_1 | \lambda, \block{}{}{1})$. Then, we sample $\sigma_2$ from $p(\sigma_2 | \lambda, \block{}{}{1}, \sigma_1)$ where we move from node $b_1$ to $b_1 \setminus \{\sigma_1\}$ and so on. In a sense we randomly sample a path in the trellis of the Hasse diagram, where nodes are weighted according to $\mathcal{R}$.
To sample a full $\sigma$, this procedure is applied to all blocks $l = 1, \dots, L$ in a given partial ranking; such that for the $l$'th block $b_l$ we sample from $p(\sigma_k = s_k | \lambda, \block{}{}{1:l}, \sigma_{1:k-1} = s_{1:k-1})$, for $k = c_{l-1} + 1, \dots, c_l$. Note that we condition on $\block{}{}{1:l}$ instead of $\block{}{}{l}$ in this generic expression, since both $\lambda$ and $\block{}{}{1:l}$ are needed to compute $\mathcal{R}_l$.

\textbf{Summary:} Overall, our Gibbs sampling procedure would involve repeating the loop below for $T$ \textit{iteration}s, where $t \in \{1, \dots, T\}$ denote the specific iterations of the algorithm. We initialize $\lambda$ by sampling from its prior, such that $ \plaus{(0)}{j} \sim \mathcal{G}(\alpha,\beta), \forall j\in[\numcond]$. The $t$'th iteration of the algorithm would involve the following updates:
\begin{align*}
    P^{(t)}_\sigma &\sim p(P_\sigma|\plaus{(t-1)}{}, B)\\
    \tau^{(t)} &\sim p(\tau|\plaus{(t-1)}{}, P^{(t)}_\sigma)\\
    \plaus{(t)}{} &\sim p(\plaus{}{}|\tau^{(t)})
\end{align*}
for $t = 1, \dots, T$. After discarding an appropriate number of burn in samples $I$, the samples $(\plaus{(t)}{})^{T}_{t=I}$ constitute the samples from the posterior of $p(\plaus{}{}|b)$.

\textbf{Extension to multiple annotators:} We now address the case where we have multiple annotators. For $R$ annotators we let $b = (\block{}{}{}^1, \dots, \block{}{}{}^R)$, and denote the rankings of the annotator $r$ with $\block{}{}{}^r = (\block{}{}{1}^r, \dots, \block{}{}{L_r}^r)$, where $L_r$ denotes the number of blocks for annotator $r$.
While likelihood computation can be done independently across annotators, i.e. $p(\block{}{}{}|\lambda) = \prod^{R}_{r=1} p(\block{}{}{}^r|\lambda)$, the generative model of Equation \eqref{eq:pl_generative_model} and the associated Gibbs sampling algorithm needs to be be modified to accommodate multiple annotators. We restate the generative model model as
\begin{alignat*}{2}
\label{eq:pl_generative_model_multiple_readers}
\lambda_k & \sim \mathcal{G}(\alpha,\beta),\quad &&k = 1 \dots K\\
\tau^r_k & \sim \mathcal{E}(\lambda_k),\quad &&k = 1 \dots K,\, r = 1 \dots R  \\
\sigma^r & = \arg\sort (\tau^r),\quad &&r = 1 \dots R  \\
B^r &= Q^r P_{\sigma^r},\quad &&r = 1 \dots R,
\end{alignat*}
where we redefine the random variables without superscripts as $\tau = (\tau^1, \dots, \tau^R)$, $B = (B^1, \dots, B^R)$, $\sigma = (\sigma^1, \dots, \sigma^R)$, and $Q = (Q^1, \dots, Q^R)$. We now describe the changes to the sampling steps. In this extended model, sampling from $p(\lambda| \tau)$ is defined as
\begin{align*}
    p(\lambda_k| \tau_k) & = \mathcal{G}(\lambda_k; \alpha+n_k, \beta+\sum^R_{r=1}\tau^r_k),
\end{align*}
where $n_k$ is the number of times the item $k$ is observed among partial rankings, excluding the last blocks.
Modification needed for the steps for sampling $p(\tau| \lambda, P_\sigma)$ and $p(P_\sigma| \lambda, B)$ to address multiple annotators are even simpler, given the conditional independences implied by the graphical model. For both of these cases, we repeat the steps described above independently for each annotator. More explicitly:
\begin{align*}
    p(\tau| \lambda, P_\sigma) = \prod^R_{r=1}p(\tau^r| \lambda, P_{\sigma^r}), \qquad
    p(P_\sigma| \lambda, B) = \prod^R_{r=1}p(P_{\sigma^r}| \lambda, B^r).
\end{align*}
The algorithm is otherwise identical to the single annotator case described above.

\subsection{Likelihood computation for partial rankings}
\label{app:proof}

\textbf{Example:} Given a PL model $\lambda = (\lambda_1, \lambda_2, \lambda_3, \dots, \lambda_{K})$, 
assume that we wish to compute the probability of the partial ranking
$$ \{1, 2, 3\} \succ \{4, \dots, K\}. $$
This is the total probability of all permutations of the form
\begin{align*}
    \sigma = \sigma_{1:3} * \sigma_{4:K}
\end{align*}
where $\sigma_{1:3} \in \mathcal{S}(b_1)$, $\sigma_{4:K} \in \mathcal{S}( \bar{b}_1 )$ and $*$ denotes concatenation.
We set $b_1 = \{1, 2, 3\}$ and $\bar{b}_1 = \{4, \dots, K\}$. We let $\bar{Z}_1 = \lambda_4 + \dots + \lambda_K$.
The probability of a permutation under the PL distribution is 
\begin{align*}
    p(\sigma_{1:3} * \sigma_{4:K}) = \frac{\lambda_{\sigma_1}}{(\bar{Z}_1 + \lambda_{\sigma_1} + \lambda_{\sigma_2} + \lambda_{\sigma_3})} 
\frac{\lambda_{\sigma_2}}{(\bar{Z}_1 + \lambda_{\sigma_2} + \lambda_{\sigma_3})} \frac{\lambda_{\sigma_3}}{(\bar{Z}_1 + \lambda_{\sigma_3})}
\frac{\lambda_{\sigma_4}}{\bar{Z}_1} \frac{\lambda_{\sigma_5}}{(\bar{Z}_1 - \lambda_{\sigma_4})} \dots \frac{\lambda_{\sigma_K}}{(\bar{Z}_1 - \sum_{k=4}^{K-1} \lambda_{\sigma_k})},
\end{align*}
\begin{align*}
  p(b_1 \succ \bar{b}_1) & =   \sum_{\sigma_{1:3} \in \mathcal{S}(b_1)} \frac{\lambda_{\sigma_1}}{ (\bar{Z}_1 + \lambda_{\sigma_1} + \lambda_{\sigma_2} + \lambda_{\sigma_3})} 
\frac{\lambda_{\sigma_2}}{(\bar{Z}_1 + \lambda_{\sigma_2} + \lambda_{\sigma_3})} \frac{\lambda_{\sigma_3}}{(\bar{Z}_1 + \lambda_{\sigma_3})}
\sum_{\sigma_{4:K} \in \mathcal{S}(b_2)} \frac{\lambda_{\sigma_4}}{\bar{Z}_1}  \dots \frac{\lambda_{\sigma_K}}{(\bar{Z}_1 - \sum_{k=4}^{K-1} \lambda_{\sigma_k})}. 
\end{align*}
The second term is the total probability of a Plackett--Luce distribution on $\bar{b}_1$ so it is equal to one, thus can be ignored. We can rearrange the first term as below:
\begin{align*}
    p(b_1 \succ \bar{b}_1) & =   \sum_{\sigma_{1:3} \in \mathcal{S}(b_1)} \frac{\lambda_{\sigma_1}}{(\bar{Z}_1 + \lambda_{\sigma_1} + \lambda_{\sigma_2} + \lambda_{\sigma_3})} 
\frac{\lambda_{\sigma_2}}{(\bar{Z}_1 + \lambda_{\sigma_2} + \lambda_{\sigma_3})} \frac{\lambda_{\sigma_3}}{(\bar{Z}_1 + \lambda_{\sigma_3})} \nonumber\\
 & = \frac{\lambda_1 \lambda_2 \lambda_3}{(\bar{Z}_1 + \lambda_1 + \lambda_2 + \lambda_3)}  \Big( \frac{1}{(\bar{Z}_1 + \lambda_2 + \lambda_3)} \Big( \frac{1}{(\bar{Z}_1 + \lambda_3)} +  \frac{1}{(\bar{Z}_1 + \lambda_2)} \Big)\nonumber \\
 & + \frac{1}{(\bar{Z}_1 + \lambda_1 + \lambda_3)} \Big(\frac{1}{(\bar{Z}_1 + \lambda_3)} 
+  \frac{1}{(\bar{Z}_1 + \lambda_1)} \Big)
 + \frac{1}{(\bar{Z}_1 + \lambda_1 + \lambda_2)} \Big( \frac{1}{(\bar{Z}_1 + \lambda_2)} 
 + \frac{1}{(\bar{Z}_1 + \lambda_1)} \Big) \Big). 
\end{align*}
We see that the desired probability is 
\begin{align*}
  p(b_1 \succ \bar{b}_1) & =  (\lambda_1 \lambda_2 \lambda_3) \times \mathcal{R}_1({\{1, 2, 3\}}),
\end{align*}
where $\mathcal{R}_1$ is a function that is defined on all subsets of $b_1$ recursively as follows
\begin{align*}
\mathcal{R}_1({\{1, 2, 3\}}) & = \frac{\mathcal{R}_1({\{1,2\}}) + \mathcal{R}_1({\{1,3\}}) + \mathcal{R}_1(\{2,3\})}{(\bar{Z}_1 +\lambda_1 + \lambda_2 + \lambda_3)},\\
\mathcal{R}_1({\{1, 2\}}) & = \frac{\mathcal{R}_1({\{1\}}) + \mathcal{R}_1({\{2\}})}{(\bar{Z}_1 +\lambda_1 + \lambda_2)}  \quad
\mathcal{R}_1({\{1, 3\}})  = \frac{\mathcal{R}_1({\{1\}}) + \mathcal{R}_1({\{3\}})}{(\bar{Z}_1 +\lambda_1 + \lambda_3)}\quad
\mathcal{R}_1({\{2, 3\}}) & = \frac{\mathcal{R}_1({\{2\}}) + \mathcal{R}_1({\{3\}})}{(\bar{Z}_1 +\lambda_2 + \lambda_3)},\\
\mathcal{R}_1({\{1\}}) & = \frac{1}{(\bar{Z}_1 +\lambda_1)}\quad
\mathcal{R}_1({\{2\}})  = \frac{1}{(\bar{Z}_1 +\lambda_2)}\quad
\mathcal{R}_1({\{3\}})  = \frac{1}{(\bar{Z}_1 +\lambda_3)}. 
\end{align*}

\textit{\textbf{Theorem:}
Given a PL model with plausibilities $\lambda = (\lambda_1, \dots, \lambda_K)$, and a partitioning $b = (\block{}{}{1}, \block{}{}{2}, \dots, \block{}{}{L})$
of $K$ distinct classes, we say that a permutation of $K$ elements is compatible with the partitioning $b$ when $\sigma_{1:c_L} = \sigma_{1:c_1} * \sigma_{c_1+1:c_2} * \dots * \sigma_{c_{L-1}+1:c_L}$
satisfies $\sigma_{1:c_1} \in \mathcal{S}(b_1)$, $\sigma_{c_1+1:c_2} \in \mathcal{S}(b_2)$,
$\dots$, $\sigma_{c_{L-1}+1:c_L} \in \mathcal{S}(b_L)$.
The probability of observing a permutation that is compatible with the partitioning $b$ given $\lambda$ is
\begin{align*}
    p(\block{}{}{}|\plaus{}{}) & = p(\block{}{}{1} \succ \block{}{}{2} \succ \dots \succ \block{}{}{L}) = p(\block{}{}{1}| \lambda) p(\block{}{}{2}| \lambda, \block{}{}{1} ) \dots p(\block{}{}{L}| \lambda, \block{}{}{1:L-1}) = \Big( \prod_{k=1}^{c_L} \lambda_k \Big) \prod_{l=1}^L \mathcal{R}_l(b_l)
\end{align*}
where
\begin{align*}
    p(\block{}{}{l} | \lambda, \block{}{}{1:l-1}) & = \Big(\prod_{k\in\block{}{}{l} } \lambda_{k} \Big) \cdot \mathcal{R}_l({\block{}{}{l}}).
\end{align*}
Here, $\mathcal{R}_l(A)$ is a function that is defined recursively for any subset $A \subset \block{}{}{l}$ as
\begin{align*}
 \mathcal{R}_{l}(A) = \left\{ \begin{array}{cc} 
 1 & A = \emptyset \\
 \Big(\sum_{a \in A} \mathcal{R}_{l}(A \setminus \{a\})\Big)/{\Big(\bar{Z}_{l} + \sum_{a \in A} \lambda_a\Big)}  & A \neq \emptyset  
 \end{array} \right.  
\end{align*}
where $\bar{Z}_{l} = \sum_{k \in \bar{b}_l} \lambda_k$ denotes the total plausibility of the remaining elements $\bar{b}_l \equiv \{\block{}{}{l+1} \cup \dots \cup \block{}{}{L}\}$.
}

{\small
\textbf{Proof:}
We first prove the base case for $l=1$, where we will calculate
\begin{align*}
p(b_1 | \lambda) = p(b_1 \succ \bar{b}_1)
\end{align*}
with $\bar{b}_1 = \block{}{}{2} \cup \dots \cup \block{}{}{L}$.
The structure of the computation will be identical for $l > 1$, where $\bar{b}_l = \block{}{}{l+1} \cup \dots \cup \block{}{}{L}$
and $
p(\block{}{}{l} | \lambda, \block{}{}{1:l-1}) = p(b_l \succ \bar{b}_l | b_1 \succ \dots \succ b_{l-1} \succ b_l \cup \bar{b}_l)
$. For $l = 1$, $p(b_1 \succ \bar{b}_1)$ is the total probability of all permutations of the form
\begin{align*}
    \sigma = \sigma_{1:c_1} * \sigma_{c_1+1:c_L}
\end{align*}
where $\sigma_{1:c_1} \in \mathcal{S}(b_1)$, $\sigma_{c_1+1:c_L} \in \mathcal{S}(\bar{b}_1)$, $*$ denotes concatenation and $c_L = K$. Under the PL distribution, a given permutation has the following probability
\begin{align*}
    p(\sigma_{1:c_1} * \sigma_{c_1+1:K}) = 
    \Big(
    \prod_{k = 1}^{c_1} \frac{\lambda_{\sigma_k}}{(\bar{Z}_{1} + \sum_{r \in b_1 \setminus \{\sigma_{1:k-1}\}} \lambda_{r})} 
    \Big)
    \Big(
    \prod_{k=c_1+1}^{K}\frac{\lambda_{\sigma_k}}{(\bar{Z}_{1} - \sum_{r \in \{\sigma_{c_1+1:k-1}\}} \lambda_{r})} 
    \Big)
\end{align*}
where $\bar{Z}_{1} = \sum_{k \in \bar{b}_1} \lambda_k$. The probability we wish to calculate is
\begin{align*}
\begin{split}
    p(b_1 \succ \bar{b}_1) &= \sum_{\sigma_{1:c_1} \in \mathcal{S}(b_1)} \sum_{\sigma_{c_1+1:c_L} \in \mathcal{S}(\bar{b}_1)} p(\sigma_{1:c_1} * \sigma_{c_1+1:K})\\
    & = \sum_{\sigma_{1:c_1} \in \mathcal{S}(b_1)}
        \Big(
    \prod_{k=1}^{c_1} \frac{\lambda_{\sigma_k}}{(\bar{Z}_1 + \sum_{r \in b_1 \setminus \{\sigma_{1:k-1}\}} \lambda_{r})} 
    \Big)
    \sum_{\sigma_{c_1+1:c_L} \in \mathcal{S}(\bar{b}_1)}
    \Big(
    \prod_{k=c_1+1}^{c_L} \frac{\lambda_{\sigma_k}}{(\bar{Z}_l - \sum_{r \in \bar{b}_1 \setminus \{\sigma_{c_1+1:k-1}\}} \lambda_{r})}. 
    \Big)
\end{split}
\end{align*}
The second term is the total probability of a PL distribution on classes $k \in \bar{b}_1$ with plausibilities $\lambda_k$, so it is identical to one. Hence, we have
\begin{align*}
\begin{split}
p(b_1 \succ \bar{b}_1) & = \sum_{\sigma_{1:c_1} \in \mathcal{S}(b_1)}
    \Big(\prod_{k=1}^{c_1} \lambda_{\sigma_k} \Big) \Big( \prod_{k=1}^{c_1} \frac{1}{(\bar{Z}_1 + \sum_{r \in b_1 \setminus \{\sigma_{1:k-1}\}} \lambda_{r})} 
    \Big)\\
& = \Big(\prod_{k = 1}^{c_1} \lambda_{k} \Big) \sum_{\sigma_{1:c_1} \in \mathcal{S}(b_1)}
     \Big( \prod_{k=1}^{c_1} \frac{1}{(\bar{Z}_1 + \sum_{r \in b_1 \setminus \{\sigma_{1:k-1}\}} \lambda_{r})} 
    \Big)  \equiv \Big(\prod_{k=1}^{c_1} \lambda_{k} \Big) \cdot \mathcal{R}_1(b_1),
\end{split}\\
 \mathcal{R}_1(b_1) & \equiv  \sum_{\sigma_{1:c_1} \in \mathcal{S}(b_1)}
     \Big( \prod_{k=1}^{c_1} \frac{1}{(\bar{Z}_1 + \sum_{r \in b_1 \setminus \{\sigma_{1:k-1}\}} \lambda_{r})}
    \Big).
\end{align*}
Now, the function $\mathcal{R}_1(b_1)$ can be evaluated by reorganizing the individual terms as
\begin{align*}
\begin{split}
    \mathcal{R}_1(b_1) & = \frac{1}{(\bar{Z}_1 + \sum_{r \in b_1} \lambda_{r})} 
    \sum_{\sigma_{1} \in b_1} 
      \frac{1}{(\bar{Z}_1 + \sum_{r \in b_1 \setminus \{\sigma_{1}\}} \lambda_{r})} 
    \sum_{\sigma_{2} \in b_1\setminus \{\sigma_1\}} 
      \frac{1}{(\bar{Z}_1 + \sum_{r \in b_1 \setminus \{\sigma_{1:2}\}} \lambda_{r})} \\
      & \times 
    \sum_{\sigma_{3} \in b_1\setminus \{\sigma_{1:2}\}} 
      \frac{1}{(\bar{Z}_1 + \sum_{r \in b_1 \setminus \{\sigma_{1:3}\}} \lambda_{r})} 
      \dots \\
      & \dots 
    \sum_{\sigma_{c_1-2} \in b_1\setminus \{\sigma_{1:{c_1-3}}\}} 
      \frac{1}{(\bar{Z}_1 + \sum_{r \in b_1 \setminus \{\sigma_{1:c_1-2}\}} \lambda_{r})} 
    \sum_{\sigma_{c_1-1} \in b_1\setminus \{\sigma_{1:{c_1-2}}\}} 
      \frac{1}{(\bar{Z}_1 + \sum_{r \in b_1 \setminus \{\sigma_{1:c_1-1}\}} \lambda_{r})}.
\end{split}
\end{align*}
We see that all the required computations depend on sets of the form $b_1 \setminus \{\sigma_{1:k}\}$ where $k=1, \dots c_1-1$ and not on the individual permutations. This observation allows us to reduce the computation from evaluating $|b_1|!$ terms to evaluating the function $\mathcal{R}_1$ on all subsets of $b_1$, of which there are $2^{|b_1|}$ (and we use $w^{b_1}$ to denote the power set of $b_1$). To see this, notice that
\begin{align*}
    \mathcal{R}_1(b_1) & = \frac{1}{(\bar{Z}_1 + \sum_{r \in b_1} \lambda_{r})} 
    \sum_{\sigma_{1} \in b_1} \mathcal{R}_1(b_1 \setminus \{\sigma_{1}\}) \\
    \mathcal{R}_1(b_1 \setminus \{\sigma_{1}\}) & =  \frac{1}{(\bar{Z}_1 + \sum_{r \in b_1 \setminus \{\sigma_{1}\}} \lambda_{r})} 
    \sum_{\sigma_{2} \in b_1\setminus \{\sigma_1\}} \mathcal{R}_1(b_1 \setminus \{\sigma_{1:2}\})\\
    \mathcal{R}_1(b_1 \setminus \{\sigma_{1:2}\}) & =  
    \frac{1}{(\bar{Z}_1 + \sum_{r \in b_1 \setminus \{\sigma_{1:2}\}} \lambda_{r})} 
    \sum_{\sigma_{3} \in b_1\setminus \{\sigma_{1:2}\}} 
    \mathcal{R}_1(b_1 \setminus \{\sigma_{1:3}\}) \\
    \dots \\
    \mathcal{R}_1(b_1 \setminus \{\sigma_{1:c_1-2}\}) & =
    \frac{1}{(\bar{Z}_1 + \sum_{r \in b_1 \setminus \{\sigma_{1:c_1-2}\}} \lambda_{r})} 
    \sum_{\sigma_{c_1-1} \in b_1\setminus \{\sigma_{1:{c_1-2}}\}} \mathcal{R}_1(b_1 \setminus \{\sigma_{1:c_1-1}\})  \\
    \mathcal{R}_1(b_1 \setminus \{\sigma_{1:c_1-1}\}) & =    
      \frac{1}{(\bar{Z}_1 + \sum_{r \in b_1 \setminus \{\sigma_{1:c_1-1}\}} \lambda_{r})}. 
\end{align*}
Hence $\mathcal{R}$ is a function that satisfies the following relation for any $A \subset b_1$:
\begin{align}
 \mathcal{R}_1(A) = \left\{ \begin{array}{cc} 
 1 & A = \emptyset, \\
 \Big({\sum_{a \in A} \mathcal{R}_1({A \setminus \{a\}}}) \Big)/{(\bar{Z}_1 + \sum_{a \in A} \lambda_a)} 
 & A \neq \emptyset,~ A \in w^{b_1}.
 \end{array} \right.   
\end{align}
For the $l$'th term, we will compute $\mathcal{R}_l(A)$ for all subsets $A$ of $b_l$ where $\bar{Z}_{l} = \sum_{k \in \bar{b}_l} \lambda_k$ denotes the total plausibility of the remaining elements $\bar{b}_l = \block{}{}{l+1} \cup \dots \cup \block{}{}{L}$:
\begin{align*}
 \mathcal{R}_l(A) = \left\{ \begin{array}{cc} 
 1 & A = \emptyset, \\
 \Big({\sum_{a \in A} \mathcal{R}_l({A \setminus \{a\}}}) \Big)/{(\bar{Z}_l + \sum_{a \in A} \lambda_a)} 
 & A \neq \emptyset, ~ A \in w^{b_l}.
 \end{array} \right.   
\end{align*}
The total probability of the partition is given by
\begin{align*}
    p(\block{}{}{}|\plaus{}{}) & = p(\block{}{}{1} \succ \block{}{}{2} \succ \dots \succ \block{}{}{L})
    = p(\block{}{}{1}| \lambda) p(\block{}{}{2}| \lambda, \block{}{}{1} ) \dots p(\block{}{}{L}| \lambda, \block{}{}{1:L-1})
    = \Big( \prod_{k=1}^{c_L} \lambda_k \Big) \prod_{l=1}^L \mathcal{R}_l(b_l).
\end{align*}
as stated in the theorem.

Note that we assume $b$ to be a partitioning of $K$ classes, i.e. $c_L = K$, which implies $( \prod_{k\in b_L} \lambda_k ) \mathcal{R}_l(b_l) = 1$. In some practical applications, the last block $L$ might collect a possibly large number of ``unranked'' classes as explained in Section~\ref{subsec:method-diff}. Therefore in such cases we can avoid some potentially expensive computation for the last block since we can say that $p(\block{}{}{}|\plaus{}{}) =( \prod_{k\notin b_L} \lambda_k ) \prod_{l=1}^{L-1} \mathcal{R}_l(b_l) = ( \prod_{k=1}^{c_L} \lambda_k ) \prod_{l=1}^L \mathcal{R}_l(b_l)$.
}

\section{Average overlap with partial rankings}
\label{appendix-average-overlap}

In this section, we extend the average overlap definition of Section \ref{subsec:methods-measuring} to partial rankings. We first define average overlap (at $L$) between two full rankings 
 $\sigma = (\choice{}{}{1},\ldots,\choice{}{}{K})$ and $\sigma' = (\choice{}{}{1}',\ldots,\choice{}{}{K}')$ as
\begin{align*}
    \text{AverageOverlap@L}(\sigma, \sigma') =\frac{1}{L} \sum_{k=1}^L \text{Overlap}(\{\sigma_{1:k}\}, \{\sigma'_{1:k}\}),
\end{align*}
where we recall that overlap, defined in Equation \eqref{eq:methods-overlap} and repeated here for convenience, is given by 
\begin{align*}
\text{Overlap}(C, Y) = \frac{|C \cap Y|}{|C|}.
\end{align*}
The average overlap metric is higher when the overlap occurs between top ranks rather than lower ones; 
a property that has been found to be useful when comparing differential diagnoses \citep{EngBJD2019}.
We can also express average overlap using matrix notation:
\begin{align*}
    \text{AverageOverlap@L}(\sigma, \sigma') = \trace\{(T {P}_{\sigma'})^\top D_L (T {P}_{\sigma})  \},
    \label{eq:methods-average-overlap-matrix-multiplication}
\end{align*}
where $D_L$ is a diagonal $K \times K$ matrix with diagonal entries
\begin{align*}
    [\nicefrac{1}{L}, \nicefrac{1}{2 L}, \nicefrac{1}{3 L}, ..., \nicefrac{1}{L^2}, 0, 0, ..., 0],
\end{align*}
$T$ is a lower triangular  $K \times K$ matrix with ones in the diagonal and below, and $P_\sigma$ is a $K \times K$  matrix representation of the corresponding permutation $\sigma$ defined as 
\begin{align*}
    [P_\sigma]_{i, j} = 
    \begin{dcases}
        1   & \text{if } \sigma_i = j,\\
        0   & \text{otherwise,}
    \end{dcases}
\end{align*}
paralleling its usage in Section~\ref{subsec:methods-gibbs}. For example, the permutation $\sigma = (4, 3, 1, 2)$ is represented by
\begin{align*}
    P_{(4, 3, 1, 2)} = 
    \begin{pmatrix}
        0   & 0 & 0 & 1 \\
        0   & 0 & 1 & 0 \\
        1   & 0 & 0 & 0 \\
        0   & 1 & 0 & 0
    \end{pmatrix}.
\end{align*}
Our goal is to compute average overlap between two partial rankings of $K$ classes.
To this end, we will replace the permutation matrices with \emph{soft} permutation matrices. Additionally we normalize the metric to ensure that average overlap between two identical partial rankings is 1. 

Given the matrix representation of a partial ranking $b$ as $L \times K$ matrix $B$ where $K$ classes are assigned to $L$ blocks, we define
\begin{align*}
\tilde{P}_B = P_\sigma B^\top \diag(1/c) B,
\end{align*}
where $c$ is a vector of row sums of $B$ ($c = B \mathbf{1}_K $).
The matrix $\tilde{P}_B$ is the expected permutation matrix over all permutation matrices consistent with a partial ranking, 
assuming each permutation of the elements of block have the same probability. In other words, element $(i, j)$ in $\tilde{P}_B$ is the  probability of class $j$ being in position $i$ according to the partial ranking $b$. 
For an example, we can express a partial ranking $b = \{4\} \succ \{3, 1\} \succ \{2\}$ using matrix $B$ and the associated soft permutation matrix is $\tilde{P}_B$
\begin{align*}
    B = \begin{pmatrix}
        0   & 0 & 0 & 1 \\
        1   & 0 & 1 & 0 \\
        0   & 1 & 0 & 0
    \end{pmatrix} & &
    \tilde{P}_{B} = 
    \begin{pmatrix}
        0   & 0 & 0 & 1 \\
        0.5   & 0 & 0.5 & 0 \\
        0.5   & 0 & 0.5 & 0 \\
        0   & 1 & 0 & 0
    \end{pmatrix} = P_\sigma B^\top \diag([1, 1/2, 1]) B,
\end{align*}
where $P_\sigma$ is any hard permutation matrix that is compatible with the partial ranking $b$.
Given the soft permutation matrix representation of partitioned partial rankings, we first define the unnormalized average overlap between two partial rankings
\begin{align*}
    \text{UAO@L}(b, b') = \trace\{(T \tilde{P}_{B'})^\top D_L (T \tilde{P}_{B})  \}.
\end{align*}
We then normalize to ensure that our new metric is at most 1: 
\begin{align*}
    \text{meanAO@L}(b, b') = \frac{\text{UAO@L}(b, b')}{\sqrt{\text{UAO@L}(b, b)\times\text{UAO@L}(b', b')}}.
\end{align*}
It can be checked that $\text{meanAO@L}(b, b) = 1$.

\section{Case study details}
\label{sec:app-data}

\begin{figure*}
    \centering
    \includegraphics[width=0.24\textwidth]{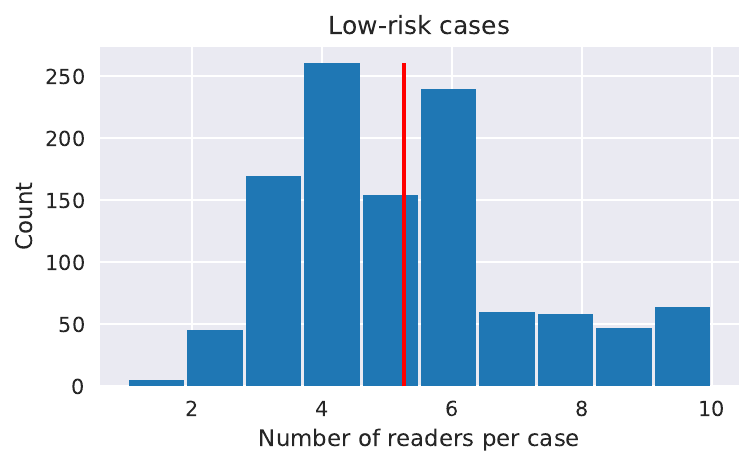}
    \includegraphics[width=0.24\textwidth]{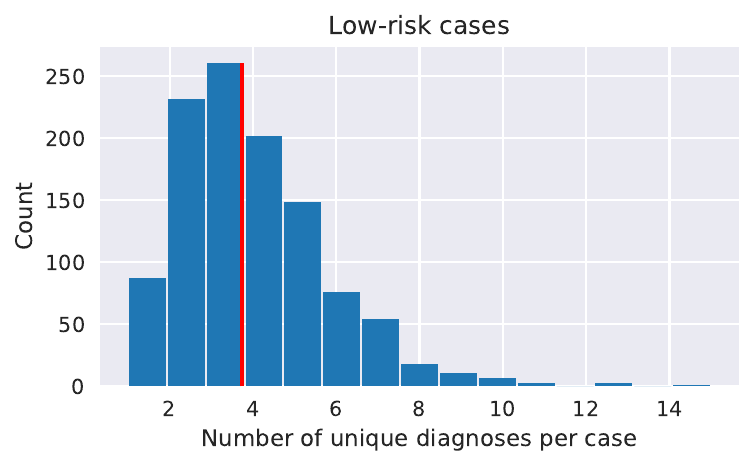}
    \includegraphics[width=0.24\textwidth]{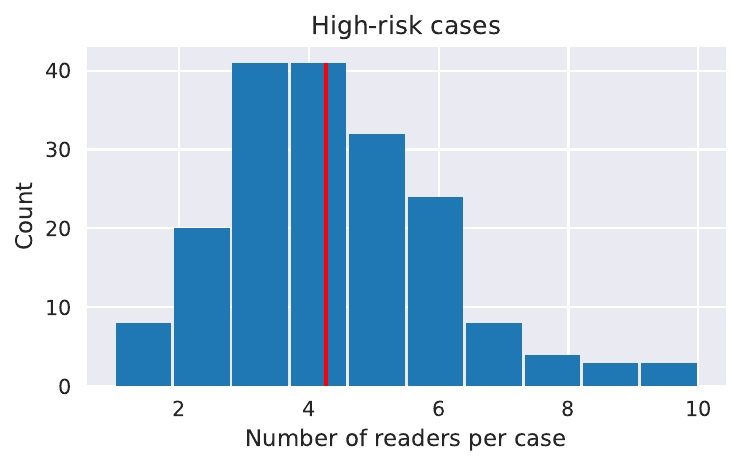}
    \includegraphics[width=0.24\textwidth]{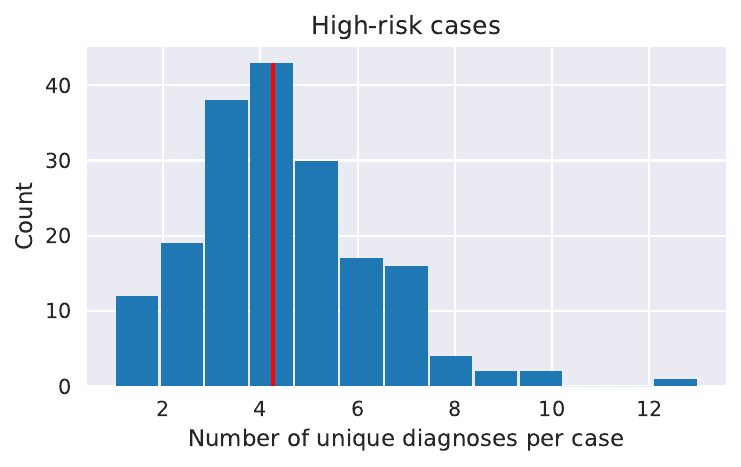}
    \caption{We plot the number of annotators per case and the number of unique conditions annotated by these annotators for low- (left) and high-risk (right) cases. Risk categories are based on the risk-level of the top-1 IRN condition as in previous work \citep{RoyMIA2022}. Surprisingly, high-risk cases were annotated by, on average, less annotators while clearly yielding more unique conditions.}
    \label{fig:app-reader-statistics}
\end{figure*}

We use the de-identified dermatology dataset previously used in~\citep{LiuNATURE2020} which was collected from $17$ different sites across California and Hawaii.
Each case in the dataset consists of up to $3$ RGB images, taken by medical assistants using consumer-grade digital cameras in a clinical environment. The images exhibit a large amount of variation in terms of affected anatomic location, background objects, resolution, perspective and lighting. 
Our dataset has patients with different skin types varying between Fitzpatrick scale of $1-6$.
We resized each image to $448\times448$ pixels for our training and evaluation purposes. The dataset consists of $16,225$ cases, where $12,335$ cases are used for model training, $1951$ cases for validation and $1939$ cases for test set. Note that in our dataset there are a few patients with multiple cases collected during repeated visits over the duration of the data collection process. For the data split, we ensured a patient-level split to avoid any bias.
Furthermore, we ensured a similar distribution of skin conditions and risk categories. The latter corresponds to a classification of case outcomes for a given skin condition, i.e., each skin condition has been mapped to a risk category in low, medium and high.
We operate on a label space of $419$ skin conditions similar to~\citep{LiuNATURE2020}, following a long-tailed distribution.
The dataset was labelled by US or Indian board certified dermatologists with $5-30$ years of experience.
Annotation was conducted over a period of time in tandem to data collection between $2010-2018$. As a result, different cases were annotated by different dermatologists (see Figure \ref{fig:app-reader-statistics} for statistics). Moreover, there are varying numbers of annotations per case but we made sure to have at least $3$ annotations per case.

\subsection{Annotation details}

All dermatologists were actively seeing patients in clinics and passed a certification test to ensure they were comfortable grading cases in our labeling tool. To provide the differential diagnoses, each annotator selected one or multiple conditions from the standardized Systematized Nomenclature of Medicine-Clinical Terms (SNOMED-CT)\footnote{\url{http://snomed.org/}} using a search-as-you-type interface, and assigned a confidence rating per condition. The conditions were later mapped to the 419 conditions as described previously \citep{LiuNATURE2020}. The confidence ratings were then used to rank conditions since confidence assignments are not comparable across dermatologists. As multiple conditions can obtain the same confidence ratings, this results in the \emph{partial} rankings described in Section \ref{subsec:method-diff}. These rankings are the basis for our study and several prior works \citep{EngBJD2019,RoyMIA2022,AziziARXIV2022}.
As the associated risk levels are of particular interest for analysis in prior work \citep{RoyMIA2022}, it is important to highlight that different annotator annotations also correspond to different risk annotations. Note that risk is not specific to each case but rather based on a fixed condition to risk mapping.

\subsection{Training details}
\label{app:training}

As we have multiple images per case, we are dealing with a multi-instance classification task with $419$ labels.
As we are dealing with limited training data, we use the BiT-L model~\citep{KolesnikovECCV2020}, pre-trained on JFT dataset~\citep{SunICCV2017} containing large-scale natural images from the internet. 
The encoder provides a feature representation for each image instance in a case. This output is then passed through an instance level average pooling layer, which generates a common feature-map for all the instances for a given case. 
This is passed to a classification head through a drop-out layer. The classification head has an intermediate fully-connected hidden layer with $1024$-dimensions followed by a final classification with a softmax layer which provide probabilities for each of the $419$ classes.
To keep model training process consistent to prior work~\citep{LiuNATURE2020, RoyMIA2022}, we use IRN as (soft) supervision in a cross-entropy loss.
For data augmentation, we use horizontal and vertical flips, rotations, variations of brightness, contrast and saturation, and random Gaussian blurring.
We use a Adam optimizer for training. Each model is trained for $10,000$ steps. Convergence of all the models were ensured on the validation set. We use a batch-size of $16$ cases for training.
For each model, we select the final checkpoint based on the highest top-3 accuracy against the top-1 IRN ground truth label, consistent with previous work.
We train a pool of 30 models by randomly varying the size of the ResNet architecture (depth $\times$ width: $101 \times 3$, $101 \times 1$, $50 \times 3$, $152 \times 2$, $152 \times 4$), pre-logit dropout rate (keep probability: $0.5 - 0.7$) and learning rate ($10^{-6} - 5\times10^{-4}$). As model performance is not the focus of this work, we randomly selected four of these models for our experiments.

\subsection{PL and \PIRN details}

As detailed in Section \ref{subsec:methods-pl}, for PL, we repeat annotators in order to simulate higher reliability. Specifically, corresponding to the x-axis in Figure \ref{fig:results-comparison}, we evaluate 1, 2, 3, 5, and 10 repetitions. For \PIRN, we evaluate temperatures $\gamma = 10, 20, 30, 50, 100$.

\section{Additional results}
\label{sec:app-results}

\begin{figure*}[t]
    \centering
    \includegraphics[height=3cm]{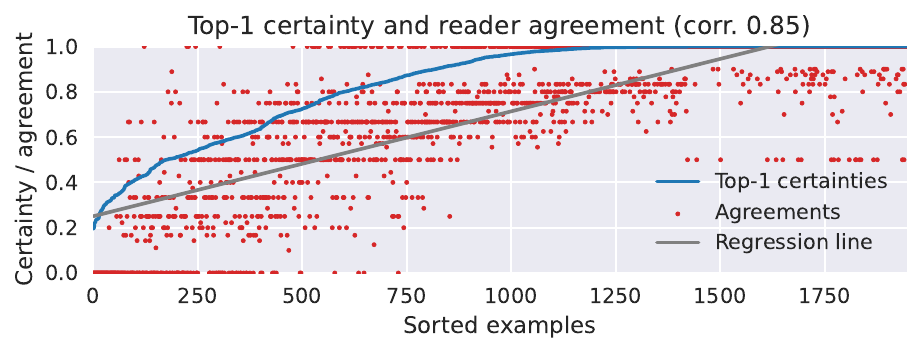}
    \caption{PL annotation certainty ({\color{blue}blue}) at medium reliability (see Figure \ref{fig:results-certainty}) plotted with annotator agreement ({\color{red}red}), see text for details. With a correlation coefficient of $0.85$, both measures are highly correlated, indicating that annotation certainty indeed captures the uncertainty induced by disagreement.}
    \label{fig:app-results-certainty}
\end{figure*}
\textbf{PL top-1 annotation certainty and agreement:} Figure \ref{fig:app-results-certainty} correlates top-1 annotation certainty from PL ({\color{blue}blue}, at medium reliability corresponding to Figure \ref{fig:results-certainty}) with annotator agreement ({\color{red}red}). Here, agreement is computed as follows: For $R$ annotators, given a selected annotator $r$, we compute IRN plausibilities using all annotator opinions except those from $r$. We then check whether the top-1 IRN label is included in the conditions provided by annotator $r$. We then average across annotators $r \in [R]$. More formally:
\begin{align*}
    \frac{1}{R}\sum_{r = 1}^R\delta[\arg\max_{k \in [K]} \text{IRN}(k;b^1,\ldots,b^{r-1},b^{r+1},\ldots, b^R) \in \cup_{l=1}^{L_r-1}b^r_l],
\end{align*}
where $L_r$ is the number of blocks in annotation provided by $r$, and we ignore $b^r_{L_r}$ since we use the last block to collect all unranked classes.
We found high correlation between this notion of annotator agreement and our annotation certainty, as also highlighted by the {\color{gray}gray} regression line. This further highlights that our statistical aggregation model and the derived measure annotation certainty captures disagreement and the corresponding uncertainty well.

\begin{figure*}
    \centering
    \includegraphics[height=2.5cm]{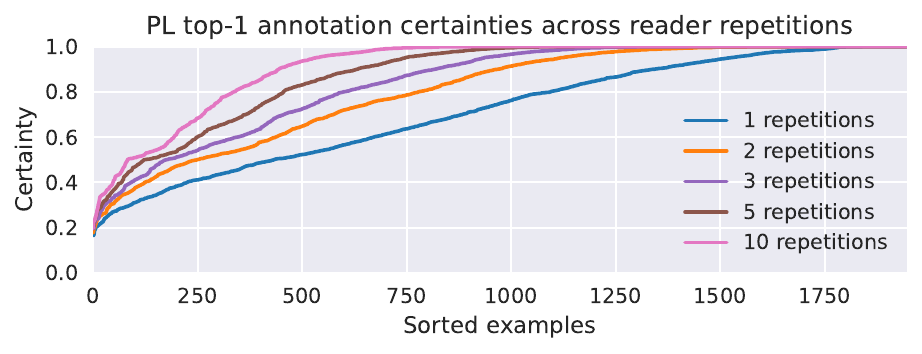}
    \includegraphics[height=2.5cm]{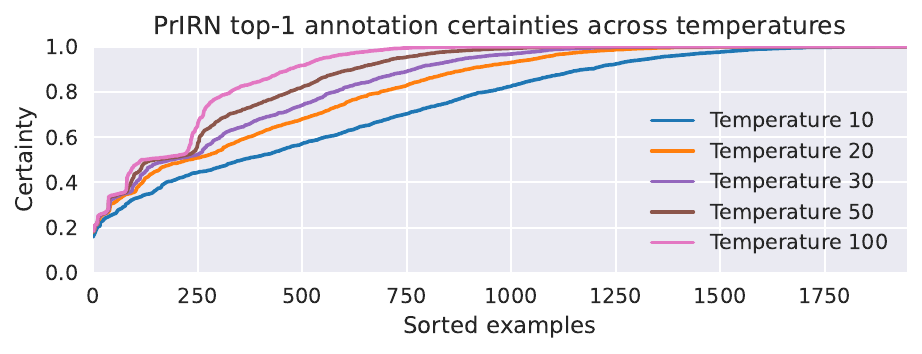}
    \caption{Top-1 annotation certainty for PL (left) and \PIRN across reliabilities. As detailed in the text, for PL, reliability is approximated by repeating annotations; for \PIRN, we can directly increase the corresponding temperature parameter.}
    \label{fig:app-results-reliability}
\end{figure*}
\textbf{Reliabity in PL and \PIRN:} Figure \ref{fig:app-results-reliability} plots top-1 annotation certainty for PL and \PIRN across reliabilities. As detailed in Section \ref{sec:methods-metrics}, reliability can be understood as the number of annotators in both cases. For PL, we explicitly repeat annotators as integrating a temperature into the statistical aggregation model makes sampling from the posterior infeasible; for \PIRN, in contrast, Section \ref{subsec:method-birn} allows the direct use of a temperature. As indicated by Figure \ref{fig:results-comparison} (left), average annotation certainty increases with reliability. This suggests that annotation certainty increases across examples. However, this is not entirely true; we clearly see a saturating effect for at least 250 cases where higher reliability does not seem to increase annotation certainty. This further supports our observation that the observed disagreement for many cases is actually due to inherent uncertainty, i.e., these cases are inherently ambiguous for annotators.

\red{
\begin{figure*}[t]
    \centering
    \includegraphics[height=3cm]{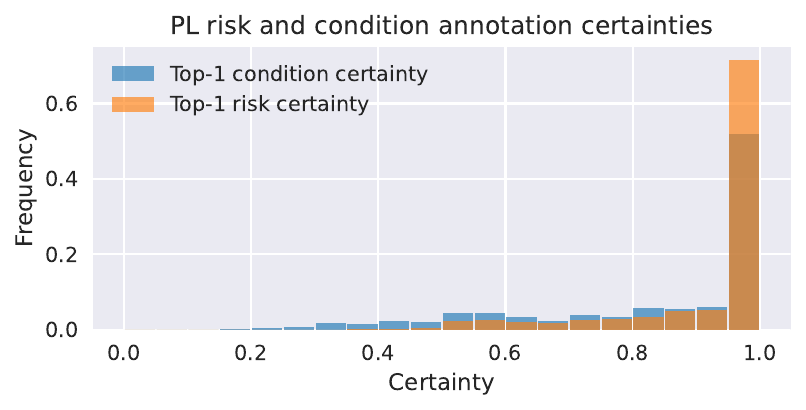}
    \caption{Distributions of top-1 risk and condition certainty, complementary to Figure \ref{fig:results-risk}.}
    \label{fig:app-results-risk-certainty}
\end{figure*}
\textbf{Top-1 risk and condition certanties:} Figure \ref{fig:app-results-risk-certainty} shows the distribution of top-1 risk and condition certainties, complementing Figure \ref{fig:results-risk} in the main paper. Clearly, more cases exhibit high top-1 risk certainty compared to top-1 condition certainty because there are only three risk categories of 419 different conditions. However, the fraction of examples with certainty below 99\% is with 38.2\% still very high (compared to 59\% for conditions).
}

\begin{figure*}[t]
    \centering
    \begin{minipage}{0.05\textwidth}
        \rotatebox{90}{Image and annotations}
    \end{minipage}
    \begin{minipage}{0.93\textwidth}
        \begin{minipage}{0.225\textwidth}
            \includegraphics[width=1\textwidth]{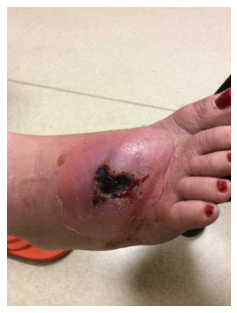}
        \end{minipage}
        \hfill
        \begin{minipage}{0.75\textwidth}
            \small
            A0: \{Pyoderma gangrenosum, Venous stasis ulcer\}\\
            A1: \{Arterial ulcer, Calciphylaxis cutis\}\\
            A2: \{Cellulitis\}
        \end{minipage}
    \end{minipage}
    
    \vskip 3px
    {\color{black!50!white}\rule{\textwidth}{1pt}}
    \vskip 3px
    
    \begin{minipage}{0.05\textwidth}
        \rotatebox{90}{Plausibilities}
    \end{minipage}
    \begin{minipage}{0.93\textwidth}
        \small\hspace{4cm}PL plausibilities\hspace{6cm} PRIRN plausibilities
        
        \includegraphics[trim={0 0 0 0.65cm},clip,width=0.49\textwidth]{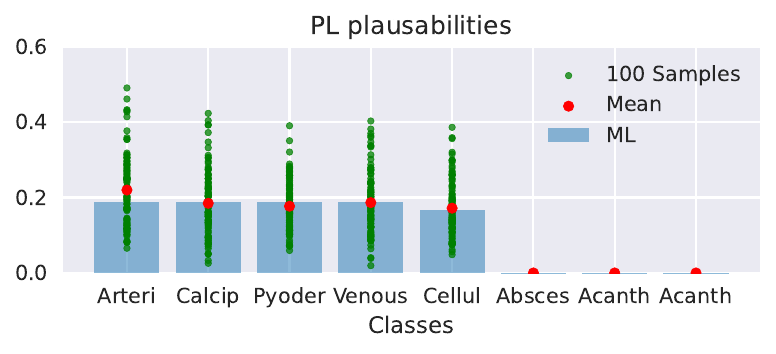}\includegraphics[trim={0 0 0 0.65cm},clip,width=0.49\textwidth]{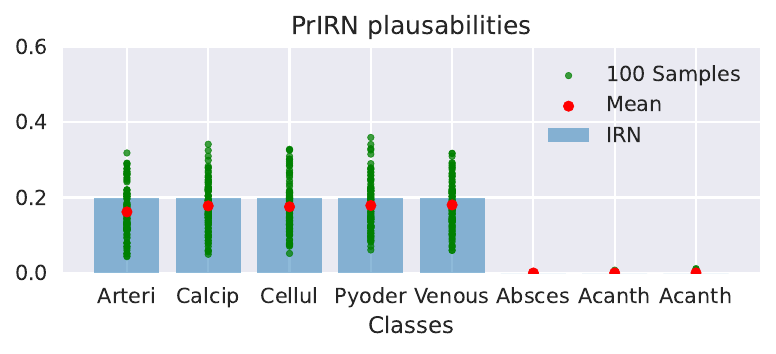}
    \end{minipage}
    
    \vskip 3px
    {\color{black!50!white}\rule{\textwidth}{1pt}}
    \vskip 3px
    
    \begin{minipage}{0.05\textwidth}
        \rotatebox{90}{Pred.}
    \end{minipage}
    \begin{minipage}{0.93\textwidth}
        \begin{minipage}{0.49\textwidth}
            \small
            Model A: $C_{\text{top-}3} =$\\ \{Diabetic ulcer, Pyoderma Gangrenosum, Venous Stasis Ulcer\}
        \end{minipage}
        \hfill
        \begin{minipage}{0.49\textwidth}
            \small
            Model D: $C_{\text{top-}3} =$\\ \{Arterial ulcer, Cellulitis, Pyoderma Gangrenosum\}
        \end{minipage}
    \end{minipage}
    
    \vskip 3px
    {\color{black!50!white}\rule{\textwidth}{1pt}}
    \vskip 3px
    
    \begin{minipage}{0.05\textwidth}
        \rotatebox{90}{Eval.}
    \end{minipage}
    \begin{minipage}{0.93\textwidth}
        \begin{minipage}{0.49\textwidth}
            \small
            \centering
            \begin{tabular}{|r | c | c|}
                Model A & PL & \PIRN\\\hline
                $\E_{p(\lambda| b, x)}\{ \delta[Y_{\text{top-}1}(\lambda) \subset C_{\text{top-}k}(x)]\}$ & $0.39$ & $0.41$
            \end{tabular}
        \end{minipage}
        \hfill
        \begin{minipage}{0.49\textwidth}
            \small
            \centering
            \begin{tabular}{|r | c | c|}
                Model D & PL & \PIRN\\\hline
                $\E_{p(\lambda| b, x)}\{ \delta[Y_{\text{top-}1}(\lambda) \subset C_{\text{top-}k}(x)]\}$ & $0.58$ & $0.61$
            \end{tabular}
        \end{minipage}
    \end{minipage}
    
    \vskip 3px
    {
    \color{black!50!white}\rule{\textwidth}{1pt}
    }
    \vskip 3px
    
    \begin{minipage}{0.05\textwidth}
        \rotatebox{90}{Image and annotations}
    \end{minipage}
    \begin{minipage}{0.93\textwidth}
        \begin{minipage}{0.225\textwidth}
            \includegraphics[width=1\textwidth]{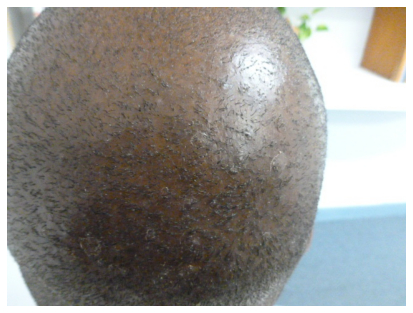}
        \end{minipage}
        \hfill
        \begin{minipage}{0.75\textwidth}
            \small
            A0: \{Dissecting cellulitis of scalp, Tinea\} \{Seborrheic dermatitis\}\\
            A1: \{Acne keloidalis, Folliculitis decalvans\}\\
            A2: \{Folliculitis\}
        \end{minipage}
    \end{minipage}
    
    \vskip 3px
    {\color{black!50!white}\rule{\textwidth}{1pt}}
    \vskip 3px
    
    \begin{minipage}{0.05\textwidth}
        \rotatebox{90}{Plausibilities}
    \end{minipage}
    \begin{minipage}{0.93\textwidth}
        \small\hspace{4cm}PL plausibilities\hspace{6cm} PRIRN plausibilities
        
        \includegraphics[trim={0 0 0 0.65cm},clip,width=0.49\textwidth]{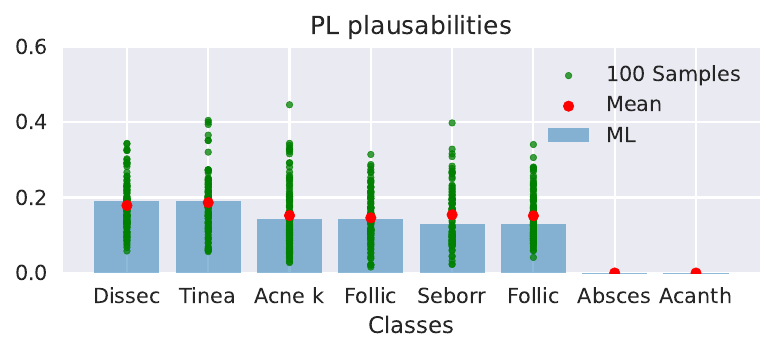}        \includegraphics[trim={0 0 0 0.65cm},clip,width=0.49\textwidth]{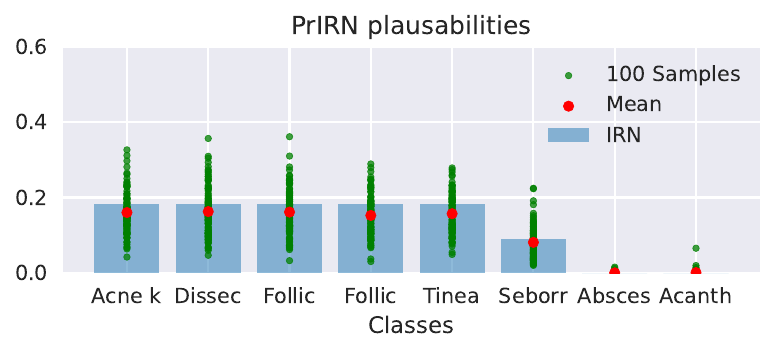}
    \end{minipage}
    
    \vskip 3px
    {\color{black!50!white}\rule{\textwidth}{1pt}}
    \vskip 3px
    
    \begin{minipage}{0.05\textwidth}
        \rotatebox{90}{Pred.}
    \end{minipage}
    \begin{minipage}{0.93\textwidth}
        \begin{minipage}{0.49\textwidth}
            \small
            Model A: $C_{\text{top-}3} =$\\ \{Acne, Folliculitis, Folliculitis decalvans\}
        \end{minipage}
        \hfill
        \begin{minipage}{0.49\textwidth}
            \small
            Model B: $C_{\text{top-}3} =$\\ \{Acne keloidalis, Folliculitis, Folliculitis decalvans\}
        \end{minipage}
    \end{minipage}
    
    \vskip 3px
    {\color{black!50!white}\rule{\textwidth}{1pt}}
    \vskip 3px
    
    \begin{minipage}{0.05\textwidth}
        \rotatebox{90}{Eval.}
    \end{minipage}
    \begin{minipage}{0.93\textwidth}
        \begin{minipage}{0.49\textwidth}
            \small
            \centering
            \begin{tabular}{|r | c | c|}
                Model A & PL & \PIRN\\\hline
                $\E_{p(\lambda| b, x)}\{ \delta[Y_{\text{top-}1}(\lambda) \subset C_{\text{top-}k}(x)]\}$ & $0.27$ & $0.4$
            \end{tabular}
        \end{minipage}
        \hfill
        \begin{minipage}{0.49\textwidth}
            \small
            \centering
            \begin{tabular}{|r | c | c|}
                Model B & PL & \PIRN\\\hline
                $\E_{p(\lambda| b, x)}\{ \delta[Y_{\text{top-}1}(\lambda) \subset C_{\text{top-}k}(x)]\}$ & $0.42$ & $0.62$
            \end{tabular}
        \end{minipage}
    \end{minipage}
    \caption{Two additional qualitative results corresponding to cases with very low annotation certainty, following the presentation of Figure \ref{fig:results-motivation}.}
    \label{fig:app-results-2}
\end{figure*}
\textbf{Qualitative results:} Figure \ref{fig:app-results-2} shows two additional qualitative cases where we observed very low annotation certainty (i.e., in Figure \ref{fig:app-results-reliability}, these likely lie on the very left). In both cases, annotators provided widely different conditions, leading PL and \PIRN to distribute the plausibility mass nearly equally across the annotated conditions. Again, this also has significant impact on evaluating the top-3 prediction sets of our classifiers.

\begin{figure*}[t]
    \centering
    \begin{minipage}{0.05\textwidth}
        \rotatebox{90}{Annotations}
    \end{minipage}
    \begin{minipage}{0.93\textwidth}
        A0: \{SK/ISK\}\\
        A1: \{Actinic Keratosis\}\\
        A2: \{SK/ISK\}\\
        A3: \{Verruca vulgaris\}\\
        A4: \{Lentigo\}\\
        A5: \{Lentigo\}
    \end{minipage}
    
    \vskip 3px
    {\color{black!50!white}\rule{\textwidth}{1pt}}
    \vskip 3px
    
    \begin{minipage}{0.05\textwidth}
        \rotatebox{90}{Plausibilities}
    \end{minipage}
    \begin{minipage}{0.93\textwidth}
        \small\hspace{4cm}PL plausibilities\hspace{6cm} PRIRN plausibilities
        
        \includegraphics[trim={0 0 0 0.65cm},clip,width=0.49\textwidth]{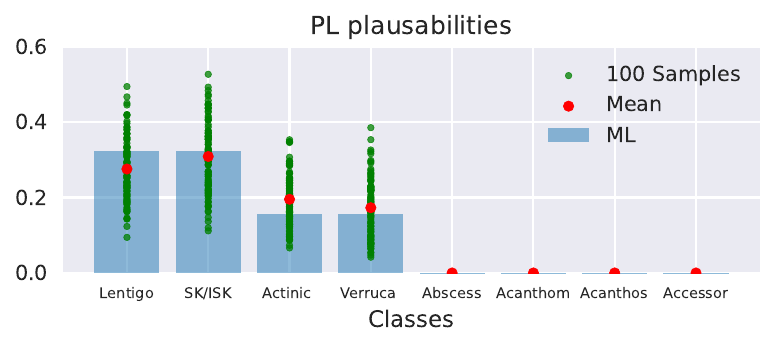}\includegraphics[trim={0 0 0 0.65cm},clip,width=0.49\textwidth]{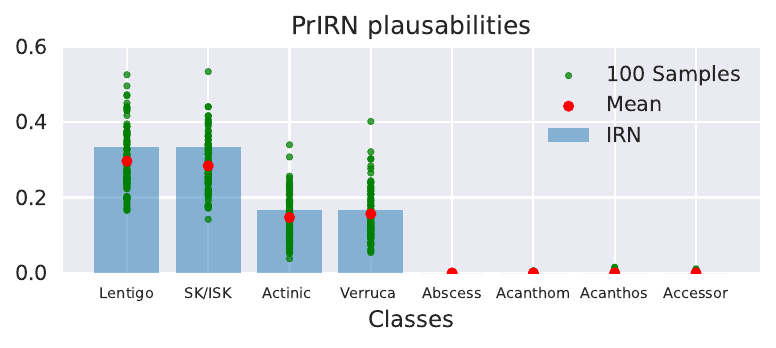}
    \end{minipage}
    
    \vskip 3px
    {\color{black!50!white}\rule{\textwidth}{1pt}}
    \vskip 3px
    
    \begin{minipage}{0.05\textwidth}
        \rotatebox{90}{Annot.}
    \end{minipage}
    \begin{minipage}{0.93\textwidth}
        \small
        A0: \{Cylindroma of skin, SK/ISK\} \{Cyst\}\\
        A2: \{Nevus sebaceous\}\\
        A3: \{Basal Cell Carcinoma\}\\
        A4: \{Nevus sebaceous\}
    \end{minipage}
    
    \vskip 3px
    {\color{black!50!white}\rule{\textwidth}{1pt}}
    \vskip 3px
    
    \begin{minipage}{0.05\textwidth}
        \rotatebox{90}{Plausibilities}
    \end{minipage}
    \begin{minipage}{0.93\textwidth}
        \small\hspace{4cm}PL plausibilities\hspace{6cm} PRIRN plausibilities
        
        \includegraphics[trim={0 0 0 0.65cm},clip,width=0.49\textwidth]{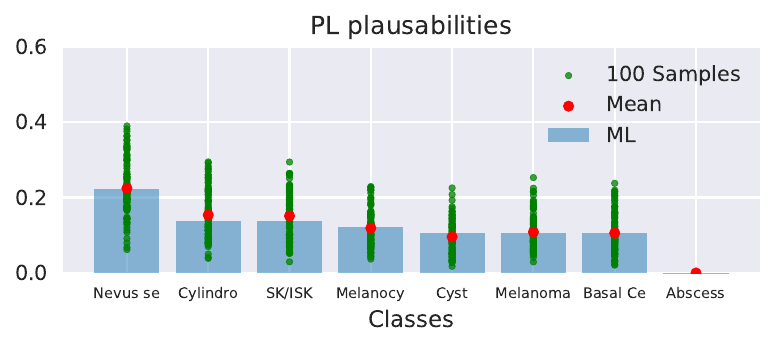}\includegraphics[trim={0 0 0 0.65cm},clip,width=0.49\textwidth]{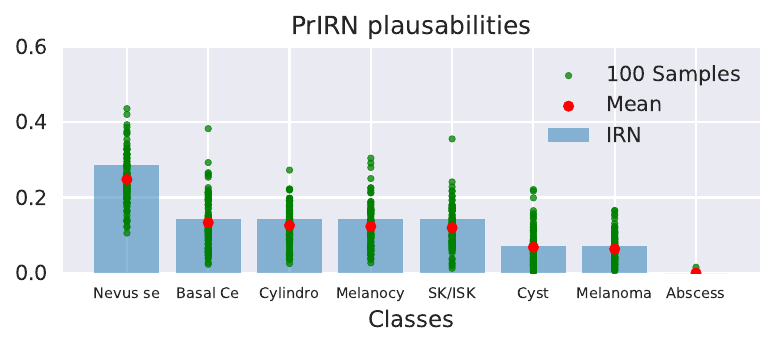}
    \end{minipage}
    
    \vskip 3px
    {\color{black!50!white}\rule{\textwidth}{1pt}}
    \vskip 3px
    
    \begin{minipage}{0.05\textwidth}
        \rotatebox{90}{Annot.}
    \end{minipage}
    \begin{minipage}{0.93\textwidth}
        \small
        A0: \{Sweet syndrome\}\\
        A1: \{Drug Rash, Eczema\}\\
        A2: \{Actinic Keratosis\} \{Cutaneous lupus\}
    \end{minipage}
    
    \vskip 3px
    {\color{black!50!white}\rule{\textwidth}{1pt}}
    \vskip 3px
    
    \begin{minipage}{0.05\textwidth}
        \rotatebox{90}{Plausibilities}
    \end{minipage}
    \begin{minipage}{0.93\textwidth}
        \small\hspace{4cm}PL plausibilities\hspace{6cm} PRIRN plausibilities
        
        \includegraphics[trim={0 0 0 0.65cm},clip,width=0.49\textwidth]{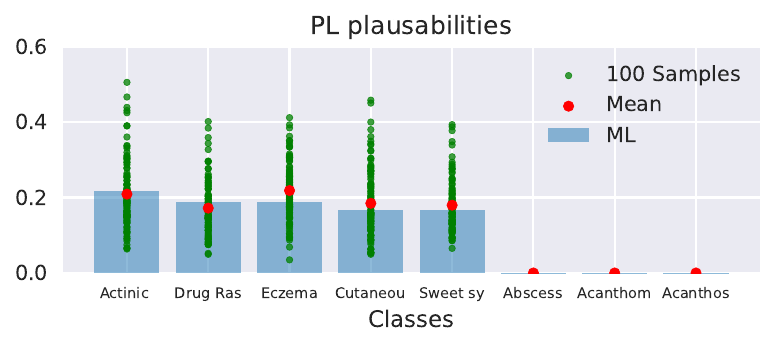}\includegraphics[trim={0 0 0 0.65cm},clip,width=0.49\textwidth]{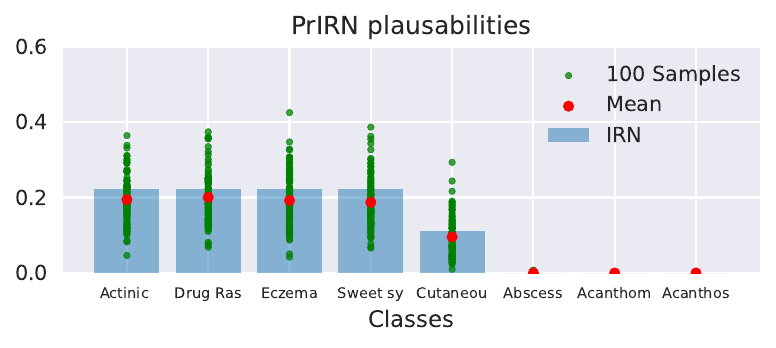}
    \end{minipage}
    
    \vskip 3px
    {\color{black!50!white}\rule{\textwidth}{1pt}}
    \vskip 3px
    
    \begin{minipage}{0.05\textwidth}
        \rotatebox{90}{Annotations}
    \end{minipage}
    \begin{minipage}{0.93\textwidth}
        \small
        A0: \{Psoriasis\} \{Drug Rash\}\\
        A1: \{Drug Rash, Eczema\} \{Psoriasis\}\\
        A2: \{Psoriasis\} \{Cutaneous T Cell Lymphoma, Eczema\}\\
        A3: \{Drug Rash, Eczema\} \{Cutaneous T Cell Lymphoma\}
    \end{minipage}
    
    \vskip 3px
    {\color{black!50!white}\rule{\textwidth}{1pt}}
    \vskip 3px
    
    \begin{minipage}{0.05\textwidth}
        \rotatebox{90}{Plausibilities}
    \end{minipage}
    \begin{minipage}{0.93\textwidth}
        \small\hspace{4cm}PL plausibilities\hspace{6cm} PRIRN plausibilities
        
        \includegraphics[trim={0 0 0 0.65cm},clip,width=0.49\textwidth]{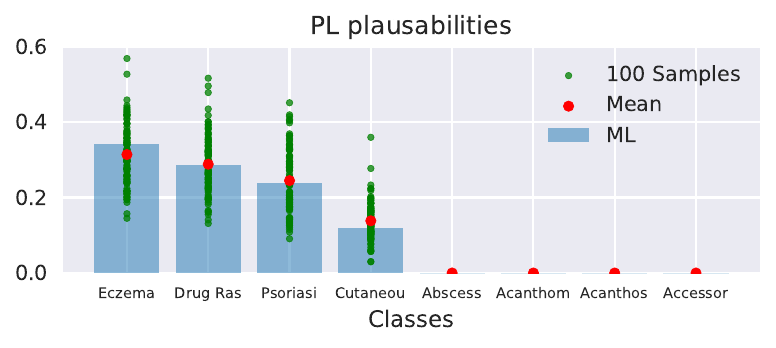}\includegraphics[trim={0 0 0 0.65cm},clip,width=0.49\textwidth]{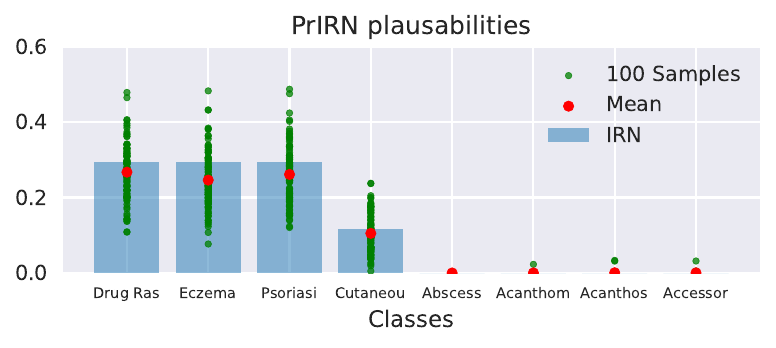}
    \end{minipage}
    \caption{Two additional qualitative results focusing on some differences between PL and \PIRN plausibilities.}
    \label{fig:app-results-3}
\end{figure*}

\red{
\textbf{Differences between PL and \PIRN:} In Figure \ref{fig:app-results-3} we show further qualitative results highlighting some cases with differences between the PL and \PIRN plausibilities.
}

\end{appendix}

\end{document}